\documentclass{article}

\usepackage{microtype}
\usepackage{graphicx}
\usepackage{subfigure}
\usepackage{booktabs} 
\usepackage{subcaption}
\usepackage{multirow}
\usepackage{hyperref}
\usepackage{diagbox}

\usepackage[accepted]{icml2024}

\usepackage{amsmath}
\usepackage{amssymb}
\usepackage{mathtools}
\usepackage{amsthm}

\usepackage[capitalize,noabbrev]{cleveref}

\theoremstyle{plain}
\newtheorem{theorem}{Theorem}

\newtheorem{remark}[theorem]{Remark}

\newtheorem{definition}[theorem]{Definition}

\usepackage[textsize=tiny]{todonotes}

\allowdisplaybreaks[2]
\usepackage{subcaption}
\usepackage{mismath}

\usepackage{mathtools}
\usepackage{amsmath}
\usepackage{amsthm}
\usepackage{amssymb}
\usepackage{algorithm}
\usepackage{algorithmic}
\usepackage{graphicx}
\usepackage{setspace}
\usepackage{color}
\usepackage{cite}
\usepackage{wrapfig}
\usepackage{lipsum}
\usepackage{xcolor}
\usepackage{pgffor}

\newcommand{\adjp}[1]{\operatorname{adj} \left( #1 \right) }
\newcommand{\adjj}[1]{\operatorname{adj} #1}

\definecolor{turquoise}{RGB}{0, 131, 133}

\usepackage{makecell}
\usepackage{multicol}
\usepackage{multirow}
\usepackage{siunitx}

\icmltitlerunning{On Diffusion Process in SE(3)-invariant Space}

\begin{document}

\twocolumn[
\icmltitle{On Diffusion Process in SE(3)-invariant Space}

\begin{icmlauthorlist}
\icmlauthor{Zihan Zhou}{cuhksz}
\icmlauthor{Ruiying Liu}{cuhksz}
\icmlauthor{Jiachen Zheng}{cuhksz}
\icmlauthor{Xiaoxue Wang}{chemlex}
\icmlauthor{Tianshu Yu}{cuhksz,airs}


\end{icmlauthorlist}

\icmlaffiliation{cuhksz}{The Chinese University of Hong Kong, Shenzhen}
\icmlaffiliation{airs}{Shenzhen Institute of Artificial Intelligence and Robotics for Society}
\icmlaffiliation{chemlex}{Chemlex}

\icmlcorrespondingauthor{Tianshu Yu}{yutianshu@cuhk.edu.cn}

\icmlkeywords{Machine Learning, ICML, Diffusion, Manifold Diffusion, SE(3)-invariant, 3D Generation}

\vskip 0.3in
]

\printAffiliationsAndNotice{}
\begin{abstract}

Sampling viable 3D structures (e.g., molecules and point clouds) with SE(3)-invariance using diffusion-based models proved promising in a variety of real-world applications, wherein SE(3)-invariant properties can be naturally characterized by the inter-point distance manifold. However, due to the non-trivial geometry, we still lack a comprehensive understanding of the diffusion mechanism within such SE(3)-invariant space. This study addresses this gap by mathematically delineating the diffusion mechanism under SE(3)-invariance, via zooming into the interaction behavior between coordinates and the inter-point distance manifold through the lens of differential geometry. Upon this analysis, we propose accurate and projection-free diffusion SDE and ODE accordingly. Such formulations enable enhancing the performance and the speed of generation pathways; meanwhile offering valuable insights into other systems incorporating SE(3)-invariance.
 
\end{abstract}

\section{Introduction} \label{sec:intro}
Recently, there has been an increasing interest in utilizing diffusion-based generative models to sample 3-dimensional (3D) structures~\citep{shi2021learning, luo2021predicting, xu2022geodiff, jing2022torsional, zhou2023molecular, fan2023generative, fan2023ec, zhang2023sdegen}. These approaches have demonstrated promising modeling ability in capturing the wide distribution of 3D coordinates, wherein the incorporation of the inherent roto-translational property of SE(3)-invariant space is considered as one of the key factors potentially leading to these successes~\citep{de2022riemannian}.

For 3D structures, the inter-node distance matrix is inherently equipped with SE(3)-invariance, thus being a natural and flexible choice for modeling the underlying diffusion process. To perform diffusion-based generation on this manifold, estimating the associated score function is essential~\citep{song2020denoising}. Unfortunately, the geometry over this manifold is highly irregular and non-trivial, yielding significant difficulty in mathematical and accurate estimation. While some recent works simplify the diffusion process with Gaussian~\citep{xu2022geodiff} or Maxwell-Boltzmann~\citep{zhou2023molecular} assumptions, these heuristics may result in sub-optimal performance. Till now, extant literature has not systematically undertaken an in-depth analysis of the diffusion process in this manifold.

In parallel, some other trials seek to avoid direct manipulations on this manifold. Certain methodologies address the SE(3) group~\citep{ganea2021geomol}, offering a means to overcome the SE(3)-invariance obstacle through prior knowledge. In the context of molecular conformation generation tasks, specific endeavors employ RDkit to initially generate local 3D structures, subsequently employing diffusion techniques to derive torsional angles~\citep{jing2022torsional}. Alternatively, models geared towards protein generation delve into the SE(3) Lie group, capitalizing on the intrinsic dependency of protein properties on bond angles and lengths~\citep{anand2022protein, yim2023se}. However, the applicability of these modeling schemes is confined to specific tasks and proves inflexible to extend to generic 3D coordinate generation. In the case of point clouds, the majority of existing studies do not explicitly formulate designs for SE(3)-invariance~\citep{zhou20213d, luo2021diffusion, mo2023dit, zeng2022lion}. Instead, they commonly employ diffusion techniques directly on node embeddings. Notably, some achieve commendable performance by leveraging transformer architectures, anticipating the model to implicitly learn SE(3)-invariance in brute force.

\begin{figure}
    \centering
    \includegraphics[width=.15\textwidth]{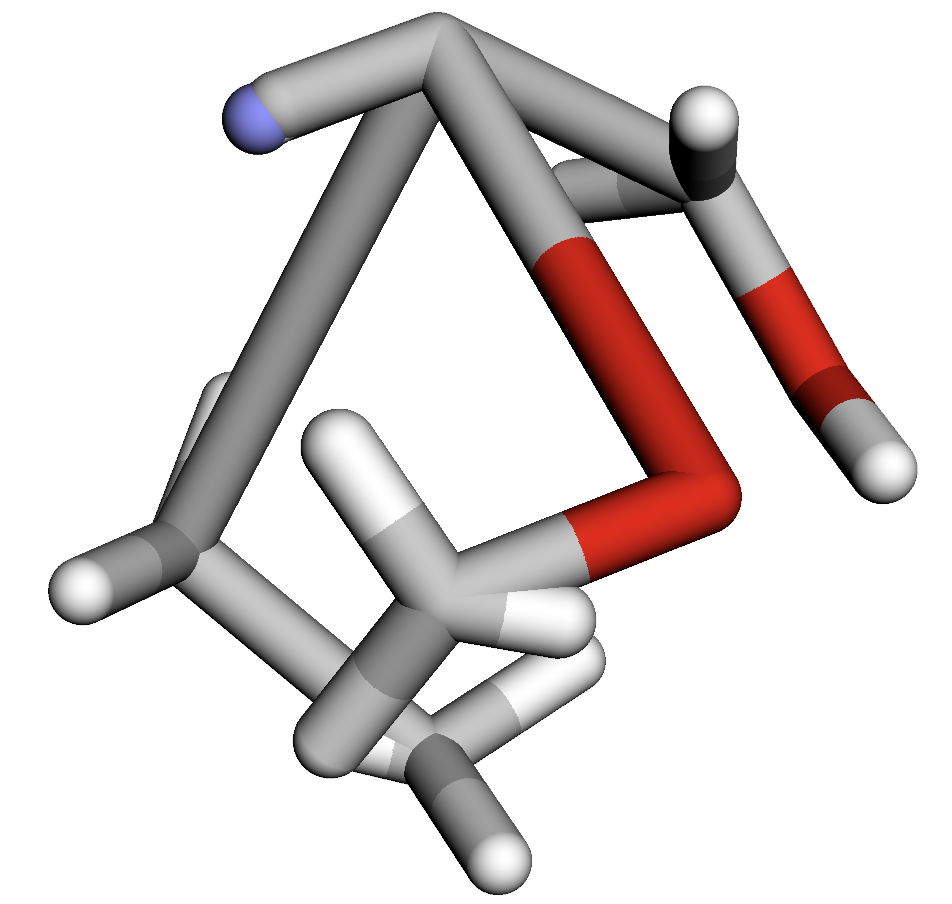}
    \includegraphics[width=.15\textwidth]{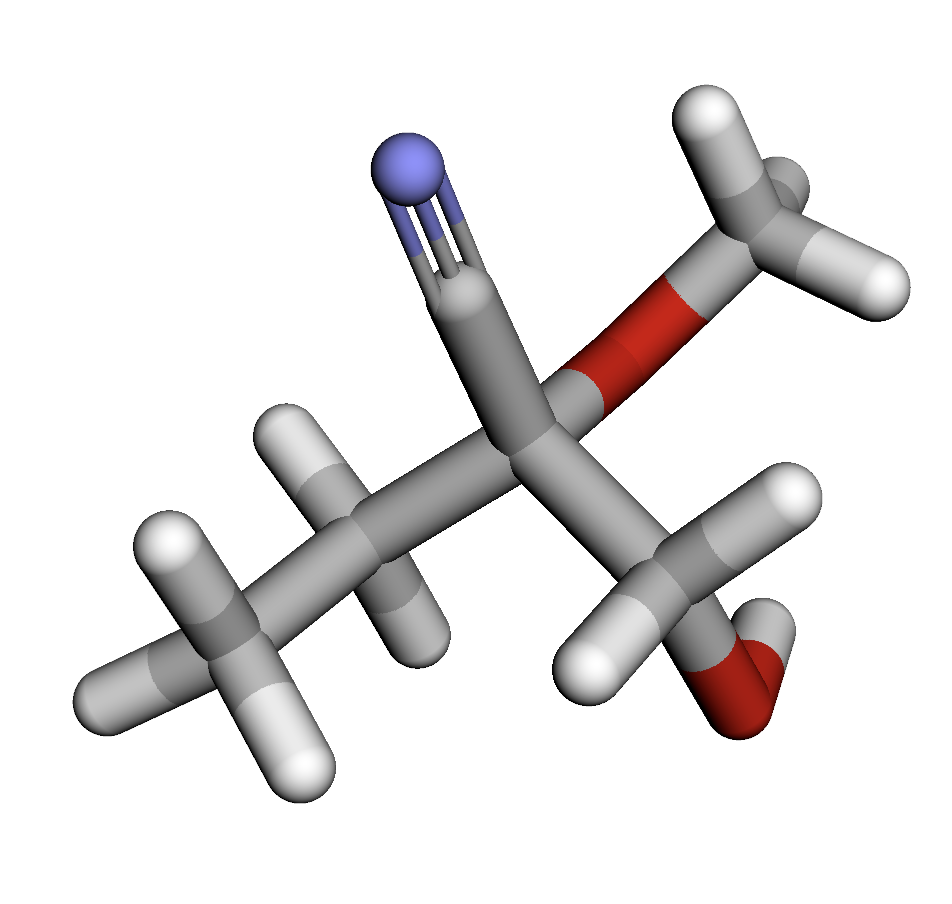}
    \caption{We utilized a sampling process with the off-the-shelf DPM-solver~\citep{lu2022dpm} and our method in reduced sampling iterations (1000 steps) on a GeoDiff~\citep{xu2022geodiff} model that was trained for 5000 steps. An example of a molecular conformer sampled via DPM-solver appears to be unreasonable (left), while our method produces a high-quality conformer (right).}
    \label{fig:intro-contract}
\end{figure}
The lack of a mathematical understanding of existing works raises additional concerns. Without accurate modeling, the reverse ODE and SDE formats of the diffusion process in SE(3)-invariant space remains opaque. Therefore, existing approximations usually sample 3D structures via naive Langevin dynamics (LD) with thousands of steps~\citep{shi2021learning, xu2022geodiff}. Directly applying modern solvers (e.g., DPM solver and high-order solvers~\citep{dormand1980family, lu2022dpm}) is problematic and will produce unmeaningful structures. See Fig.~\ref{fig:intro-contract} for an example. Besides, a coarse diffusive framework on Riemannian manifold~\citep{de2022riemannian} requires projection operation on each sampling step, failing to be effectively accelerated as well.

To better understand the diffusion process in SE(3)-invariant space and facilitate downstream tasks, in this paper, we systematically investigate the correlated diffusion process in this space from the perspective of differential geometry and projected differential equations. To this end, we mathematically derive the transformation between inter-point distance change and coordinate shift, delivering linear operations with bounded errors to mimic a projection into SE(3)-invariant space. As such, we propose projection-free SDE and ODE formulations to reverse the diffusion process. These findings offer useful insights into the mathematical mechanism of diffusion in SE(3)-invariant space. Extensive experiments are conducted on molecular conformation generation~\citep{axelrod2022geom} and pose skeleton generation~\citep{ionescu2013human3}. Our scheme can sample high-quality 3D coordinates with only a few steps. 
In summary, our contributions are:
\begin{itemize}
    \item We systematically investigate how to accurately model the diffusion-based generation in SE(3)-invariant space by considering the adjacency matrices for 3D coordinate generation tasks. 
    \item We analyze the mappings between the SE(3)-invariant Cartesian space and matrix manifold and propose a linear operation to approximate the projection operation with bounded errors.
    \item We propose a novel projection-free sampling scheme by solving the reverse SDE and ODE via samplings between coordinates and adjacency matrices, allowing for simple acceleration. 
\end{itemize}

\section{Related works}
\paragraph{Coordinate generation from SE(3)-invariant distribution.}
Generating 3D coordinates of geometric structures (e.g., molecular conformation and point clouds) is a challenging task. 
In the context of molecular conformation or protein structure generation, some notable approaches focus on modeling the bound length and bound angles by considering the SE(3) group~\citep{ganea2021geomol, jing2022torsional, yim2023se, anand2022protein}. In contrast, DMCG~\citep{zhu2022direct} presents an end-to-end solution by directly predicting atom coordinates while ensuring roto-translational invariance through an SE(3)-invariant loss function. More recently, there has been growing interest in modeling the inter-atomic distances, which naturally possess SE(3)-invariance. Certain approaches~\citep{shi2021learning, zhang2023sdegen, luo2021predicting, xu2022geodiff, zhou2023molecular} introduce perturbations to the inter-atomic distances and subsequently estimate the corresponding coordinate scores. In the domain of point cloud generation, most works directly introduce noise to coordinates or latent features~\citep{luo2021diffusion, mo2023dit, zeng2022lion}, assuming that the learned point embeddings can implicitly incorporate SE(3) invariance~\citep{wen2021learning, zamorski2020adversarial, wu2016learning}.

\paragraph{Sampling acceleration of diffusion-based models.}
Standard diffusion-based generation typically takes thousands of steps. To accelerate this, some new diffusion models such as consistency models~\citep{song2023consistency} are proposed. Consistency models directly map noise to data, enabling the generation of images in only a few steps. EC-Conf~\citep{fan2023ec} is an application of consistency models in substantially reduced steps for molecular conformation generation. Another straightforward approach is reducing the number of sampling iterations. DDIM~\citep{song2020denoising} uses a hyper-parameter to control the sampling stochastical level and finds that decreasing the reverse iteration number improves sample quality due to less stochasticity. This phenomenon can be attributed to the existence of a probability ODE flow associated with the stochastic Markov chain of the reverse diffusion process~\citep{song2020score}, leading to the advent of numerical ODE solvers for acceleration. 
DPM-solver~\citep{lu2022dpm} leverages the semi-linear structure of probability flow ODE to develop customized ODE solvers, and also provides high-order approximates for the probability ODE flow which can generate high-quality samples in only 10 to 20 iterations. Then DPM is extended to DPM-solver++~\citep{lu2022dpm+} for sampling with classifier-free guidance. 
However, such a dedicated solver can only be applied in Euclidean space, and developing diffusion solvers for SE(3)-invariant space or on the pairwise distance manifold remains an open problem.

\section{Preliminary}

\subsection{Euclidean diffusion and score matching objective} 
Given a random variable $\mathbf{x}_0 \in \mathbb{R}^n$ following an unknown data distribution $q_0$, denoising diffusion probabilistic models~\citep{song2019generative, ho2020denoising, song2020score} define a forward process $\{ \mathbf{x}_t \}_{t \in [0, T]}$ following an Ito SDE
\begin{equation}
    \mathrm{d} \mathbf{x}_t = f(t) \mathbf{x}_t \mathrm{d} t + g(t) \mathrm{d} \mathbf{w}_t, \quad \mathbf{x}_0 \sim q_0,
\end{equation}
where $\mathbf{w}_t \in \mathbb{R}^n$ is the standard Brownian motion and $f(t)$, $g(t)$ are chosen functions. Such a forward process has an equivalent reverse process starting from time $T$ to $0$:
\begin{equation}
    \begin{aligned}
    \mathrm{d} \mathbf{x}_t &= \left[ f(t) \mathbf{x}_t - g^2(t) \nabla_{\mathbf{x}} \log q_t (\mathbf{x}_t) \right] \mathrm{d} t + g(t) \mathrm{d} \mathbf{w}_t,  \\
    \end{aligned}
\end{equation}
where $\mathbf{x}_T \sim q_{T}(\mathbf{x}_T \mid \mathbf{x}_0) \approx q_{T}(\mathbf{x}_T)$, and the marginal probability densities $\{ q_{t}(\mathbf{x}_t) \}_{t=0}^T$ of the above SDE is the same as the following probability flow ODE~\citep{song2020score}:
\begin{equation}
    \frac{\mathrm{d} \mathbf{x}_t}{\mathrm{d} t} = f(t) \mathbf{x}_t - \frac{1}{2} g^2(t) \nabla_{\mathbf{x}} \log q_t (\mathbf{x}_t). \label{eq: general reverse ode1}
\end{equation}
This implies that if we can sample from $q_{T}(\mathbf{x}_T) \approx q_{T}(\mathbf{x}_T \mid \mathbf{x}_0)$ and solve Eq.~\eqref{eq: general reverse ode1}, then the obtained $\mathbf{x}_0$ follows the distribution of $q_0 (\mathbf{x}_0)$. The only unknown terms in Eq.~\eqref{eq: general reverse ode1} are $q_{T}(\mathbf{x}_T)$ and $\nabla_{\mathbf{x}} \log q_t (\mathbf{x}_t)$. By choosing some specific $f(t)$ and $g(t)$, the distribution $q_{T}(\mathbf{x}_T \mid \mathbf{x}_0)$ converges to $ q_{T}(\mathbf{x}_T)$ as $T \rightarrow \infty$ and $q_{T}$ is an easy distribution for sampling like Gaussian distribution. The score of the diffused data $\nabla_{\mathbf{x}} \log q_t (\mathbf{x}_t)$ can be approximatively learned by a score-based model.

\subsection{Diffusion on manifolds} 
In mathematical terms, a probability distribution $p(\mathbf{x})$ is considered invariant concerning the $\operatorname{SE}(3)$ group if $p (T(\mathbf{x})) = p (\mathbf{x})$ for any arbitrary transformations $T$ belonging to the $\operatorname{SE}(3)$ group. In terms of 3D coordinate generation following an SE(3)-invariant distribution, an effective and general strategy is to employ adjacency matrices or inter-atomic distances, which naturally exhibit invariance concerning the $\operatorname{SE}(3)$ groups~\citep{gasteiger2020directional, shi2021learning, xu2022geodiff}. An adjacency matrix $d = [ d_{ij} ] \in \mathbb{R}^{n \times n}_{+}$ is said to be \emph{feasible} if there is a set of coordinates $\mathcal{C} = [\mathbf{x}_1, \ldots, \mathbf{x}_n]^\top $ such that $d_{ij} = \| \mathbf{x}_i - \mathbf{x}_j \|, \forall i, j = 1, \ldots, n$. Opposite to the usual diffusion on feature vectors laying in Euclidean space $\mathbb{R}^{n}$, feasible adjacency matrices only lay in a subset of the Euclidean space $\mathbb{R}^{n \times n}$. This subset is a manifold and we call this manifold the adjacency matrix manifold. If we directly perturb or add Gaussian noise to the adjacency matrix, we may obtain an infeasible adjacency matrix. To overcome such difficulty, we need to consider the context of diffusion on Riemannian manifolds~\citep{de2022riemannian}, which forces the diffusion objective to always lie in the specified manifolds. However, diffusion in manifolds often requires thousands of iterations and projection operations in each iteration, which hinders an effective sampling algorithm with fewer steps.

\section{Method}
\begin{figure*}[tb]
    \centering
    \includegraphics[width=.7\textwidth]{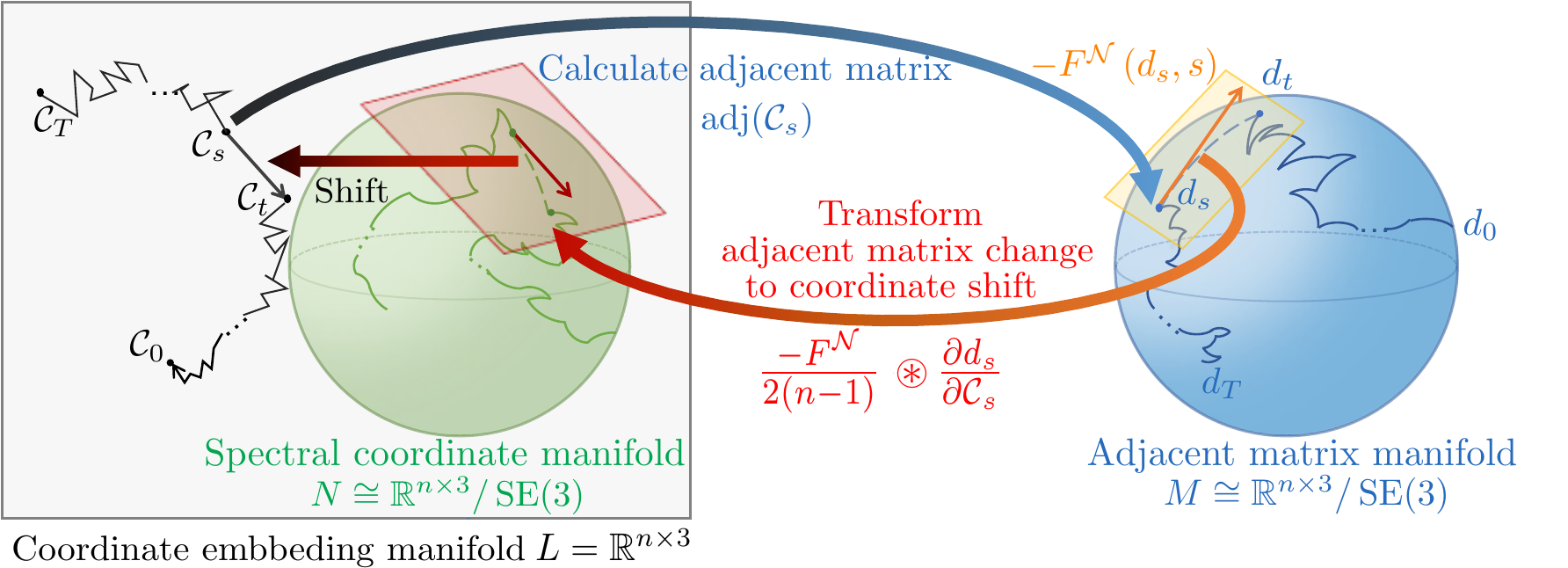}
    \caption{Mathematical framework for the reverse diffusion in the SE(3)-invariant space $\mathbb{R}^{n \times 3} / \operatorname{SE}(3)$. The grey rectangle denotes the coordinate embedding matrix manifold $L = \mathbb{R}^{n \times 3}$. The green ball denotes an SE(3)-invariant manifold $N$ of spectral coordinates (see Theorem~\ref{thm: mapping from adjacency matrix to coordinates}) that is a submanifold of $L$. The blue ball denotes the adjacency matrix manifold $M$. Detailed definitions for the above three manifolds can be found in Appendix~\ref{sec: manifold definitions}. Each coordinate in the spectral coordinate manifold $N$ has a unique adjacency matrix in the adjacency matrix manifold $M$, and vice versa. During each step of the denoising process, given a noised sample $\mathcal{C}_s \in L$, our model predicts the change of the adjacency matrix $d_s = \adjp{\mathcal{C}_s}$, illustrated by the orange tangent vector $-F^{\mathcal{N}}\left( d_s, s \right)$ in the tangent space $T_{d_s}M$. Then we transform such a change of the adjacency matrix to the spectral coordinate shift by $\frac{-F^{\mathcal{N}}}{2(n-1)} \circledast \frac{\partial d_s}{\partial \mathcal{C}_s}$ (denoted by the red tangent vector in the tangent space of the spectral coordinate manifold). At the implementation level, we can further transform the tangent vector of spectral coordinates into the change of coordinates in the coordinate embedding space $L$ since this operation does not affect the values of the adjacency matrix.}
    \label{fig: framework}
\end{figure*}

In this work, we aim to explicitly compute the reverse SDE/ODE of SE(3)-invariant diffusion models. We consider SE(3)-invariant diffusion models equipped with pairwise distances and seek to develop a forward and reverse diffusion process on the adjacency matrix with the SE(3)-invariance. The overall mathematical scheme can be seen from Fig.~\ref{fig: framework}. 

\subsection{Diffusion SDE of coordinates via adjacency matrices}
Directly injecting noise on the adjacency matrix is inapplicable since this will violate the geometric properties of 3D structures (e.g., triangle inequality). Riemannian manifold diffusion~\citep{de2022riemannian} addresses this issue by performing projection on each step but typically requiring thousands of steps. Besides, there is no literature on accelerating the Riemannian diffusion process. For the above two reasons, instead of directly working on the adjacency matrix manifold, we choose to inject noise on coordinates but consider modeling the corresponding evolving adjacency matrix, similar to~\citet{xu2022geodiff}. Given a set of 3D coordinates $\mathcal{C}_0 \in \mathbb{R}^{n \times 3} / \operatorname{SE}(3)$ and let $\mathcal{C}_0$ be represented or embedded by a $n \times 3$ matrix, i.e., $\mathcal{C}_0 = [\mathbf{x}_1, \ldots, \mathbf{x}_n]^\top \in \mathbb{R}^{n \times 3}$, we define a forward diffusion process of the coordinates embedding (we sometimes consider $\mathcal{C}_0$ as a vector to simplify our notations)~\citep{song2020score}
\begin{equation}
    \mathrm{d} \mathcal{C}_t = \mathbf{f}(\mathcal{C}_t, t) \mathrm{d} t + \mathbf{G}(\mathcal{C}_t, t) \mathrm{d} \mathbf{w}_t, \label{eq: method part coordinate diffusion}
\end{equation}
where $\mathcal{C}_0 \sim q_0(\mathcal{C}_0)$ and $q_0$ is the distribution of the dataset, $t \in [0, T]$, $\mathbf{f}$ and $\mathbf{G}$ are two hyperparameters of noise scheduler and $\mathbf{w}_t \in \mathbb{R}^n$ is the standard Brownian motion. Let $p_t(\mathcal{C}_t)$ be the distribution of $\mathcal{C}_t$. We use $\gamma_1(t)$ or $d_t$ to denote the corresponding adjacency matrix along with the diffusion process of the coordinates. Under some specific design of $\mathbf{f}$ and $\mathbf{G}$ (see Appendix~\ref{app: Fokker-Planck equation}), the forward diffusion process of the adjacency matrix is given by
\begin{equation}
    \mathrm{d} \gamma_1(t) = f(\gamma_1(t)) \mathrm{d} t + g(t) \mathrm{d} \mathcal{B}_t^{M}, \label{eq: method part adjacency matrix diffusion}
\end{equation}
for some $f$ and $g$ which are dependent on $\mathbf{f}$ and $\mathbf{G}$, and $\mathcal{B}_t^{M}$ is the Brownian motion on the adjacen matrix manifold $M$. Eq.~\eqref{eq: method part adjacency matrix diffusion} is a diffusion process in the manifold of the adjacency matrix. At each time of the diffusion, $\gamma_1(t)$ is always a feasible adjacency matrix. If we discretize the reverse diffusion process of Eq.~\eqref{eq: method part adjacency matrix diffusion} in $N$ steps, then the reverse iterations are given by~\citep{de2022riemannian}
\begin{equation}
    d^{k-1} = \pi_{M, d^{k}} \left( d^{k} - F^{\mathcal{N}}(d^k, kT/N) \right),
\end{equation}
where 
\begin{equation}
    \begin{aligned}
        F^{\mathcal{N}}(d_t, t) &= \tau \left(f(d_t) - g^2(t) \nabla_{d} \log p_{t} \left( d_t \right) \right) \\
        & \quad \quad  + \sqrt{\tau} g(t) \mathbf{z}^{t, d_t},
    \end{aligned}
\end{equation}
where $\tau = T / N$, $\mathbf{z}^{t, d_t}$ is a standard normal random variable in the tangent space of $d_t$ and injected into $d_t$, and $\pi_{M, d^k}$ is the project operator whose output is a feasible adjacency matrix. $p_t(d_t) = q_t(\mathcal{C}_t)$ and $\nabla_{d} \log p_{t} \left( d_t \right)$ is the score function of the adjacency matrix which is hard to compute or approximate. Following~\citet{zhou2023molecular}, we assume that the distribution of distance change is a combination of Gaussian and Maxwell-Boltzmann distribution (see Eq.~\eqref{eq: MB distance distribution}). It is non-trivial to explicitly compute the formula for the projection operator $\pi_{M, d_k}$. We instead consider the corresponding reverse diffuion of the coordinates. Let $\varphi$ denote a smooth function that maps a feasible adjacency matrix to one of its coordinates. Such a mapping defines a manifold called the spectral coordinate manifold and we denote it as $N$. Note that each set of coordinates in the spectral coordinate manifold $N$ corresponds to a unique adjacency matrix and vice versa. Then we have 
\begin{equation}
    \mathcal{C}^{k-1} = \varphi \circ \pi_{M, d^{k}} \left( d^{k} - F^{\mathcal{N}}(d^k, kT/N) \right) \in N. \label{eq: reverse C in form pi}
\end{equation}

To sample from the above iterations, we need to compute the formula for $\varphi \circ \pi_{M, d^{k}}$ which can be considered as a transformation from a non-necessary feasible adjacency matrix to a set of coordinates in manifold $N$. Finding the exact solution for $\varphi \circ \pi_{M, d^{k}}$ is also challenging. In the following, we will approximate the formula. Note that $\varphi \circ \pi_{M, d^{k}}$ is a function that maps a non-necessarily adjacency matrix to a set of coordinates whose adjacency matrix is closest to the function's input. Let $\hat{d} = d^k + \sum_{i < j} \alpha_{ij} \bar{e}_{ij}$ and $\bar{e}_{ij}$ is a zero matrix except for entries $(i, j)$ and $(j, i)$ being 1. We can define $\varphi \circ \pi_{M, d^{k}}$ as an optimization problem as 
\begin{equation}
\resizebox{0.85\hsize}{!}{$\varphi \circ \pi_{M, \tilde{d}} (\hat{d}) \triangleq \underset{\mathcal{C} = [\mathbf{x}_1, \ldots, \mathbf{x}_n] \in N}{\operatorname{argmin}} \sum_{i<j} \left( \left\| \mathbf{x}_i - \mathbf{x}_j \right\| - \hat{d}_{ij} \right)^2.$}
\end{equation}

Note that in the sampling process, we do not need to enforce the sampled coordinates in the spectral coordinate manifold $N$. Hence, we can ignore the constraint $\mathcal{C} \in N$ in the optimization problem at the implementation level. Hence, the optimization problem becomes
\begin{equation}
\resizebox{0.85\hsize}{!}{$\varphi \circ \pi_{M, \tilde{d}} (\hat{d}) \triangleq \underset{\mathcal{C} = [\mathbf{x}_1, \ldots, \mathbf{x}_n]}{\operatorname{argmin}} \sum_{i<j} \left( \left\| \mathbf{x}_i - \mathbf{x}_j \right\| - \hat{d}_{ij} \right)^2.$}
\end{equation}

Compared with $d^{k}$, this adjacency matrix $\hat{d}$ has a small perturbation on each entry. We consider a simple case where $\hat{d} = d^k + \delta \bar{e}_{uv}$. We want to transform such a change of the adjacency matrix to the coordinate shift. We approximate that if the $(u, v)$ and $(v, u)$ entries of the adjacency matrix $d^k$ increase by $\delta$, then node $u$ and node $v$ go in their opposite directions by $\frac{\delta}{2 (n - 1)}$ where $n$ is the number of nodes. The term $n-1$ can be considered as the other $n - 1$ nodes are rebelling against the extending of the edge $uv$. Such an approximation has a bounded error and does not explode as $n$ increases. For a more detailed analysis of such an approximation, we refer readers to Appendix~\ref{sec: approximation of differential} and Appendix~\ref{app: scatter-mean approx error bound} and we summarize the error bound in the following theorem.

\begin{theorem}
    Consider the optimization problem $f(\hat{d}) \triangleq \underset{\hat{\mathcal{C}} = [\hat{\mathbf{x}}_1, \ldots, \hat{\mathbf{x}}_M]} \min \sum_{i<j} \left( \left\| \hat{\mathbf{x}}_i - \hat{\mathbf{x}}_j \right\| - \hat{d}_{ij} \right)^2$, where $\hat{d} = \tilde{d} + \delta \bar{e}_{uv}$. The optimal value $f(\hat{d})$ approximated by $\hat{\mathcal{C}} = \tilde{\mathcal{C}} + \frac{\delta}{2 (n-1)} \frac{\partial \tilde{d}_{uv}}{\partial \tilde{\mathcal{C}}}$ where the adjacency matrix of $\tilde{\mathcal{C}}$ is $\tilde{d}$, is bounded by $\frac{2n^2 - 7n + 6}{2 (n-1)^2} \delta^2$. \label{thm: error bound}
\end{theorem}

We can extend the approximation results of simple case $\hat{d} = d^k + \delta \bar{e}_{uv}$ to the generic case $\hat{d} = d^k + \sum_{i < j} \alpha_{ij} \bar{e}_{ij}$ by the linearity of the differential (see Appendix~\ref{app: generic case approximation}). Mathematically, we use $\simeq$ to denote the left and right-hand sides of the equation having the same adjacency matrix. Then Eq.~\eqref{eq: reverse C in form pi} becomes 
\begin{equation}
    \mathcal{C}^{k-1} \simeq \mathcal{C}^{k} - \frac{F^{\mathcal{N}}(d^{k}, kT / N)}{2 (n-1)} \circledast \frac{\partial d^{k}}{\partial \mathcal{C}^{k}}, \label{eq: method discrete c inverse by d}
\end{equation}
where $n$ is the number of nodes, $F^{\mathcal{N}}(d^{k}, kT / N) \circledast \frac{\partial d^{k}}{\partial \mathcal{C}^{k}} = \frac{1}{2} \sum_{i, j = 1, \ldots, n} F_{ij}^{\mathcal{N}}(d^{k}, kT / N) \frac{\partial d_{ij}^k}{\partial \mathcal{C}^k}$. For the detailed mathematics derivation, we refer readers to Appendix~\ref{sec: manifold definitions} to Appendix~\ref{sec: approximation of differential}. Although there may exist advanced but iterative algorithms for distance geometry~\citep{more1999distance, chambolle2019total} that may complex the gradient, we believe our simple but bounded approximated solution with linear operation can ease the gradient propagation in the training process and accelerate the sampling speed.

In Eq.~\eqref{eq: method discrete c inverse by d}, there are two unknown terms $f$ and $g$ in the function $F^{\mathcal{N}}$ depending on the choice of $\mathbf{f}$ and $\mathbf{G}$, which we want to eliminate as much as possible. By inspecting the discretization of forward SDE of coordinates and adjacency matrices, we find that Eq.~\eqref{eq: method discrete c inverse by d} corresponds to the SDE 
\begin{equation}
    \label{eq: reverse SDE in c}
    \begin{aligned}
        \mathrm{d} \mathcal{C}_t &= \mathbf{G}(\mathcal{C}_t, t) \mathrm{d} \mathbf{w}_t + \Bigl( \mathbf{f}(\mathcal{C}_t, t) \\
        & \quad \quad - \frac{g^2(t) }{2 (n-1)} \nabla_{d} \log p_{t} ( d_t ) \circledast \frac{\partial d_t}{\partial \mathcal{C}_t} \Bigr) \mathrm{d} t.
    \end{aligned}
\end{equation}
A detailed derivation of Eq.~\eqref{eq: reverse SDE in c} can be found in Appendix~\ref{sec: Elimination of functions related to the adjacency matrix}. 

\begin{remark}[Necessity of alternating between coordinates and distances]
    Alternating between spectral coordinates and distances within the analytical framework is imperative. This necessitates the conversion of coordinates to distances, wherein the model is deployed to equip with the SE(3)-invariance. Conversely, transitioning from distances to coordinates through $\varphi \circ \pi_M$ mimics the projection of a tangent vector onto the SE(3)-invariant manifold. Therefore, the interplay between these two distinct manifolds becomes indispensable for a comprehensive understanding of the underlying dynamics.
\end{remark}

\subsection{Reverse ODE of coordinates via adjacency matrices}
Deriving the general reverse ODE for arbitrary forward diffusion on coordinates is challenging. In this paper, we only consider the variance exploding diffusion ($\mathbf{f} \equiv \mathbf{0}$)~\citep{song2019generative, song2020score} and leave broader cases to our future work. To derive the reverse ODE, one may consider the Fokker-Planck equation for adjacency matrices. Given a set of coordinates $\mathcal{C} = [\mathcal{C}_1, \ldots, \mathcal{C}_n]^\top \in \mathbb{R}^{n \times 3}$, we inject i.i.d. Gaussian noise to each node coordinate $\mathcal{C}_i$ and consider the change of distance between node $u$ and $v$. We found that if $\mathcal{C}_i^{t + \tau} = \mathcal{C}_i^t + \sqrt{\tau} \mathbf{z}_i^t$, $\mathbf{z}_i^t \sim \mathcal{N}(\mathbf{0}, \sigma^2 \mathbf{I})$, then $d_{uv}^{t + \tau}$ has the distribution of
\begin{equation}\label{eq:d_uv}
    d_{uv}^{t + \tau} = d_{uv}^t + \frac{2 \sqrt{\tau} \mathbf{z}^\top (\mathcal{C}_u - \mathcal{C}_v) + \tau \|\mathbf{z}\|^2}{\| \sqrt{\tau} \mathbf{z} + \mathcal{C}_u - \mathcal{C}_v\| + \| \mathcal{C}_u - \mathcal{C}_v\|},
\end{equation}
where $\mathbf{z} \sim \mathcal{N}(\mathbf{0}, 2\sigma^2 \mathbf{I})$. Utilizing the above equality, if $d_{uv}^{t + \tau}$ and $d_{uv}^{t}$ are obtained from Eq.~\eqref{eq: method part coordinate diffusion} and Eq.~\eqref{eq: method part adjacency matrix diffusion}, we have
\begin{subequations}
   \begin{align}
        &\mathbb{E} \left[ d_{uv}^t -  d_{uv}^{t + \tau} \right] \\
        &\qquad = \sqrt{\tau \pi} \mathbf{G}(t) L^{1/2}_{1/2} \left( \frac{- \| \mathcal{C}_u - \mathcal{C}_v \|^2}{2 \sqrt{2\tau} \mathbf{G}(t)} \right) \\
        &\qquad \approx - 3 \tau \mathbf{G}^2(t) / d_{uv}^t + \lito(\tau), \\
        &\mathbb{E} \left[ \left( d_{uv}^t - d_{uv}^{t + \tau} \right)^2 \right] = 2 \tau \mathbf{G}^2(t) + \lito(\tau),
    \end{align} 
\end{subequations}

where $L$ is the generalized Laguerre polynomial~\citep{sonine1880recherches}. The above results play a pivotal role in formulating the reverse ODE of both adjacency matrices and coordinates. Unlike the diffusion process in Euclidean space, in the form of the expectation of the distance change, a term denoted as $d_{uv}^t$ emerges in the denominator. This implies that the magnitude of the distribution shift is contingent upon not only the selection of $\mathbf{G}(t)$ but also the present state of the adjacency matrix. We find that if we ignore the entry independence of the adjacency matrix and define the matrix division as an entry-wise matrix division, the reverse ODE of the adjacency matrix and coordinates are respectively 

\par
{
\small
\vspace{-1em}
\begin{align}
    &\mathrm{d} \gamma_1(t) = \left( 3 \frac{\mathbf{G}^2(t)}{\gamma_1(t)} - \mathbf{G}^2(t) \nabla_{\gamma_1} \log p_t \left( \gamma_1(t) \right) \right) \mathrm{d} t , \\
    &\frac{\mathrm{d} \mathcal{C}_t}{\mathrm{d} t} = \frac{\mathbf{G}^2(t)}{2 \left( n - 1 \right)} \left( \frac{3}{d_t} - \nabla_{d} \log p_t \left( d_t \right) \right) \circledast \frac{\partial d_t}{\partial \mathcal{C}_t}. \label{eq: reverse ODE in c}
\end{align}
}

A detailed analysis from Eq.~\eqref{eq:d_uv} to Eq.~\eqref{eq: reverse ODE in c} can be found in Appendix~\ref{sec: reverse ODE}. 

\subsection{Transformation for sparse graph}
To sample from reverse diffusion, we need to train a model $\mathbf{s}_{\boldsymbol{\theta}}$ to predict $\nabla_{d} \log p_t \left( d_t \right)$. Note that for a graph with $n$ nodes, $\nabla_{d} \log p_t \left( d_t \right)$ is an $n$ by $n$ matrix and usually neural networks are designed for sparse graphs. Hence, we only consider predicting a portion of entries in the score function of distance. Note that for arbitrary graph topology, if the minimum edge degree is no less than 4, then the edge length set is bijection to its coordinate set under SE(3)-equivalence. In this case, analysis of mappings from the adjacency matrix change into the coordinate shift still holds and 
\begin{equation}
\resizebox{.42\textwidth}{!}{
    $\nabla_{d} \log p_t \left( d_t \right) \circledast \frac{\partial d_t}{\partial \mathcal{C}_t} \approx \sum_{i = 1}^{n} \frac{1/2}{\operatorname{degree}_i} \underset{j \in N(i)}{\sum} \frac{\partial d_{ij}^{t}}{\partial \mathcal{C}_t} \nabla_{d_{ij}^{t}} \log p_{t} (d_t), \label{eq: partially connected scatter-mean}$
}
\end{equation}

where $N(i)$ denotes the neighborhood of node $i$ and $\operatorname{degree}_i$ represents the degree of node $i$. Under the methodology in~\citet{xu2022geodiff}, we utilize a Message Passing Neural Network (MPNN)~\citep{gilmer2017neural}. The MPNN framework is employed to compute node embeddings $\boldsymbol{h}_v^{(t)}$ for all nodes $v \in V$ within the input graph $G$. This computation is performed through a series of $M$ layers of iterative message passing:
\begin{equation}
    \boldsymbol{h}_u^{(m+1)} = \psi \left(\boldsymbol{h}_u^{(m)}, \sum_{v\in N_u }  \boldsymbol{h}_v^{(m)}\cdot \phi (\boldsymbol{e}_{uv}, d_{uv}) \right),
\end{equation}
and the loss function is set to

{\small
\vspace{-2em}
\begin{align}
    &\mathcal{L}\left(\boldsymbol{\theta} ;\left\{\sigma_i\right\}_{t=1}^T\right)  \triangleq \frac{1}{T} \sum_{t=1}^T \frac{\sigma_t}{2} \mathbb{E}_{d_0, d_t} \left[ l \left( \boldsymbol{\theta}, d_t, d_0 \right) \right], \label{eq: score matching loss} \\
    &l \left( \boldsymbol{\theta}, d_t, d_0 \right)  \triangleq \| \mathbf{s}_{\boldsymbol{\theta}}(d_t, t) - \nabla_{d} \log p_t \left( d_t \mid d_0 \right) \circledast \frac{\partial d_t}{\partial \mathcal{C}_t} \|_2^2,
\end{align}
}%

where $\mathbf{G}^2(t) = \frac{\mathrm{d} \sigma_t^2}{\mathrm{d} t}$. Following ~\citet{zhou2023molecular}, we assume the distribution of distance change follows
\begin{equation}
    \nabla_{d} \log p_t \left( d_t \mid d_0 \right) = \frac{a_t}{d_t}-\frac{d_t - d_0}{2\sigma_t^2}, \label{eq: MB distance distribution}
\end{equation}
where $a_t$ is a hyperparameter introduced to combine the Maxwell-Boltzemann and Gaussian distribution. When Eq.~\eqref{eq: score matching loss} is minimized, the obtained model $\mathbf{s}_{\boldsymbol{\theta}}(d_t, t) \approx \nabla_{d} \log p_t (d_t) \circledast \frac{\partial d_t}{\partial \mathcal{C}_t}$. However, since the pair-wise distances $d_t$ (a random variable) appear in the denominator, this may lead to an unstable training process. Inspecting the reverse ODE given by Eq.~\eqref{eq: reverse ODE in c}, we note that there is also a $d_t$ in the denominator. Hence, to eliminate the appearance of random variables in the denominator, we revise the score-matching objective to (which is also the hypothesis in~\citet{xu2022geodiff})
\begin{equation}
    \nabla_{d} \log p_t \left( d_t \mid d_0 \right) = -\frac{d_t - d_0}{2 \sigma_t^2}, \label{eq: approximate distance distribution}
\end{equation}
and we approximate the reverse ODE to be
\begin{equation}
    \frac{\mathrm{d} \mathcal{C}_t}{\mathrm{d} t} = - \frac{\mathbf{G}^2(t)}{2 \left( n - 1 \right)} \nabla_{d} \log p_t \left( d_t \right) \circledast \frac{\partial d_t}{\partial \mathcal{C}_t}. \label{eq: naive ODE in c}
\end{equation}

\subsection{Reverse process correction}
In the proposed reverse diffusion process of both SDE and ODE, three approximations are made. Firstly, a linear operation is utilized to approximate the solution of the projection from the tangent space of the adjacency matrix to the manifold. Secondly, the dependence of the entries in the adjacency matrix is ignored. Thirdly, the score function of distance distribution is assumed to be the Gaussian one.
To rectify these approximation errors, an additional value referred to as the ``corrector'' is introduced to the reverse SDE/ODE. The rationale behind this modification is that during the denoising process from $\mathcal{C}_s$ from $q_s \left( \mathcal{C}_s \right)$ to $q_t \left( \mathcal{C}_t \right)$, the obtained sample $\mathcal{C}_t$ after one step does not accurately align with the distribution of $q_t \left( \mathcal{C}_t \right)$ due to the presence of approximation errors. To mitigate these errors, it is desired to push $\mathbf{x}$ to take a larger step size. Therefore, the corrector is incorporated to amplify the step size. A more comprehensive analysis and discussion of this approach can be found in Appendix~\ref{sec: corrector analysis}.

\begin{table*}[tb]
\begin{center}
\begin{sc}
\caption{Comparison results between LD and our proposed SDE and ODE on QM9 and Drugs datasets under different sampling steps. Higher values for COV indicate that the model can generate samples with higher diversity. Lower values for MAT imply the model can generate high-quality samples with lower differences compared with the ground truth samples.}
\label{tab: GEOM}
\begin{tabular}{cl|cccc|cccc}
\toprule
    \multirow{2}{*}{Datasets} & \multirow{2}{*}{\makecell{Methods\\/\# Steps}} & \multicolumn{2}{c}{COV-R(\%) $\uparrow$} & \multicolumn{2}{c|}{MAT-R($\SI{}{\angstrom}$) $\downarrow$} & \multicolumn{2}{c}{COV-P(\%) $\uparrow$}& \multicolumn{2}{c}{MAT-P($\SI{}{\angstrom}$) $\downarrow$} \\ 
    & & Mean & Median & Mean & Median & Mean & Median & Mean & Median \\ 
    \midrule

    \multirow{7}{*}{QM9} & LD \kern 0.16667em \; 5000 & 90.07 & 93.39 & 0.2090 & 0.1988 & 52.79 & 50.29 & 0.4448 & 0.4267 \\
    & LD \kern 0.16667em \; 100 & 88.07 & 90.20 & 0.2711 & 0.2808 & 54.80 & 52.42 & 0.5090 & 0.4552 \\ 

    \cmidrule{2-10}

    & SDE \kern 0.16667em 1000 & 89.25 & 94.02 & 0.2215 & 0.2180 & 55.15& 52.42 & 0.4323 & 0.4164 \\ 
        
    & SDE \kern 0.16667em 100 & 88.34 & 91.59 & 0.2763 & 0.2757 & 53.48 & 51.25 & 0.4870 & 0.4564 \\ 

    & ODE 1000 & 90.24  & 95.04  & 0.2052  & 0.1956  &  53.76 & 52.26  & 0.4304  & 0.4251 \\ 
    & ODE 100 & 90.81 & 96.56 & 0.2691 & 0.2765 & 48.73 & 47.16 & 0.5190 & 0.5026 \\ 

    \midrule
    \midrule

    \multirow{7}{*}{Drugs} & LD \kern 0.16667em \; 5000 & 89.13 & 97.88 & 0.8629 & 0.8529 & 61.47 & 64.55 & 1.1712 & 1.1232 \\
    & LD \kern 0.16667em  \; 1000 & 82.92 & 96.29 & 0.9525 & 0.9334 & 48.27 & 46.03 & 1.3205 & 1.2724 \\ 
    
    \cmidrule{2-10}
    
    & SDE \kern 0.16667em 1000 & 83.39 & 91.02 & 0.9728 & 0.9784 & 58.43 & 60.28 & 1.2393 & 1.1814 \\ 
        
    & SDE \kern 0.16667em 100 & 86.44 & 93.08 & 0.9261 & 0.9132 & 64.53 & 67.83 & 1.1584 & 1.1279 \\ 

    & ODE 1000 & 90.64 & 98.85 & 0.8493  & 0.8270  & 59.67  & 62.12  & 1.1915  &  1.1420 \\ 
    & ODE 100 & 87.76 & 98.59 & 0.9105 & 0.8846 & 53.32 & 54.27 & 1.8528 & 1.3563 \\ 

    \bottomrule

\end{tabular}
\end{sc}
\end{center}
\end{table*}

\section{Experiments}
In this section, we empirically evaluate our proposed SDE (Eq.~\eqref{eq: reverse SDE in c}) and ODE (Eq.~\eqref{eq: naive ODE in c}) for generating 3D coordinates, focusing on two tasks: molecular conformation generation and human pose generation. Initially, we outline the general baseline sampling methods, (annealed) Langevin dynamics (LD), followed by a detailed description of each task. Additional experimental settings and results are available in Appendix~\ref{app: experiment extension}.
\subsection{Baseline}
We compare our sampling approaches with the standard sampling method via LD~\citep{song2019generative}:
\begin{equation}
    \mathcal{C}_{t-1} = \mathcal{C}_{t} + \alpha_t \nabla_{\mathcal{C}} \log q_t (\mathcal{C}_t) + \sqrt{2\alpha_t} \mathbf{z}_{t}^{\mathcal{C}_{t}},
    \label{eq: LD sample}
\end{equation}
for $t=N, \ldots, 1$, where each entry of $\mathbf{z}_{t}^{\mathcal{C}_{t}}$ adheres to a standard normal distribution, $\alpha_t = \delta \sigma_t^2$ and $\delta$ signifies the hyper-parameter for step size that needs to be selected by grid search. In this section, we present the optimal results derived from experimenting with various step sizes, with comprehensive details on alternative step sizes available in Appendix~\ref{app: hyperparameter of molecule} and Appendix~\ref{app: hyperparameter of humanpose}. The term $\nabla_{\mathcal{C}} \log q_t (\mathcal{C}_t)$ is approximated by $\nabla_{d} \log p_t \left( d_t \right) \circledast \frac{\partial d_t}{\partial \mathcal{C}_t}$, and an in-depth explanation can be found in~\citet{xu2022geodiff}. 

Modifying the step size $\delta$ allows for generation in fewer steps via LD, providing a conventional and equitable basis for comparison with our method. We assess both LD and our methods using the same trained model, trained over 5000 steps, with varied step counts in the sampling phase. Details on training and the backbone model are available in Appendix~\ref{app: training details}.

\subsection{Molecular conformation generation}

\paragraph{Datasets.} 
We employ two datasets, namely GEOM-Drugs and GEOM-QM9~\citep{axelrod2022geom} to validate the performance of our sampling method. The dataset split is the same as~\citet{xu2020learning}. The GEOM-QM9 test set contains 24,143 conformers originating from 200 molecules, averaging about 20 conformers per molecule. In the test set of the GEOM-Drugs dataset, we encounter 14,324 conformers from 200 molecules, with an average of approximately 47 atoms per molecule. In line with the previous work~\citep{xu2020learning}, we expand the generation of ground truth conformers to double their original quantity, resulting in more than 20k+ and 40k+ generated conformers.

\paragraph{Evaluation.}
We adopt established evaluation metrics, namely Coverage (COV) and Matching (MAT), incorporating Recall (R) and Precision (P) aspects to assess the performance of sampling methods~\citep{xu2020learning, ganea2021geomol}. COV quantifies the proportion of ground truth conformers effectively matched by generated conformers, gauging diversity in the generated set. Meanwhile, MAT measures the disparity between ground truth and generated conformers, complementing quality assessment. Furthermore, refined metrics COV-R and MAT-R place added emphasis on the comprehensiveness of the ground truth coverage, while COV-P and MAT-P are employed to gauge the precision and accuracy of the generated conformers. More details can be found in Appendix~\ref{app: geom experiment setting}.

\paragraph{Comparison with LD method.}
We compare the performance of our proposed SDE and ODE sampling methods against LD sampling across various steps, with the results detailed in Tab.~\ref{tab: GEOM}. The evaluation encompasses eight metrics, and the displayed results are obtained under hyper-parameter settings that ensure a well-balanced comparison among these evaluation criteria. Additional results under varying hyperparameter conditions are available in Appendix~\ref{app: hyperparameter of molecule}. Overall, our proposed methods can consistently achieve slightly better results than LD. These equivalent outcomes underscore the validity of our SDE and ODE sampling methods. In practice, the LD method's path efficiency could be enhanced by modifying the step size. However, our SDE and ODE models are grounded in systematic mathematical reasoning within the SE(3)-invariant space, providing a robust theoretical basis.
For qualitative analysis, Appendix~\ref{app: visualizaiton of molecule} showcases visualizations of the molecules generated by our SDE and ODE sampling methods. These results affirm that our approach efficiently and effectively produces meaningful 3D molecular geometries, thereby validating the effectiveness of our method.

\subsection{Human pose generation}
\begin{figure*}[tb]
    \centering
    \begin{tabular}{ *8{>{\centering\arraybackslash}m{1.7cm}}}
         (a) LD &\includegraphics[width=.1\textwidth, trim={50 50 50 50, clip}] {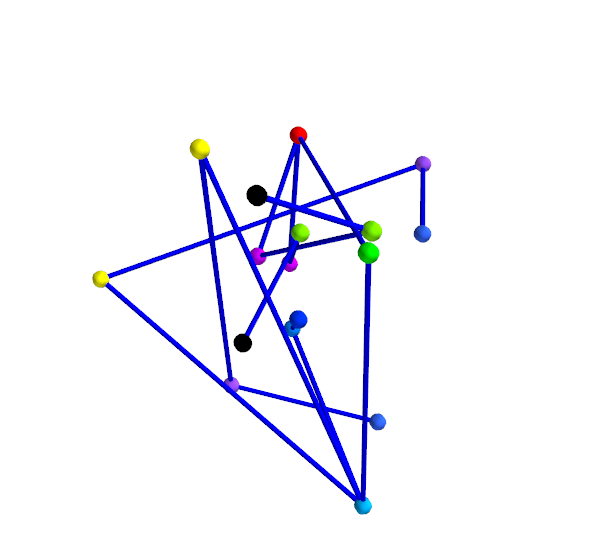}
             &\includegraphics[width=.1\textwidth, trim={50 50 50 50, clip}] {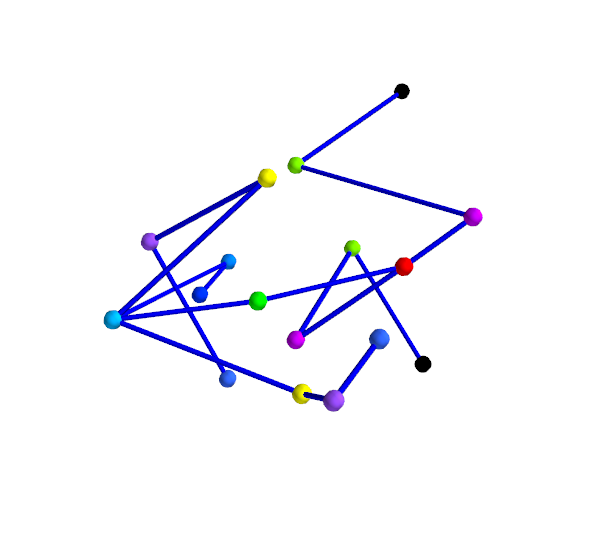}
             &\includegraphics[width=.1\textwidth, trim={50 50 50 50, clip}] {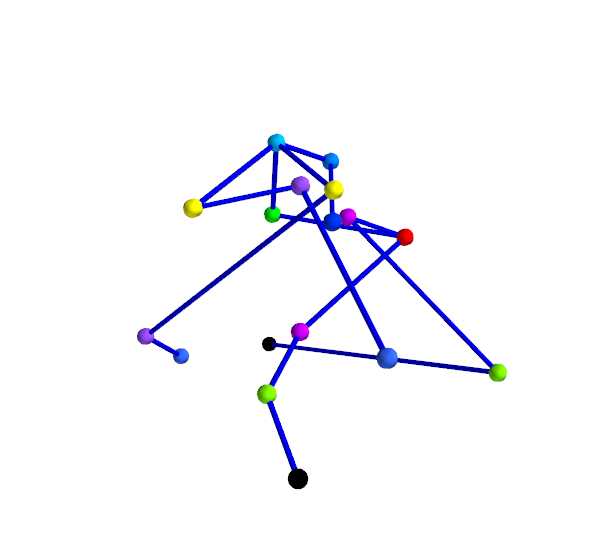}
             &\includegraphics[width=.1\textwidth, trim={50 50 50 50, clip}] {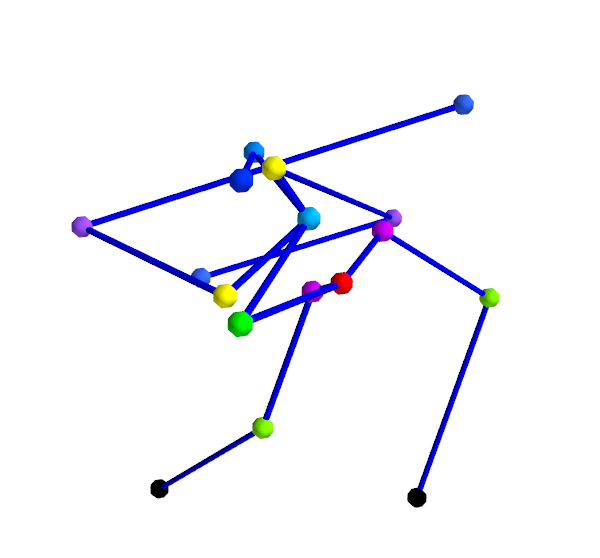}
             &\includegraphics[width=.1\textwidth, trim={50 50 50 50, clip}] {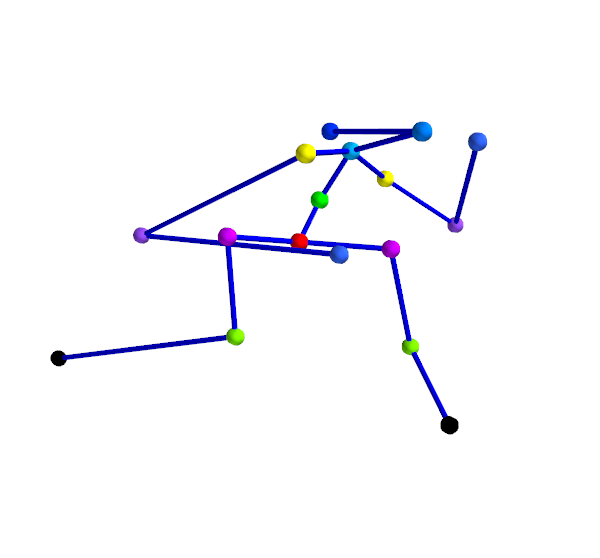}
             &\includegraphics[width=.1\textwidth, trim={50 50 50 50, clip}] {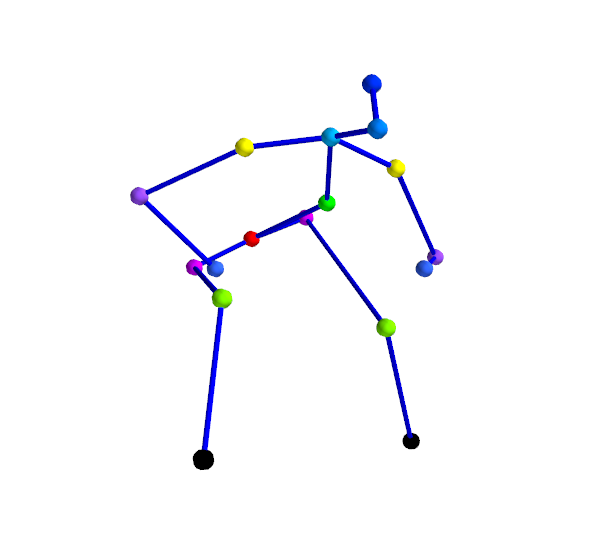}
             &\includegraphics[width=.1\textwidth, trim={50 50 50 50, clip}] {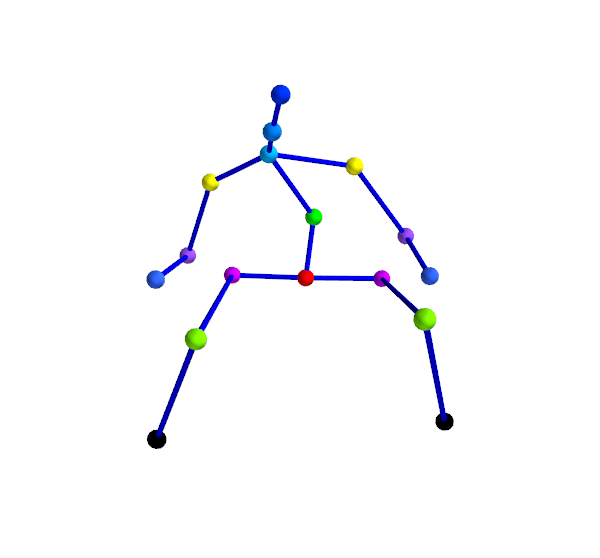}\\
        (b) Ours-SDE &\includegraphics[width=.1\textwidth, trim={50 50 50 50, clip}] {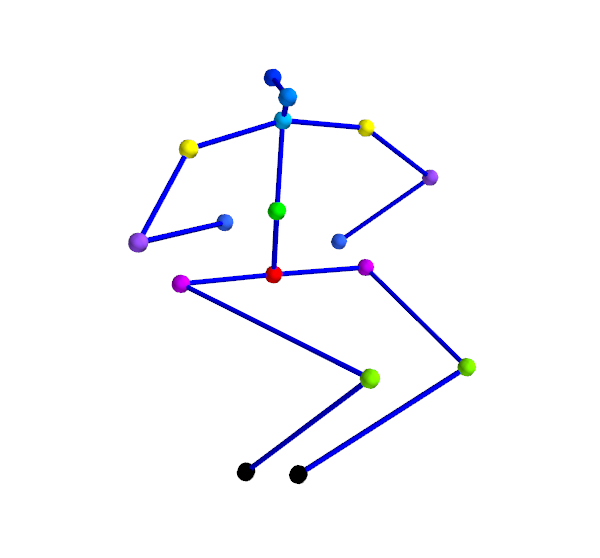}
             &\includegraphics[width=.1\textwidth, trim={50 20 50 50, clip}] {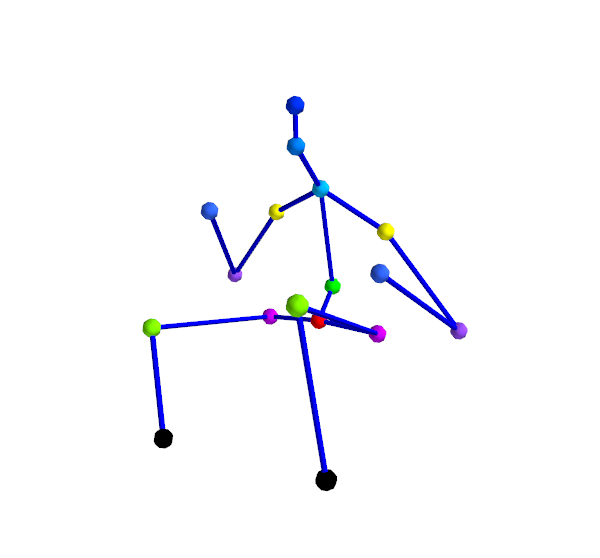}
             &\includegraphics[width=.1\textwidth, trim={50 20 50 50, clip}] {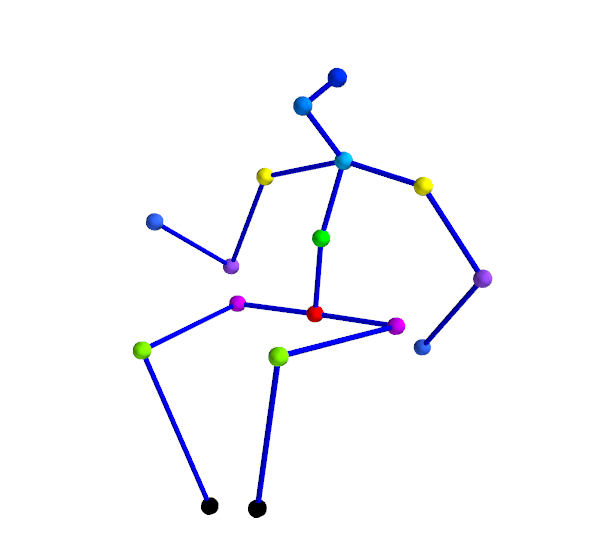}
             &\includegraphics[width=.1\textwidth, trim={50 20 50 50, clip}] {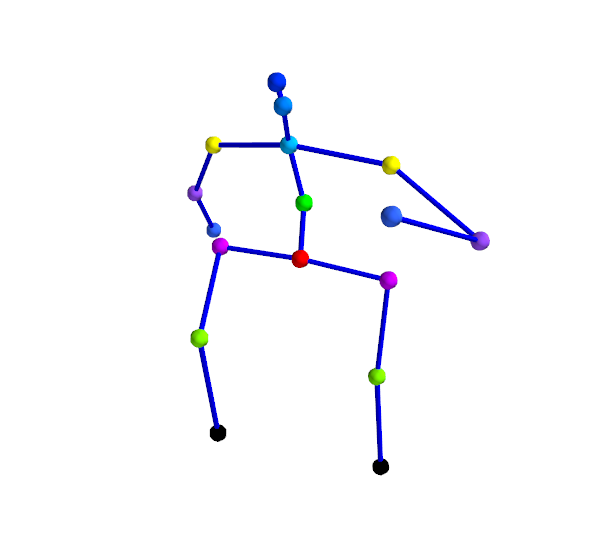}
             &\includegraphics[width=.1\textwidth, trim={50 20 50 50, clip}] {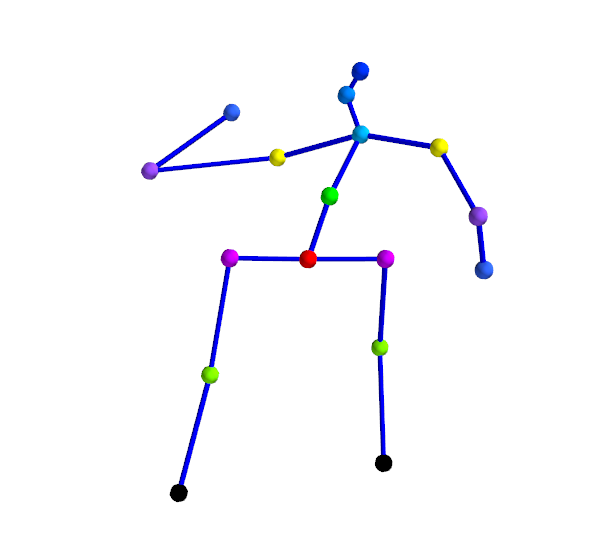}
             &\includegraphics[width=.1\textwidth, trim={50 20 50 50, clip}] {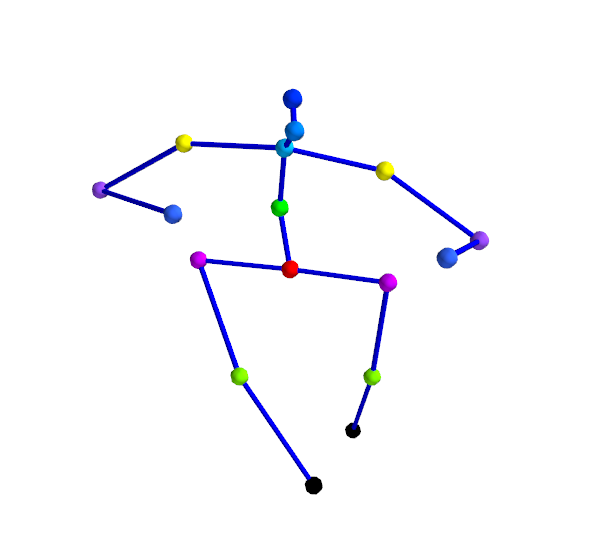}
             &\includegraphics[width=.1\textwidth, trim={50 20 50 50, clip}] {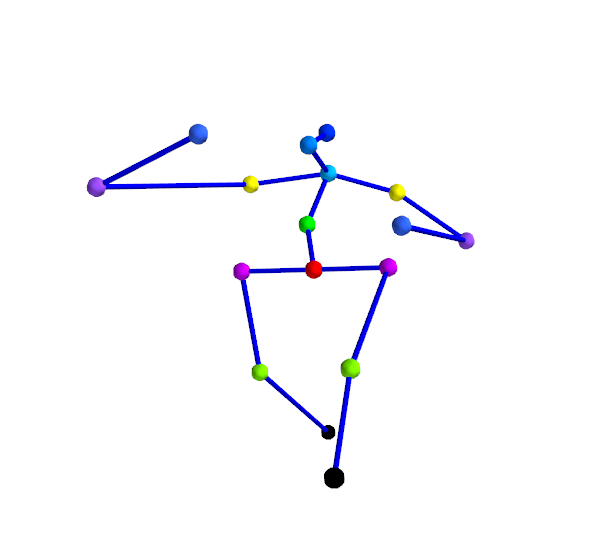}\\
        (c) Ours-ODE &\includegraphics[width=.1\textwidth, trim={50 50 50 50, clip}] {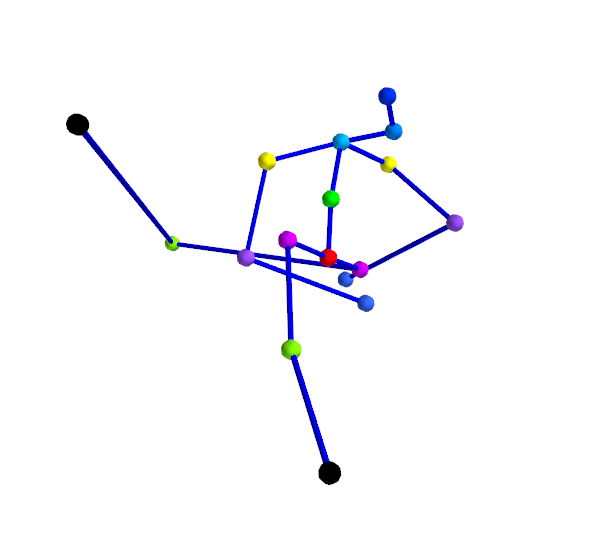}
             &\includegraphics[width=.1\textwidth, trim={50 50 50 50, clip}] {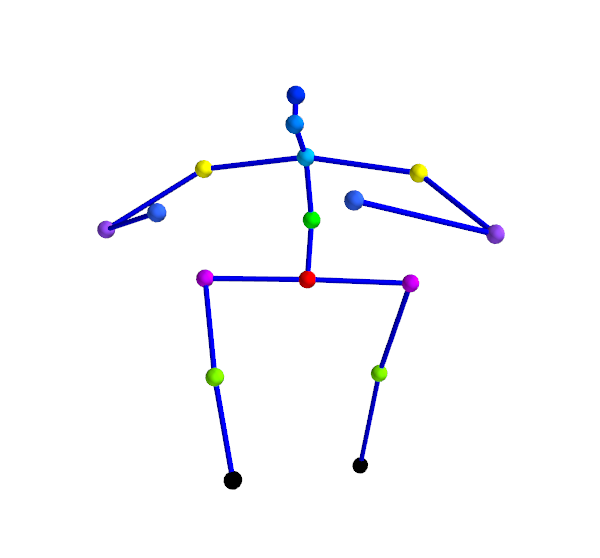}
             &\includegraphics[width=.1\textwidth, trim={50 50 50 50, clip}] {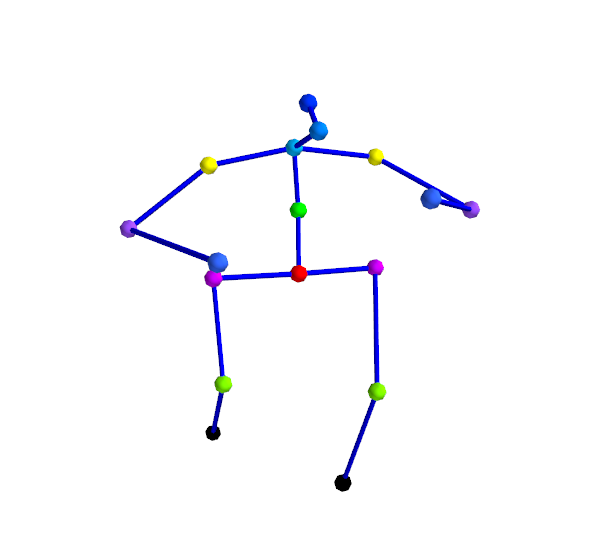}
             &\includegraphics[width=.1\textwidth, trim={50 50 50 50, clip}] {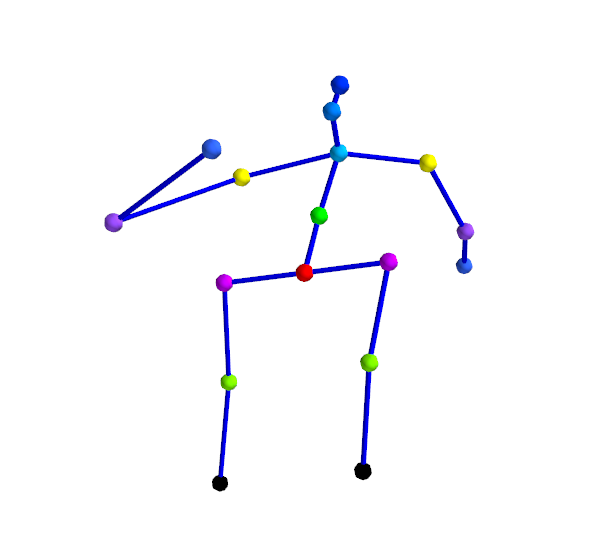}
             &\includegraphics[width=.1\textwidth, trim={50 50 50 50, clip}] {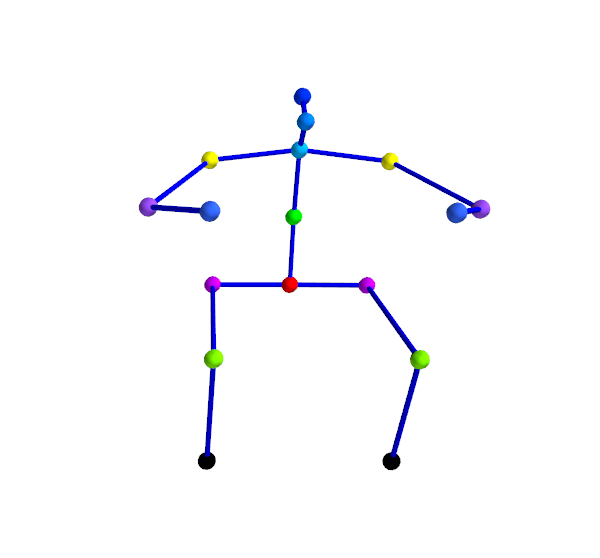}
             &\includegraphics[width=.1\textwidth, trim={50 50 50 50, clip}] {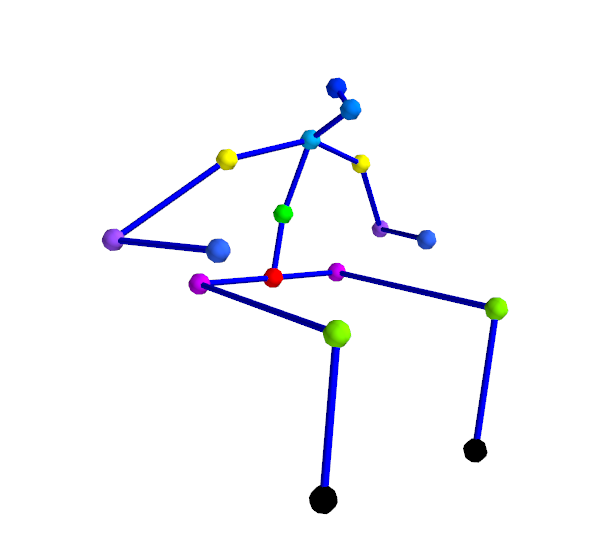}
             &\includegraphics[width=.1\textwidth, trim={50 50 50 50, clip}] {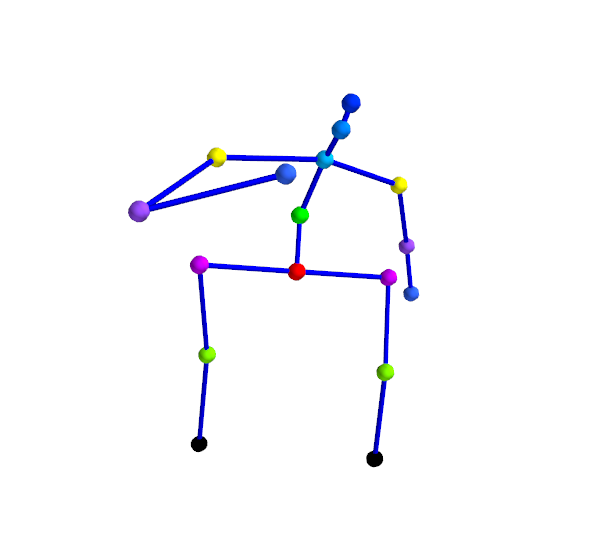}\\
             & steps=25& steps=50& steps=100& steps=250& steps=500& steps=1000& steps=2000
    \end{tabular}
    \caption{Visualization of sampling results obtained through LD and our proposed SDE/ODE with different sampling steps is presented. The human poses are labeled as ``eating''. Samples are obtained from the exactly same trained model but drawn from different sampling algorithms. We grid search the hyperparameters in each sampling algorithm for a fair comparison.}
    
\label{tab:humanpose_compare}
\end{figure*}

\paragraph{Datasets and pre-processing.} 
We assess our sampling methods using the Human3.6M dataset~\citep{ionescu2013human3}, acknowledged as the most extensive dataset for human motion analysis. This dataset portrays seven actors engaged in 15 diverse actions, including walking, eating, discussions, sitting, and phoning. Actors are delineated by a skeletal structure comprising 32 joints, from which 17 key points are extracted to represent a human pose. Adhering to the standard train-test evaluation protocol for Human3.6M dataset~\citep{iskakov2019learnable, kocabas2019self}, we employ subjects 1, 5, 6, 7, and 8 for training, while subjects 9 and 11 are utilized for validation, resulting in 312,188 and 109,867 poses for training and validation, respectively.

The human body can be conceptualized as a graph, with joints as nodes linked by edges symbolizing limbs. Each node is defined by its 3D coordinates. Our goal is to efficiently generate multiple plausible 3D human poses. We normalize the 3D coordinates and introduce ``virtual edges" to achieve comprehensive graph connectivity. Additional details on these processes are elaborated in Appendix~\ref{app: humanpose experiment}.

\paragraph{Evaluation.}  
Since human behavior is composed of many human poses, it is difficult to judge a specific behavior only from a single action. However, our purpose is to compare the rapidity and effectiveness of our sampling method. Thus we design a metric to evaluate the quality of the generated single action. Drawing inspiration from the minimum matching distance (MMD) employed in the evaluation of point cloud generation tasks and evaluation metrics in molecular conformation generation, we devise a novel evaluation approach known as Alignment Minimum Matching Distance (AD). This method not only captures the distribution variance between the generated samples and the comprehensive reference dataset but also gauges the quality of the generated sample itself. The formula for AD is defined as
\begin{equation}
    \operatorname{AD} \triangleq \frac{1}{|S_g|} \sum_{\mathcal{X} \in S_g} \min_{\mathcal{X}_r \in S_r} \| \operatorname{Align}(\mathcal{X}, \mathcal{X}_r) - \mathcal{X}_r) \|^2_F,
\end{equation}
where $S_g$ and $S_r$ denote generated and reference human pose, respectively. $\| \cdot \|_F$ denotes the Frobenius norm. The function $\operatorname{Align}$ is defined to minimize the Frobenius norm of $T(\mathcal{X}) - \mathcal{X}_r$ by translations and rotations.

\paragraph{Comparison with LD method.}
We apply our SDE/ODE and LD methods to generate 500 human poses labeled as eating, varying the steps from 25 to 2000. The quantitative results are presented in Table~\ref{tab: human result}, highlighting the best outcomes for each method under optimal hyper-parameter settings. For comprehensive details on hyperparameters and varied results, refer to Appendix~\ref{app: hyperparameter of humanpose}. We found that our proposed SDE/ODE achieves much better results with a substantially reduced number of iterations compared with LD. The effectiveness of LD diminishes significantly as the number of steps decreases, whereas our methods maintain performance consistency across varying step counts.
Qualitative comparisons in Fig.~\ref{tab:humanpose_compare} reveal that while LD requires hundreds of steps to produce reasonable samples, our SDE/ODE method accomplishes this with far fewer steps. Additional samples are visualized in Appendix~\ref{app: visualizaiton of humanpose}. The outcomes of the human pose generation task suggest that our method, anchored in comprehensive theoretical derivation within the SE(3)-invariant manifold, potentially demonstrates enhanced versatility and applicability.

\begin{table}[tb] 
    \centering 
    \caption{Comparision of mean AD scores of the three sampling methods under 100, 250, 500, and 1000 steps, where a lower AD value indicates better performance.}
    \label{tab: human result}
    \begin{sc}
        \begin{tabular}{cccc}
        \toprule
        \diagbox{\# Steps}{Method} & LD & SDE & ODE  \\ 
        \midrule 
        100   & 5.1739 & 2.3737  & 2.8978 \\ 
        250   & 4.3071 & 2.3452  & 2.5739 \\ 
        500   & 4.0020 & 2.3463  & 2.2811  \\ 
        1000  & 3.7369 & 2.3315  & 2.1620 \\ 
        
        \bottomrule
        \end{tabular}
    \end{sc}
    \vspace{-15pt}
\end{table}

\section{Conclusion}

This paper elucidates the complex diffusion processes in SE(3)-invariant spaces by examining the interaction between coordinates and adjacency matrix manifolds. We introduce noise to coordinates in the forward process and use differential geometry for reverse process analysis, focusing on inter-point distance changes. This leads to projection-free SDE and ODE sampling methods, improving efficiency as well as interpretability. Our empirical tests in molecular conformation and human pose generation confirm the high efficiency and quality of our 3D coordinate generation methods. This work offers significant insights into SE(3)-invariant manifolds and broad applicability in 3D coordinate generation tasks.

\section*{Broader impact}
This paper presents work whose goal is to advance the field of Machine Learning. There are many potential societal consequences of our work, none which we feel must be specifically highlighted here.

\bibliography{example_paper}
\bibliographystyle{icml2024}

\newpage
\appendix
\onecolumn

\section{Definition} \label{app: definition}

\begin{definition} \label{def: d function}
    We define $\adjj: \mathbb{R}^{n \times 3} \rightarrow \mathbb{R}_{+}^{n \times n}$ as the mapping from a set of coordinates of $n$ nodes to its adjacency matrix. Suppose $\mathcal{C} = [\mathbf{x}_1, \ldots, \mathbf{x}_n]$, then
    \begin{equation}
        \adjj_{ij}(\mathcal{C}) = \| \mathbf{x}_i - \mathbf{x}_j \|, \quad \forall i, j = 1, \ldots, n.
    \end{equation}
\end{definition}

\begin{definition} \label{def: feasible adjacency matrix}
    We say an adjacency matrix $A \in \mathbb{R}^{n \times n}$ is \emph{feasible} if there exists a set of non-planar $n$ coordinates $\mathcal{C}$ such that $\adjp{\mathcal{C}} = A$. Non-planar coordinates have exactly 3 positive eigenvalues (see Theorem~\ref{thm: mapping from adjacency matrix to coordinates}). If coordinates are planar (but not in a line), they only have two positive eigenvalues.
\end{definition}

\begin{definition} \label{def: paper matrix norm}
In this paper, the matrix norm $\| \cdot \|$ is defined to be the Frobenius norm, i.e, given $A \in \mathbb{R}^{m \times n}$, $\| A \| =  \| A \|_{\mathrm{F}} = \sqrt{\sum_{i=1}^m \sum_{j=1}^n a_{\mathrm{ij}}^2}$. 
\end{definition}

\begin{definition} \label{def: matrix2 times matrix4}
We define the weighted product of matrix $A \in \mathbb{R}^{n \times n}$ and matrix $B \in \mathbb{R}^{n \times n \times s \times t}$ as 
    \begin{equation}
        A \circledast B = \frac{1}{2} \sum_{i, j = 1, \ldots, n} A_{ij} B_{ij},
    \end{equation}
where $A_{ij} \in \mathbb{R}, B_{ij} \in \mathbb{R}^{s \times t}$ and $A \circledast B \in \mathbb{R}^{s \times t}$. 
\end{definition}

\begin{definition} \label{def: equi coordinate}
Given two set of coordinates $\mathcal{C}_1$ and $\mathcal{C}_2$, if $\adjp{\mathcal{C}_1} = \adjp{\mathcal{C}_2}$, we write $\mathcal{C}_1 \simeq \mathcal{C}_2$. Moreover, if two equations give the same marginal distribution density with respect to $q$ where $q$ is any SE(3)-invariant distribution density function, we refer such a marginal distribution identity as $\simeq$. 
\end{definition}

\section{Backgrounds}
\subsection{Introduction of differential geometry} \label{app: intro of differential geometry}
\begin{wrapfigure}{r}{.3\textwidth}
\vspace{-10pt}
\centering
\includegraphics[width=0.2\textwidth]{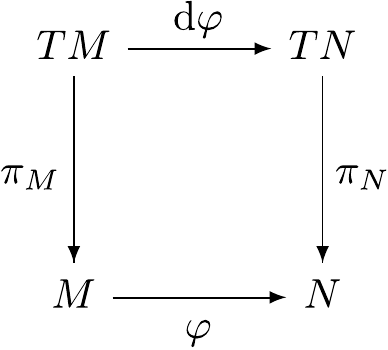}
\vspace{-10pt}
\caption{Mappings between two manifolds.}
\vspace{-20pt}
\end{wrapfigure}

In differential geometry, we consider mappings between two manifolds. Suppose that $\varphi: M \rightarrow N$ is a smooth map between smooth manifolds.
\paragraph{Tangent space $TM$.} In the following, we give an informal definition of the tangent space. The tangent space of manifold $M$ at $x \in M$, denoted by $T_{x} M$, is defined as the set of all tangent vectors $\gamma^{'}(0)$ at $x$, where $\gamma$ is an arbitrary curve in the manifold $M$ satisfying $\gamma(t): (-1, 1) \rightarrow M$ and $\gamma(0) = x$.

\paragraph{Tangent space projection $\pi_M$.} $\pi_M: TM \rightarrow M$ is the projection operator that maps a vector in the tangent space back to the manifold. The tangent space projection of $\hat{x} \in T_{x} M$ is defined by $\pi_{M, x} (\hat{x}) = \operatorname{argmin}_{y \in M} \| y - \hat{x} \|$ for some norm $\| \cdot \|$. That is, we find the point $y$ in the manifold $M$ that is closest to $\hat{x}$.

\paragraph{Differential $\mathrm{d} \varphi$.} The differential of $\varphi$ at a point $x \in M$, denoted as $\mathrm{d} \varphi_{x}$ is the best linear approximation of $\varphi$ near $x$. The differential is analogous to the total derivative in calculus. Mathematically speaking, the differential $\mathrm{d} \varphi$ is a linear mapping from the tangent space of $M$ at $x$ to the tangent space of $N$ at $\varphi (x)$, which is $\mathrm{d} \varphi_{x}: T_{x} M \rightarrow T_{\varphi (x)} N$. Given an equivalence class of curves $\gamma: (-1, 1) \rightarrow M$ where $\gamma(0) = x$, we have $\mathrm{d} \varphi_x\left(\gamma^{\prime}(0)\right)=(\varphi \circ \gamma)^{\prime}(0)$. Note that the linearity of the differential will be used in Eq.~\eqref{eq: score approximation from simple to general}.

\subsection{Introduction of spectral theorem}
One of the primary challenges encountered in the analysis of the adjacency matrix lies in the dependencies that exist between certain entries. When a subset of the entries in the adjacency matrix is accessible, it becomes possible to infer the values of other unknown entries. The spectral theorem offers insights into transforming the adjacency matrices into alternative representations, thus enabling the elimination of redundant entries. This transformation process greatly simplifies the study of the relationship between coordinates and adjacency matrices. The subsequent theorem provides a guideline for identifying a set of coordinates that corresponds to a given adjacency matrix.

\begin{theorem}[Adjacency matrix spectrum decomposition~\citep{schoenberg1935remarks, gower1982euclidean, hoffmann2019generating}]
    \label{thm: mapping from adjacency matrix to coordinates}
    For any feasible adjacency matrix $D \in \mathbb{R}_{+}^{n \times n}$ (Def.~\ref{def: feasible adjacency matrix}), there corresponds to a Gram matrix $M \in \mathbb{R}^{n \times n}$ by
    \begin{equation}
    M_{ij} = \frac{1}{2} (D_{1j} + D_{i1} - D_{ij})
    \end{equation}
    and vice versa
    \begin{equation}
    D_{ij} = M_{ii} + M_{jj} - 2 M_{ij}.
    \end{equation}
    Applying the singular value decomposition (SVD) to the Gram matrix $M \in \mathbb{R}^{n \times n}$ (associated with $D$), we can obtain exactly 3 positive eigenvalues $\lambda_1, \lambda_2, \lambda_3$ and their corresponding eigenvectors $\mathbf{v}_1, \mathbf{v}_2, \mathbf{v}_3$ satisfying $\lambda_1 \geq \lambda_2 \geq \lambda_3 > 0$. Then the set of coordinates
    \begin{equation}
        \mathcal{C}=[\mathbf{v}_1, \mathbf{v}_2, \mathbf{v}_3] \left[\begin{array}{lll}
        \centering
        \sqrt{\lambda_1} & \;\;\;0 & \;\;\;0 \\
        \;\;\; 0 & \sqrt{\lambda_2} & \;\;\;0 \\
        \;\;\;0 & \;\;\;0 & \sqrt{\lambda_3}
        \end{array}\right]
    \end{equation}
    satisfies $\adjp{\mathcal{C}} = D$. We denote such a mapping from a feasible adjacency matrix $D$ to a set of coordinates as $\varphi: \mathbb{R}^{n \times n}_{+} \rightarrow \mathbb{R}^{n \times 3}$.
\end{theorem}

\subsection{Introduction of Riemann manifold diffusion} 
 Let $M$ denote a complete, orientable connected, and boundaryless Riemannian manifold, endowed with a Riemannian metric. Let $\mathcal{B}^M_t$ be a Brownian motion on $M$. Under some regularity conditions, given a forward diffusion process by $\mathrm{d} X_t = f(X_t) \mathrm{d} t + g(t) \mathrm{d} \mathcal{B}_t^{M}$, where $X_t$ always stays in the manifold $M$, the reverse process of $X_t$ is associated with the SDE 

\begin{equation}
\mathrm{d} X_t = \left( f(X_t) - g^2(t) \nabla_{X} \log p_{t} \left( X_t \right) \right) \mathrm{d} t + g(t) \mathrm{d} \mathcal{B}_t^{M}.
\end{equation}

This reverse process can be discretized into $N$ steps by 

\begin{equation}
    X^{k-1} = \pi_{M, X^{k}} \left( X^{k} + \tau \left( -f(X^{k}) + g^2(kT/N) \nabla_{X} \log p_{kT/N} \left( X^{k} \right) \right) + \sqrt{\tau} g(kT/N) \mathbf{z}^{k} \right), \label{eq: Riemann reverse discrete}
\end{equation}

where $\mathbf{z}^{k}$ is a standard normal random variable in the tangent space of $T_{X^k} M$, and $\tau = T / N, k = N, \ldots, 1$~\citep{de2022riemannian}.

\section{Diffusion in manifolds}
\subsection{Manifolds related with SE(3)-invariant space} \label{sec: manifold definitions}

In the sampling process, the reverse diffusion trajectory in the embedding matrix manifold $L = \mathbb{R}^{n \times 3}$ is associated with a trajectory in the adjacency matrix manifold $M$. We can model the denoising trajectory in the adjacency matrix manifold to avoid considering the SE(3)-invariant property and then transform this adjacency matrix trajectory back to the embedding matrix manifold $L$. However, since the dimension of the adjacency manifold $M$ is smaller than that of the embedding manifold $L$, we cannot define a one-to-one mapping between $M$ and $L$. Instead, we consider a sub-manifold of the embedding manifold $L$, denoted by $N$ which shrinks the manifold dimension by removing the equivalent element w.r.t. translations and rotations. We want to construct a manifold $N$ containing a part of coordinate embeddings such that $M \cong N$ and $N \subset \mathbb{R}^{n \times 3}$. In the following, we give the detailed definitions of the above manifolds.

\paragraph{Embedding matrix manifold $L$.} We usually represent the coordinates of $n$ nodes by a matrix $\mathcal{C} \in \mathbb{R}^{n \times 3}$. All these representations form an embedding manifold $L$ and $L = \mathbb{R}^{n \times 3}$. 

\paragraph{Adjacency matrix manifold $M$.} $M$ is the manifold of the feasible adjacency matrix (Def.~\ref{def: feasible adjacency matrix}). Since the pairwise distances are invariant under rotations and translations, transforming the diffusion process in the embedding manifold $L$ to the manifold $M$ can avoid the difficulty in considering the SE(3)-invariance.

\paragraph{Spectral coordinate manifold $N$.} We define the spectral coordinate manifold by $N = \{ \varphi(d) \mid d \in M \}$. Since the $\varphi$ is a smooth mapping, the above-defined set is smooth. The intuition behind such a definition is that each element in $N$ is an equivalence class for all coordinates having the same adjacency matrix.

\subsection{Manifold geometries}
We consider mappings between manifolds so that we can transform the diffusion process in the embedding manifold $L$ to the adjacency matrix manifold $M$ or the spectral coordinate manifold $N$ to equip with the SE(3)-invariance. Consider a forward diffusion process on coordinates by

\begin{equation}
    \mathrm{d} \mathcal{C}_t = \mathbf{f}(\mathcal{C}_t, t) \mathrm{d} t + \mathbf{G}(\mathcal{C}_t, t) \mathrm{d} \mathbf{w}, \label{eq: general coordinate reverse flow}
\end{equation}

when we take some special forms of functions $\mathbf{f}$ and $\mathbf{G}$ (see Appendix~\ref{app: Fokker-Planck equation}), we write the corresponding evolving adjacency matrix $\gamma_1 \in \mathbb{R}_{+}^{n \times n}$ by:

\begin{equation}
    \mathrm{d} \gamma_1(t) = f(\gamma_1(t)) \mathrm{d} t + g(t) \mathrm{d} \mathcal{B}_t^{M}. \label{eq: general distance reverse flow}
\end{equation}

By the Riemann diffusion process~\citep{de2022riemannian}, the corresponding time-reverse diffusion SDE of $\gamma_1$ is given by 

\begin{equation}
\mathrm{d} \gamma_1(t) = \left( f(\gamma_1(t)) - g^2(t) \nabla_{\gamma_1} \log p_{t} \left( \gamma_1(t) \right) \right) \mathrm{d} t + g(t) \mathrm{d} \mathcal{B}_t^{M}. \label{eq: reverse SDE on d}
\end{equation}

Let
\begin{equation}
    F^{\mathcal{N}}(\gamma_1(t), t) = \tau \left(f(\gamma_1(t)) - g^2(t) \nabla_{\gamma_1} \log p_{t} \left( \gamma_1(t) \right) \right) + \sqrt{\tau} g(t) \mathbf{z}^{t, \gamma_1(t)}, \label{eq: F_normal}
\end{equation}
where $\tau = T / N$ and $\mathbf{z}^{t, \gamma_1(t)}$ is a standard normal random variable
in the tangent space of $T_{\gamma_1(t)} M$ which acts as the noise injected into $\gamma_1(t)$. Then Eq.~\eqref{eq: reverse SDE on d} is discretized by

\begin{equation}
    \gamma_1^{k-1} = \pi_{M, \gamma_{1}^{k}} \left( \gamma_{1}^{k} - F^{\mathcal{N}}(\gamma_1^k, kT/N) \right), \quad k = N, \ldots, 1.
\end{equation}

There is an associated curve $\gamma_2(t) = \varphi \circ \gamma_1(t)$ in the spectral coordinate manifold $N$ (by Theorem~\ref{thm: mapping from adjacency matrix to coordinates}) and the discretized trajectory can be computed by

\begin{equation}
    \gamma_{2}^{k-1} = \varphi \left( \gamma_{1}^{k-1} \right) = \varphi \circ \pi_{M, \gamma_{1}^{k}} \left( \gamma_{1}^{k} - F^{\mathcal{N}}(\gamma_1^k, kT/N) \right), \quad k = N, \ldots, 1. \label{eq: gamma2 reverse discre}
\end{equation}

Since $N$ is a submanifold of $L$, if we can solve the denoising process by Eq.~\eqref{eq: gamma2 reverse discre}, then at the end of the denoising process (at $\gamma_2(0)$), the obtained sample follows the distribution of $p_0$. To solve Eq.~\eqref{eq: gamma2 reverse discre}, we must analyze two functions $\varphi \circ \pi_M(\cdot)$ and $F(\cdot)$. In the following, we will approximate them respectively.

\subsection{Approximation of $\varphi \circ \pi_M$} \label{sec: approximation of differential}
In this section, we use $\tilde{d}$ to denote a feasible adjacency matrix in the denoising trajectory and we use $\hat{d}$ to denote a vector in the tangent space $T_{\tilde{d}} M$. We will use $\hat{\mathcal{C}} \in N$ to denote a set of coordinates whose adjacency matrix is close to $\hat{d}$, i.e. the solution of $\hat{C} = \varphi \circ \pi_{M, \tilde{d}}(\hat{d})$. Given $\hat{d} = \tilde{d} + \sum_{i < j} \alpha_{ij} \bar{e}_{ij} \in T_{\tilde{d}}M$, where $\bar{e}_{ij}$ is a zero matrix except for entries $(i, j)$ and $(j, i)$ being 1, our goal is to compute the value of $\varphi \circ \pi_{M, \tilde{d}}(\hat{d})$.

\subsubsection{Definition of $\varphi \circ \pi_M$} $\varphi \circ \pi_M$ is the mapping from $T_{\tilde{d}}M$ to $N$. In other words, given a non-necessarily feasible adjacency matrix $\hat{d} \in T_{\tilde{d}} M$, we want to find an element $\hat{\mathcal{C}} \in N$ such that $\| \adjj (\hat{\mathcal{C}}) - \hat{d} \| \leq \| \adjj (\mathcal{C}) - \hat{d} \|$ for all $\mathcal{C} \in N$. We write $\hat{d}$ in the form of $\displaystyle \hat{d} = \tilde{d} + \sum_{i < j} \alpha_{ij} \bar{e}_{ij} \in T_{\tilde{d}}M$, where $\bar{e}_{ij}$ is a zero matrix except for entries $(i, j)$ and $(j, i)$ being 1. $\tilde{d} \in M$ and $\tilde{d}$ is a feasible adjacency matrix. Define $\pi_{M, \tilde{d}}(\hat{d})_{ij} \triangleq \left\| \hat{\mathbf{x}}_i - \hat{\mathbf{x}}_j \right\|$, 
where $\displaystyle \hat{\mathbf{x}}_1, \ldots, \hat{\mathbf{x}}_n = \underset{\mathbf{x}_1, \ldots, \mathbf{x}_n}{\operatorname{argmin}} \sum_{i<j} \left( \left\| \mathbf{x}_i - \mathbf{x}_j \right\| - \hat{d}_{ij} \right)^2.$ Composed with $\varphi: M \rightarrow N$, we add the constraint $\mathcal{C} = [\mathbf{x}_1, \ldots, \mathbf{x}_n] \in N$ to the optimization problem and change the return values from the adjacency matrix to its associated coordinates $\mathcal{C}$ in the $N$ manifold. We have 

\begin{equation}
    \displaystyle \varphi \circ \pi_{M, \tilde{d}} (\hat{d}) = \varphi \circ \pi_{M, \tilde{d}} (\tilde{d} + \sum_{i < j} \alpha_{ij} \bar{e}_{ij}) = \underset{[\mathbf{x}_1, \ldots, \mathbf{x}_n] \in N}{\operatorname{argmin}} \sum_{i<j} \left( \left\| \mathbf{x}_i - \mathbf{x}_j \right\| - \hat{d}_{ij} \right)^2, \quad \hat{d} \in T_{\tilde{d}} M, \tilde{d} \in M. \label{eq: definition of phi pi M}
\end{equation}

In the reverse diffusion process, at $\tilde{d} = d_{t}$, our model takes the noised sample $\tilde{d}$ and predicts $\hat{d} \in T_{\tilde{d}} M$. Mathematically speaking, $T_{\tilde{d}} M$ has the same dimension as the $M$ manifold, which is less than $n^2$. However, we do not add any constraints on the model's output. The predicted $\hat{d}$ may not lay in $T_{\tilde{d}} M$. To consider the above cases, we consider a more general optimization problem by expanding the domain of $\hat{d}$ into $\mathbb{S}^{n}$ (all symmetric matrices). That is, given $\displaystyle \hat{d} = \tilde{d} + \sum_{i < j} \alpha_{ij} \bar{e}_{ij}$ for some $\alpha_{ij}$, we define 

\begin{equation}
    \displaystyle \varphi \circ \pi_{M, \tilde{d}} (\hat{d}) = \varphi \circ \pi_{M, \tilde{d}} (\tilde{d} + \sum_{i< j} \alpha_{ij} \bar{e}_{ij}) = \underset{[\mathbf{x}_1, \ldots, \mathbf{x}_n] \in N}{\operatorname{argmin}} \sum_{i<j} \left( \left\| \mathbf{x}_i - \mathbf{x}_j \right\| - \hat{d}_{ij} \right)^2, \quad \hat{d} \in \mathbb{R}^{n \times n}, \tilde{d} \in M.
\end{equation}

In the following, we will investigate how to solve the above equation by first considering a simple case of $\hat{d}$ where $\hat{d} = \tilde{d} + \delta \bar{e}_{uv}$ for some $u, v = 1, \ldots, n$ and then considering the generic case of $\hat{d}$ where $\displaystyle \hat{d} = \tilde{d} + \sum_{i < j} \alpha_{ij} \bar{e}_{ij}$.

\subsubsection{Simple case of $\varphi \circ \pi_M$}
\begin{figure}[H]
    \centering
    \includegraphics[height=.28\linewidth]{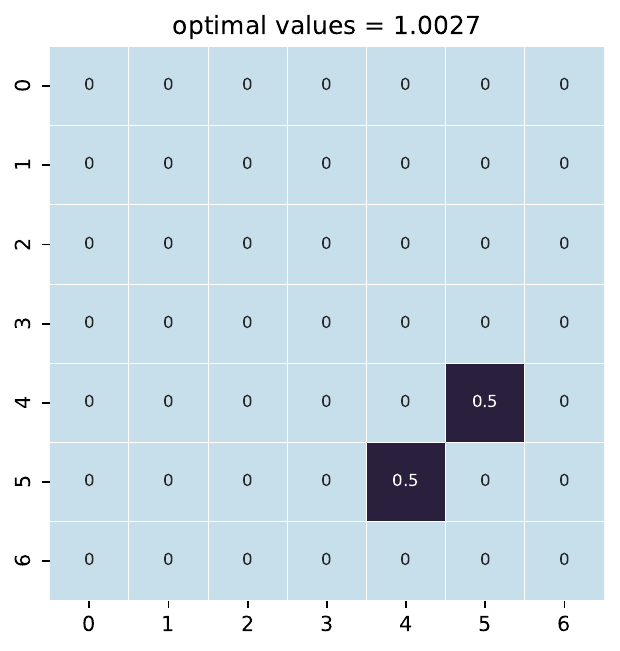}
    \includegraphics[height=.28\linewidth]{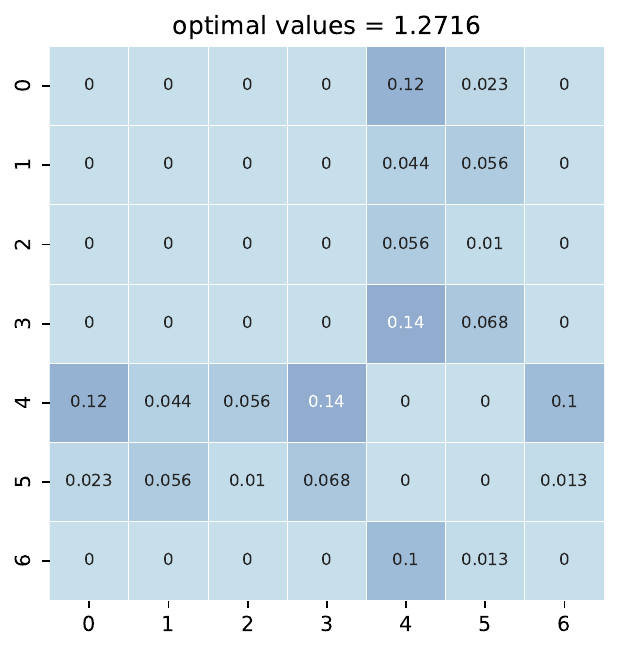}
    \includegraphics[height=.28\linewidth]{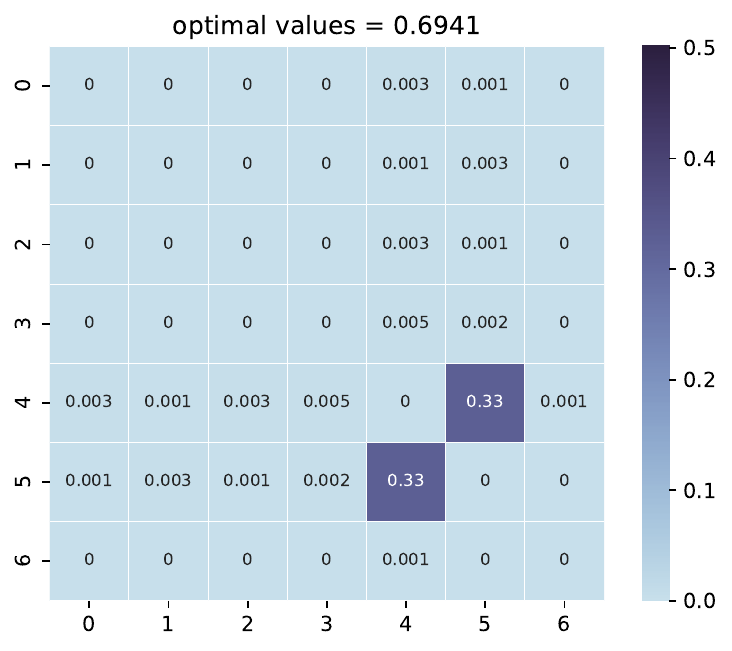}
    \caption{The illustration of how the choice of $s_u$ and $s_v$ affects the optimal values of $\varphi \circ \pi_{M, \tilde{d}}$. Given fixed $\delta, d, \hat{d}$, we set the optimal values as $\underset{\hat{\mathcal{C}} = [\mathbf{x}_1, \ldots, \mathbf{x}_n]}{\min} \sum_{i<j} \left( \left\| \mathbf{x}_i - \mathbf{x}_j \right\| - \hat{d}_{ij} \right)^2 / \delta^2$. In each figure, the matrix grids $(i, j)$ denote the entries of $( \adjj_{ij}(\mathcal{C}) - \hat{d}_{ij} )^2  / \delta^2$. We want to minimize the sum of grid values. The left figure denotes the case where we take $\hat{\mathcal{C}} = \tilde{\mathcal{C}}$. In this case, only the entry $(4, 5)$ introduces the loss to the optimal value but the value of the entry $(4, 5)$ is comparably large. The middle figure denotes the case where we take $s_{u} = s_{v} = \delta / 2$. In this case, the value of the entry $(4, 5)$ is reduced to zero but other entries become non-zero, and the total loss increases. In the right figure, we take $s_{u} = \delta / (2 \operatorname{degree}_u)$ and $s_{v} = \delta / (2 \operatorname{degree}_v)$. At this time, although the loss introduced by the entry $(u, v)$ is not reduced to $0$, the value of the entry sum is reduced.}
    \label{fig: why_n_1}
\end{figure}

Instead of directly computing $\displaystyle \varphi \circ \pi_{M, \tilde{d}} (\tilde{d} + \sum_{i < j} \alpha_{ij} \bar{e}_{ij})$, we first consider a simple case where $\hat{d} = \tilde{d} + \delta \bar{e}_{uv}$ and $\delta > 0$. That is, compared with the generic cases where $\displaystyle \hat{d} = \tilde{d} + \sum_{i < j} \alpha_{ij} \bar{e}_{ij}$, we assume that only one among $\{ \alpha_{ij} \}_{i, j = 1, \ldots, n, i < j}$ is non-zero. Computing $\displaystyle \varphi \circ \pi_{M, \tilde{d}} (\tilde{d} + \delta \bar{e}_{uv})$ is exactly the same as finding a set of coordinates $\mathcal{C} \in \mathbb{R}^{n \times 3}$ such that $\| \hat{d} - \adjp{\mathcal{C}} \|$ is minimized. Note that $\tilde{d}$ is associated with a set of coordinates $\tilde{\mathcal{C}} = [ \tilde{\mathbf{x}}_1, \ldots, \tilde{\mathbf{x}}_n ]^\top \in N$. If we take $\mathcal{C} = \tilde{\mathcal{C}}$, then the matrix $\hat{d} - \adjp{\mathcal{C}}$ has all 0 entries except entries $(u, v)$ and $(v, u)$ and these entries have the value $\delta$. We want to reduce the value of these entries so that the value of $\| \hat{d} - \adjj(\mathcal{C}) \|$ would reduce. To achieve this, we need to expand the edge length between node $u$ and $v$. A straightforward implementation is that we push node $u$ and $v$ in their opposite directions for some distance $s > 0$ and other nodes maintain in their positions. That is if we denote the obtained new coordinates by $\hat{\mathcal{C}} = [ \hat{\mathbf{x}}_1, \ldots, \hat{\mathbf{x}}_n ]^\top$, then we have 

$$ \left\{
\begin{aligned}
\hat{\mathbf{x}}_u &= \tilde{\mathbf{x}}_u + s_u \boldsymbol{\lambda}_{uv}, \\
\hat{\mathbf{x}}_v &= \tilde{\mathbf{x}}_v + s_v \boldsymbol{\lambda}_{vu}, \\
\hat{\mathbf{x}}_w &= \tilde{\mathbf{x}}_w, \quad \quad w \neq u, v,
\end{aligned}
\right.
$$

where $\boldsymbol{\lambda}_{uv} = \frac{\tilde{\mathbf{x}}_u - \tilde{\mathbf{x}}_v}{\left\| \tilde{\mathbf{x}}_u - \tilde{\mathbf{x}}_v \right\|}, \boldsymbol{\lambda}_{vu} = \frac{\tilde{\mathbf{x}}_v - \tilde{\mathbf{x}}_u}{\left\| \tilde{\mathbf{x}}_v - \tilde{\mathbf{x}}_u \right\|}$. Geometrically speaking, $\boldsymbol{\lambda}_{uv}, \boldsymbol{\lambda}_{vu}$ are opposite directions between node $u$ and $v$. If we take $s_u = s_v = \delta / 2$, then $\hat{d}_{uv} = \adjj_{uv}(\hat{\mathcal{C}})$. However, other entries in rows $u, v$ and columns $u, v$ of the matrix $\hat{d} - \adjj(\hat{\mathcal{C}})$ becomes non-zero, which may increase the matrix norm $\| \hat{d} - \adjj(\hat{\mathcal{C}}) \|$. Hence, choosing the value of $s$ is a tradeoff between the decrease of entries $(u, v), (v, u)$ and the increase of entries in rows and columns $u, v$.  To minimize $\| \hat{d} - \adjj(\hat{\mathcal{C}}) \|$, all entries except for entries $(u, v), (v, u)$ trend to remain $0$, this implies that in the graph of $\tilde{\mathcal{C}}$, edge length except for the edge length of  $uv$ trends to unchanged. When we push nodes $u$ and $v$ in their opposite directions, other edges that are incident to node $u$ and $v$ tend to rebel against such a push. Hence, we approximate that instead of both moving by a distance of $\delta / 2$, the node $u$ moves by $\delta / (2\operatorname{degree}_u)$ and the node $v$ moves by $\delta / (2\operatorname{degree}_v)$. Fig.~\ref{fig: why_n_1} gives a concrete example of the optimal values under different choices of $s_u$ and $s_v$.

When we consider a fully connected graph, $\operatorname{degree}_u = \operatorname{degree}_v = n-1$ and we have

$$ \left\{
\begin{aligned}
\hat{\mathbf{x}}_u &= \tilde{\mathbf{x}}_u + \frac{\delta}{2 (n-1)} \boldsymbol{\lambda}_{uv}, \\
\hat{\mathbf{x}}_v &= \tilde{\mathbf{x}}_v - \frac{\delta}{2 (n-1)} \boldsymbol{\lambda}_{uv}, \\
\hat{\mathbf{x}}_w &= \tilde{\mathbf{x}}_w, \quad \quad w \neq u, v,
\end{aligned}
\right.
$$

where $\boldsymbol{\lambda}_{uv} = \frac{\tilde{\mathbf{x}}_u - \tilde{\mathbf{x}}_v}{\left\| \tilde{\mathbf{x}}_u - \tilde{\mathbf{x}}_v \right\|}$. We want to write the above equations in a more compact way. Note that 

\begin{subequations}
    \begin{alignat}{2}
    (\frac{\partial \tilde{d}_{uv}}{\partial \tilde{\mathcal{C}}})_u &= (\frac{\partial \| \tilde{\mathbf{x}}_u - \tilde{\mathbf{x}}_v \|}{\partial \tilde{\mathcal{C}}})_u = \frac{\partial \| \tilde{\mathbf{x}}_u - \tilde{\mathbf{x}}_v \|}{\partial \tilde{\mathbf{x}}_u} = \frac{\tilde{\mathbf{x}}_u - \tilde{\mathbf{x}}_v}{\left\| \tilde{\mathbf{x}}_u - \tilde{\mathbf{x}}_v \right\|} &&= \boldsymbol{\lambda}_{uv}, \\
    (\frac{\partial \tilde{d}_{uv}}{\partial \tilde{\mathcal{C}}})_v &= (\frac{\partial \| \tilde{\mathbf{x}}_u - \tilde{\mathbf{x}}_v \|}{\partial \tilde{\mathcal{C}}})_v = \frac{\partial \| \tilde{\mathbf{x}}_u - \tilde{\mathbf{x}}_v \|}{\partial \tilde{\mathbf{x}}_v} = - \frac{\tilde{\mathbf{x}}_u - \tilde{\mathbf{x}}_v}{\left\| \tilde{\mathbf{x}}_u - \tilde{\mathbf{x}}_v \right\|} &&= - \boldsymbol{\lambda}_{uv}, \\
    (\frac{\partial \tilde{d}_{uv}}{\partial \tilde{\mathcal{C}}})_k &= (\frac{\partial \| \tilde{\mathbf{x}}_u - \tilde{\mathbf{x}}_v \|}{\partial \tilde{\mathcal{C}}})_k = \frac{\partial \| \tilde{\mathbf{x}}_u - \tilde{\mathbf{x}}_v \|}{\partial \tilde{\mathbf{x}}_k} = \mathbf{0}, && \quad k \neq u, v.
    \end{alignat}
\end{subequations}

Hence, we have 
\begin{equation}
    \hat{\mathcal{C}} = \tilde{\mathcal{C}} + \frac{\delta}{2 (n-1)} \frac{\partial \tilde{d}_{uv}}{\partial \tilde{\mathcal{C}}}, \label{eq: approximate scatter mean compact}
\end{equation}

where $\hat{\mathcal{C}} = [ \hat{\mathbf{x}}_1, \ldots, \hat{\mathbf{x}}_n ]^\top$. Under such an approximation, the optimal value is bounded by $\frac{2n^2 - 7n + 6}{2 (n-1)^2} \delta^2$ and the approximation error does not explode with the increase of node numbers. We refer readers to Appendix~\ref{app: scatter-mean approx error bound} for the proof.

Note that $\hat{\mathcal{C}}$ may not lie in the manifold $N$. We can find an element $\hat{\mathcal{C}_N} \in N$ such that $\adjp{\hat{\mathcal{C}_N}} = \adjp{\hat{\mathcal{C}}}$ by $\hat{\mathcal{C}_N} = \varphi \left( \adjp{\hat{\mathcal{C}}} \right)$. Then, we define $\pi_N = \varphi \circ \adjj$ and we have 

\begin{equation}
    \displaystyle \varphi \circ \pi_{M, \tilde{d}} (\tilde{d} + \delta \bar{e}_{uv}) \approx \varphi \left( \adjp{ \tilde{\mathcal{C}} + \frac{\delta}{2 (n-1)} \frac{\partial \tilde{d}_{uv}}{\partial \tilde{\mathcal{C}}} } \right). \label{eq: score approximation simple case}
\end{equation}

By the definition of the differential, $\mathrm{d} \varphi_{\tilde{d}} \left(\gamma^{\prime}(0)\right)=(\varphi \circ \gamma)^{\prime}(0)$, $\mathrm{d} \varphi_{\tilde{d}}$ maps a vector in the tangent space $T_{\tilde{d}} M$ to the tangent space $T_{\tilde{\mathcal{C}}} N$. Again, applying the above approximation for the function $\varphi \circ \pi_M$, we have 

\begin{equation}
    \displaystyle \mathrm{d} \varphi_{\tilde{d}} (\tilde{d} + \delta \bar{e}_{uv}) \approx \tilde{\mathcal{C}} + \frac{\delta}{2 (n-1)} \frac{\partial \tilde{d}_{uv}}{\partial \tilde{\mathcal{C}}}. \label{eq: differential approximation simple case}
\end{equation}

\begin{figure}[H]
    \centering
    \includegraphics[width=.6\linewidth]{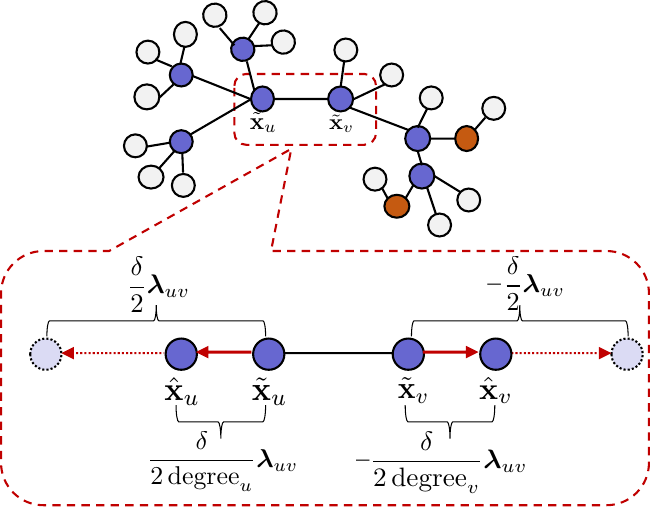}
    \caption{The illustration of approximating the solution of $\varphi \circ \pi_M(\tilde{d} + \delta \bar{e}_{uv})$}
    \label{fig:enter-label}
\end{figure}

\subsubsection{Generic case of $\varphi \circ \pi_M$} \label{app: generic case approximation}

In the above section, we approximate the value for $\displaystyle \varphi \circ \pi_{M, \tilde{d}} (\tilde{d} + \delta \bar{e}_{uv})$. In this section, we consider generic cases where $\displaystyle \hat{d} = \tilde{d} + \sum_{i < j} \alpha_{ij} \bar{e}_{ij} \triangleq \tilde{d} + \sum_{i, j} \alpha_{ij} e_{ij}$, where $e_{ij}$ is a zero matrix except for the entry $(i, j)$ and $\alpha_{ij} = \alpha_{ji}$. Consider the mapping $f: T_{\tilde{d}}M \rightarrow N$. There are two ways to transform from $TM$ to $N$. One is by $TM \rightarrow TN \rightarrow N$ and the other one is by $TM \rightarrow M \rightarrow N$. The former transformation is defined by the composition $\pi_N \circ \mathrm{d} \varphi$ and the latter by $\varphi \circ \pi_M$. Hence, $f = \varphi \circ \pi_M = \pi_N \circ \mathrm{d} \varphi$. Then, we have 

\begin{equation}
    \pi_N \circ \mathrm{d} \varphi (\tilde{d} + \delta \bar{e}_{uv}) = \varphi \circ \pi_{M, \tilde{d}} (\tilde{d} + \delta \bar{e}_{uv}) \approx \varphi \circ \adjp{ \tilde{\mathcal{C}} + \frac{\delta}{2 (n-1)} \frac{\partial \tilde{d}_{uv}}{\partial \tilde{\mathcal{C}}} }
\end{equation}

and 

\begin{subequations}
\label{eq: score approximation from simple to general}
    \begin{alignat}{2}
        \varphi \circ \pi_{M, \tilde{d}} (\hat{d}) &= \varphi \circ \pi_{M, \tilde{d}} (\tilde{d} + \sum_{i < j} \alpha_{ij} \bar{e}_{ij}) \\
        &= \pi_N \circ \mathrm{d} \varphi_{\tilde{d}} (\tilde{d} + \sum_{i < j} \alpha_{ij} \bar{e}_{ij}) \quad \quad & (\varphi \circ \pi_{M, \tilde{d}} = \pi_N \circ \mathrm{d} \varphi_{\tilde{d}}) \\
        &= \pi_N \left( (1 - \frac{n^2 - n}{2}) \mathrm{d} \varphi_{\tilde{d}} (\tilde{d}) + \sum_{i < j}  \mathrm{d} \varphi_{\tilde{d}} (\tilde{d} + \alpha_{ij} \bar{e}_{ij}) \right) \quad \quad & (\text{differential is linear}) \\
        &\approx \pi_N \left( (1 - \frac{n^2 - n}{2}) \tilde{\mathcal{C}} + \sum_{i < j} \tilde{\mathcal{C}} + \frac{\alpha_{ij}}{2 (n-1)} \frac{\partial \tilde{d}_{ij}}{\partial \tilde{\mathcal{C}}} \right) \quad \quad & (\text{induced from Eq.~\eqref{eq: differential approximation simple case}})\\
        &= \varphi \circ \adjj \left( \tilde{\mathcal{C}} + \sum_{i < j} \frac{\alpha_{ij}}{2 (n-1)} \frac{\partial \tilde{d}_{ij}}{\partial \tilde{\mathcal{C}}} \right) \quad \quad &  (\pi_N = \varphi \circ \adjj) \\
        &= \varphi \circ \adjj \left( \tilde{\mathcal{C}} + \frac{\alpha}{2 (n-1)} \circledast \frac{\partial \tilde{d}}{\partial \tilde{\mathcal{C}}} \right) \quad \quad &  (\text{by Def.~\ref{def: matrix2 times matrix4}})
    \end{alignat}
\end{subequations}

where $\frac{\partial \tilde{d}}{\partial \tilde{\mathcal{C}}} = \left( \frac{\partial \tilde{d}_{ij}}{\partial \tilde{\mathcal{C}}} \right)_{n \times n}$ and $\alpha = \left( \alpha_{ij} \right)_{n \times n}, \alpha_{ij} = \alpha_{ji}$. The formula for $\varphi \circ \pi_{M, \tilde{d}}$ looks complex. In the following, we will show that at the implementation level, we can drop the $\varphi \circ \adjj$ term and thus $\displaystyle \varphi \circ \pi_{M, \tilde{d}} (\hat{d}) = \tilde{\mathcal{C}} + \frac{\alpha}{2 (n-1)} \circledast \frac{\partial \tilde{d}}{\partial \tilde{\mathcal{C}}}$. The function becomes linear. 

In the above analysis, we restrict $\hat{\mathcal{C}}$ to be in the manifold of the spectral coordinate manifold $N$. However, in practice, we are not interested in whether $\hat{\mathcal{C}}$ lies in the manifold $N$ or not, as long as the denoising process of the adjacency matrix does not change. Note that the composition function $\varphi \circ \adjj$ does not affect the value of the adjacency matrix. Hence, at the implementation level, we can remove this function. Mathematically speaking, given $\tilde{d}^{k} \in M$ and its associated spectral coordinates $\tilde{C}^{k} \in N$, in each update, $\hat{d}^{k-1} = \tilde{d}^{k} - F^{\mathcal{N}}(\tilde{d}^{k}, kT / N)$ and we update $\tilde{C}^{k}$ by

\begin{equation}
    \tilde{\mathcal{C}}^{k-1} = \varphi \circ \adjp{ \tilde{\mathcal{C}}^{k} - \frac{F^{\mathcal{N}}(\tilde{d}^{k}, kT / N)}{2 (n-1)} \circledast \frac{\partial \tilde{d}^{k}}{\partial \tilde{\mathcal{C}}^{k}} } \simeq \tilde{\mathcal{C}}^{k} - \frac{F^{\mathcal{N}}(\tilde{d}^{k}, kT / N)}{2 (n-1)} \circledast \frac{\partial \tilde{d}^{k}}{\partial \tilde{\mathcal{C}}^{k}}.
\end{equation}

At the implementation level, substituting $F^{\mathcal{N}}$ by Eq.~\eqref{eq: F_normal}, we can update $\gamma_2$ or $\mathcal{C}$ by

\begin{subequations}
    \label{eq: simplified reverse c}
    \begin{align}
        \mathcal{C}^{k-1} &= \mathcal{C}^{k} - \frac{F^{\mathcal{N}}(d^{k}, kT / N)}{2 (n-1)} \circledast \frac{\partial d^{k}}{\partial \mathcal{C}^{k}} \\
        &= \mathcal{C}^{k} + \frac{\tau}{2 (n-1)} \left( -f(d^{k}) + g^2(kT / N) \nabla_{d} \log p_{kT / N} ( d^{k} ) \right) \circledast \frac{\partial d^{k}}{\partial \mathcal{C}^{k}} + \frac{\sqrt{\tau}}{2 (n-1)} g(kT / N) \mathbf{z}^{k, d^{k}} \circledast \frac{\partial d^{k}}{\partial \mathcal{C}^{k}}.
    \end{align}
\end{subequations}

\subsection{Elimination of functions related to the adjacency matrix} \label{sec: Elimination of functions related to the adjacency matrix}
Since we do not know the form of the functions $f$ and $g$, we may consider trying to replace $f$ and $g$ with $\mathbf{f}$ and $\mathbf{G}$, which are the hyperparameters and the noise scheduler on coordinates. Remind that given the forward diffusion process with some specific choice of $\mathbf{f}$ and $\mathbf{G}$ noise scheduler on coordinates by

\begin{equation}
    \mathrm{d} \mathcal{C}_t = \mathbf{f}(\mathcal{C}_t, t) \mathrm{d} t + \mathbf{G}(\mathcal{C}_t, t) \mathrm{d} \mathbf{w},
\end{equation}

we denote its associated diffusion of adjacency matrices as

\begin{equation}
    \mathrm{d} \gamma_1(t) = f(\gamma_1(t)) \mathrm{d} t + g(t) \mathrm{d} \mathcal{B}_t^{M}.
\end{equation}

We discretize the above two equations and we have

\begin{subequations}
    \begin{alignat}{2}
        \mathcal{C}^{k+1} &= \mathcal{C}^k + \tau \mathbf{f}(\mathcal{C}^k, kT / N) +  \sqrt{\tau} \mathbf{G}(\mathcal{C}^k, kT / N) \mathbf{z}^{k, \mathcal{C}^k}, \\
        \mathcal{C}^{k+1} &= \varphi (d^{k+1}) =  \varphi \circ \pi_{M, d^k} \left( d^k + \tau f(d^k) + \sqrt{\tau} g(kT/N) \mathbf{z}^{k, d^k} \right) \\
        &= \varphi \circ  \adjp{ \mathcal{C}^k + \frac{1}{2 (n-1)} \left( \tau f(d^k) + \sqrt{\tau} g(kT/N) \mathbf{z}^{k, d^k} \right) \circledast \frac{\partial d_t}{\partial \mathcal{C}_t} } &\quad \text{(by Sec.~\ref{sec: approximation of differential})} \\
        &= \mathcal{C}^k + \frac{1}{2 (n-1)} \left( \tau f(d^k) + \sqrt{\tau} g(kT/N) \mathbf{z}^{k, d^k} \right) \circledast \frac{\partial d_t}{\partial \mathcal{C}_t}, &\quad \text{(by Def.~\ref{def: equi coordinate})}
    \end{alignat}
\end{subequations}

where $\tau = T / N$ and each entry of $\mathbf{z}^{k, \mathcal{C}^k}$ and $\mathbf{z}^{k, d^k}$ follow the normal distribution. This implies 

\begin{equation}
    \tau \mathbf{f}(\mathcal{C}^k, kT / N) +  \sqrt{\tau} \mathbf{G}(\mathcal{C}^k, kT / N) \mathbf{z}^{k, \mathcal{C}^k} = \frac{1}{2 (n-1)} \left( \tau f(d^k) + \sqrt{\tau} g(kT/N) \mathbf{z}^{k, d^k} \right) \circledast \frac{\partial d_t}{\partial \mathcal{C}_t}.
\end{equation}

Substitute Eq.~\eqref{eq: simplified reverse c} with the above equation, we have 

\begin{subequations}
    \begin{align}
        \tilde{\mathcal{C}}^{k-1} &= \tilde{\mathcal{C}}^{k} - \left( \tau \mathbf{f}(\mathcal{C}^k, kT / N) +  \sqrt{\tau} \mathbf{G}(\mathcal{C}^k, kT / N) \mathbf{z}^{k, \mathcal{C}^k} \right) + \frac{\tau g^2(kT / N) }{2 (n-1)} \nabla_{d} \log p_{kT / N} ( d^{k} ) \circledast \frac{\partial d^{k}}{\partial \mathcal{C}^{k}}.
    \end{align}
\end{subequations}

The above discretization corresponds to the SDE

\begin{equation}
    \mathrm{d} \mathcal{C}_t = \left( \mathbf{f}(\mathcal{C}_t, t) - \frac{g^2(t) }{2 (n-1)} \nabla_{d} \log p_{t} ( d_t ) \circledast \frac{\partial d_t}{\partial \mathcal{C}_t} \right) \mathrm{d} t + \mathbf{G}(\mathcal{C}_t, t) \mathrm{d} \bar{\mathcal{\mathbf{w}}}. \label{eq: coordinates final sde}
\end{equation}

Hence, in the sampling process, if we can solve Eq.~\eqref{eq: coordinates final sde}, then the obtained $\mathcal{C}_0$ has an adjacency matrix $d_0 = \adjp{\mathcal{C}_0}$ follows the distribution of $p_0$. However, directly solving Eq.~\eqref{eq: coordinates final sde} may cost thousands of iterations. In the following, we consider a reverse ODE which has the same marginal distribution as the reverse SDE given by Eq.~\eqref{eq: coordinates final sde}.

\subsection{Reverse ODE} \label{sec: reverse ODE}
\subsubsection{Adjacency matrices change along with the perturbation of coordinates}
We want to investigate how the adjacency matrices evolve along with the perturbation of coordinates, which will be useful in computing the SDE of the adjacency matrix. We only consider a simple case where the noise is injected on coordinates following the variance exploding noise scheduler. Given a forward diffusion process of the conformation $\tilde{\mathcal{C}} = \mathcal{C} + \mathbf{z}^{\mathcal{C}} \in \mathbb{R}^{n \times 3}$, where $\mathbf{z}_i^{\mathcal{C}} \sim \mathcal{N}(\mathbf{0}, \sigma^2 \mathbf{I}_{3 \times 3}), i=1, \ldots, n$, we have

\begin{subequations}
    \begin{align}
        \tilde{\mathcal{C}}_u &= \mathcal{C}_u + \mathbf{z}_u \qquad \tilde{\mathcal{C}}_v = \mathcal{C}_v + \mathbf{z}_v \\
        \tilde{d}_{uv} &= \| \tilde{\mathcal{C}}_u - \tilde{\mathcal{C}}_v \| \\
        &= \| \tilde{\mathcal{C}}_u - \mathcal{C}_u + \mathcal{C}_u - \mathcal{C}_v + \mathcal{C}_v - \tilde{\mathcal{C}}_v \| \\
        &= \| \mathbf{z}_u - \mathbf{z}_v + \mathcal{C}_u - \mathcal{C}_v \| \\
        &\simeq \| \mathbf{z} + \mathcal{C}_u - \mathcal{C}_v\| \quad \quad (\mathbf{z} \sim \mathcal{N}(\mathbf{0}, 2\sigma^2 \mathbf{I}_{3 \times 3})) \\
        &= \| \mathcal{C}_u - \mathcal{C}_v \| + \| \mathbf{z} + \mathcal{C}_u - \mathcal{C}_v\| - \| \mathcal{C}_u - \mathcal{C}_v \| \\
        &= \| \mathcal{C}_u - \mathcal{C}_v \| + (\| \mathbf{z} + \mathcal{C}_u - \mathcal{C}_v\| - \| \mathcal{C}_u - \mathcal{C}_v \|) \frac{\| \mathbf{z} + \mathcal{C}_u - \mathcal{C}_v\| + \| \mathcal{C}_u - \mathcal{C}_v \|}{\| \mathbf{z} + \mathcal{C}_u - \mathcal{C}_v\| + \| \mathcal{C}_u - \mathcal{C}_v \|} \\
        &= \| \mathcal{C}_u - \mathcal{C}_v \| + \frac{\| \mathcal{C}_u - \mathcal{C}_v \|^2 + 2 \mathbf{z}^\top (\mathcal{C}_u - \mathcal{C}_v) + \|\mathbf{z}\|^2 - \| \mathcal{C}_u - \mathcal{C}_v \|^2}{\| \mathbf{z} + \mathcal{C}_u - \mathcal{C}_v\| + \| \mathcal{C}_u - \mathcal{C}_v \|} \\
        &= \| \mathcal{C}_u - \mathcal{C}_v \| + \frac{2 \mathbf{z}^\top (\mathcal{C}_u - \mathcal{C}_v) + \|\mathbf{z}\|^2}{\| \mathbf{z} + \mathcal{C}_u - \mathcal{C}_v\| + \| \mathcal{C}_u - \mathcal{C}_v \|} \\
        &= d_{uv} + \frac{2 \mathbf{z}^\top (\mathcal{C}_u - \mathcal{C}_v) + \|\mathbf{z}\|^2}{\| \mathbf{z} + \mathcal{C}_u - \mathcal{C}_v\| + \| \mathcal{C}_u - \mathcal{C}_v\|}
        \end{align}
\end{subequations}

In terms of the diffusion SDE, given a forward diffusion process on coordinates by

\begin{equation}
    \mathrm{d} \mathcal{C}_t = \mathbf{G}(t) \mathrm{d} \mathbf{w}, \label{eq: forward sde c ve case}
\end{equation}

where $\mathbf{G}(t)$ is a scalar function. We discretize the forward diffusion process as

\begin{equation}
    \mathcal{C}^{t + \tau} = \mathcal{C}^t + \sqrt{\tau} \mathbf{G}(t) \mathbf{z}^{t, \mathcal{C}^t},
\end{equation}

where $\mathbf{z}^{t, \mathcal{C}^t} \in \mathbb{R}^{n \times 3}, \mathbf{z}^{t, \mathcal{C}^t}_i \sim \mathcal{N}(\mathbf{0}, \mathbf{I}_{3 \times 3})$.  Let $\mathbf{m}^{t, \mathcal{C}^t} = \sqrt{\tau} \mathbf{G}(t) \mathbf{z}^{t, \mathcal{C}^t}$. Consider the associated diffusion of the adjacency matrix $d^t = \adjp{\mathcal{C}^t}$, we have, for $u \neq v$,

\begin{subequations}
    \label{eq: d difference order 1}
    \begin{align}
        & \quad d_{uv}^t - d_{uv}^{t + \tau} \\ 
        &= d_{uv}^t - \adjj_{uv} \left( \mathcal{C}^t + \mathbf{m}^{t, \mathcal{C}^t} \right) \\
        &= d_{uv}^t - \left[ d_{uv}^t + \left( 2 {\mathbf{m}_{uv}^{t}}^\top \left( \mathcal{C}_u - \mathcal{C}_v \right) + \| \mathbf{m}_{uv}^t \|^2 \right) / M_{uv}^t \right] \\
        &= - \left( 2 {\mathbf{m}_{uv}^t}^\top \left( \mathcal{C}_u - \mathcal{C}_v \right) + \| \mathbf{m}_{uv}^t \|^2 \right) / M_{uv}^t \label{eq: linear distance shift 1} \\
        &= - 2 \sqrt{2\tau} \mathbf{G}(t) {\mathbf{z}_{uv}^t}^\top \left( \mathcal{C}_u - \mathcal{C}_v \right) / M_{uv}^t + \left( 2\tau \mathbf{G}^2(t)\| \mathbf{z}_{uv}^t \|^2 \right) \cdot \left( 1 / \left( 2\| \mathcal{C}_u - \mathcal{C}_v \| \right) + \lito(\tau^{1/3}) \right) \label{eq: linear distance shift 2} \\
        &= - 2 \sqrt{2\tau} \mathbf{G}(t) {\mathbf{z}_{uv}^t}^\top \left( \mathcal{C}_u - \mathcal{C}_v \right) / M_{uv}^t + \tau \mathbf{G}^2(t)\| \mathbf{z}_{uv}^t \|^2 / d_{uv} + \lito(\tau) 
    \end{align}
\end{subequations}

where $\mathbf{m}_{uv}^t = \sqrt{\tau} \mathbf{G}(t) \left( \mathbf{z}^{t, \mathcal{C}_u} - \mathbf{z}^{t, \mathcal{C}_v} \right) \triangleq \sqrt{2\tau} \mathbf{G}(t) \mathbf{z}_{uv}^t, \mathbf{z}_{uv}^t \sim \mathcal{N}(\mathbf{0}, \mathbf{I}_{3 \times 3})$, $M_{uv}^t = \| \mathbf{m}_{uv}^t + \mathcal{C}_u - \mathcal{C}_v \| + \| \mathcal{C}_u - \mathcal{C}_v \|$.

\begin{subequations}
    \label{eq: d difference order 2}
    \begin{align}
        & \quad \left( d_{uv}^t - d_{uv}^{t + \tau} \right)^2 \\ 
        &= \left( 2 {\mathbf{m}_{uv}^t}^\top \left( \mathcal{C}_u - \mathcal{C}_v \right) + \| \mathbf{m}_{uv}^t \|^2 \right)^2 / \left( M_{uv}^t \right)^2 \\
        &= \left( 2 {\mathbf{m}_{uv}^t}^\top \left( \mathcal{C}_u - \mathcal{C}_v \right) + \| \mathbf{m}_{uv}^t \|^2 \right)^2 \cdot \left( 1 / \left( 2\| \mathcal{C}_u - \mathcal{C}_v \| \right) + \lito(\tau^{1/3}) \right)^2 \\
        &= 4 \cdot 2 \tau \mathbf{G}^2(t) \left( {\mathbf{z}_{uv}^t}^\top \left( \mathcal{C}_u - \mathcal{C}_v \right) \right)^2 / \left( 4 \| \mathcal{C}_u - \mathcal{C}_v \|^2 \right) + \lito(\tau)\\
        &= 2 \tau \mathbf{G}^2(t) \left( {\mathbf{z}_{uv}^t}^\top \frac{\mathcal{C}_u - \mathcal{C}_v}{ \| \mathcal{C}_u - \mathcal{C}_v \|} \right)^2 + \lito(\tau)
    \end{align}
\end{subequations}

From Eq.~\eqref{eq: linear distance shift 1} to Eq.~\eqref{eq: linear distance shift 2} is beacuse if $\mathbf{x} \neq \mathbf{0}$, then

\begin{subequations}
    \begin{align}
        & \quad \lvert \frac{1}{\| \sqrt{\tau} \mathbf{z} + \mathbf{x} \| + \| \mathbf{x} \|} -\frac{1}{2 \| \mathbf{x} \|} \rvert \\
        &= \frac{\lvert 2 \| \mathbf{x} \| - \| \mathbf{x} \| - \| \sqrt{\tau} \mathbf{z} + \mathbf{x} \| \rvert}{2  \| \mathbf{x} \| \left( \| \sqrt{\tau} \mathbf{z} + \mathbf{x} \| + \| \mathbf{x} \| \right)} \\
        &\leq \frac{\lvert \| \mathbf{x} \| - \| \sqrt{\tau} \mathbf{z} + \mathbf{x} \| \rvert}{2 \| \mathbf{x} \|^2} \\
        &= \frac{\lvert \| \mathbf{x} \|^2 - \| \sqrt{\tau} \mathbf{z} + \mathbf{x} \|^2 \rvert}{2 \| \mathbf{x} \|^2 \left( \| \mathbf{x} \| + \| \sqrt{\tau} \mathbf{z} + \mathbf{x} \| \right) }  \\
        &\leq \frac{\lvert 2 \sqrt{\tau} \mathbf{z}^\top \mathbf{x} + \tau \| \mathbf{z} \|^2 \rvert}{2 \| \mathbf{x} \|^3 }  \\
        &= \lito(\tau^{1/3})
    \end{align}
\end{subequations}

We have $\frac{1}{\| \sqrt{\tau} \mathbf{z} + \mathbf{x} \| + \| \mathbf{x} \|}  = \frac{1}{2 \| \mathbf{x} \|} + \lito(\tau^{1/3})$. Using Eq.~\eqref{eq: d difference order 1} and Eq.~\eqref{eq: d difference order 2} and approximating $\mathbb{E}[{\mathbf{z}_{uv}^t}^\top \left( \mathcal{C}_u - \mathcal{C}_v \right) / M_{uv}^t] \approx 0$, directly we have 

\begin{subequations}
    \begin{align}
        \mathbb{E}_{\mathbf{z}_{uv}^t} \left[ d_{uv}^t - d_{uv}^{t + \tau} \right] &\approx - 3 \tau \mathbf{G}^2(t) / d_{uv}^t + \lito(\tau), \\
        \mathbb{E}_{\mathbf{z}_{uv}^t} \left[ \left( d_{uv}^t - d_{uv}^{t + \tau} \right)^2 \right] &= 2 \tau \mathbf{G}^2(t) + \lito(\tau).
    \end{align}
\end{subequations}

Note that there exist an exact solution of $\mathbb{E}_{\mathbf{z}_{uv}^t} \left[ d_{uv}^t - d_{uv}^{t + \tau} \right]$.

\begin{subequations}
    \begin{align}
        & \quad \mathbb{E}_{\mathbf{z}_{uv}^t} \left[ d_{uv}^t - d_{uv}^{t + \tau} \right] \\ 
        &= \mathbb{E}_{\mathbf{z}_{uv}^t} \left[ d_{uv}^t - \| \mathbf{m}_{uv}^t + \mathcal{C}_u - \mathcal{C}_v \| \right]\\
        &= \mathbb{E}_{\mathbf{z}_{uv}^t} \left[ d_{uv}^t - \| \sqrt{2 \tau} \mathbf{G}(t) \mathbf{z}_{uv}^t + \mathcal{C}_u - \mathcal{C}_v \| \right] \\
        &\triangleq  d_{uv}^t - \mathbb{E}_{\mathbf{z}_{uv}^t} \left[ \| X \| \right]
    \end{align}
\end{subequations}

where $X \triangleq \mathcal{N}(\mathcal{C}_u - \mathcal{C}_v, 2\tau \mathbf{G}^2(t))$. It is known that if $X \sim \mathcal{N}(\mu, \sigma^2 \mathbf{I}_{n \times n})$, then $\| X \| / \sigma$ follows a noncentral chi distribution~\citep{park1961moments}, which has the mean $\mathbb{E}\|X\|=\sigma \sqrt{\frac{\pi}{2}} L_{1 / 2}^{(n / 2-1)}\left(\frac{-\|\mu\|^2}{2 \sigma}\right)$, where $L$ is the generalized Laguerre polynomial~\citep{sonine1880recherches}. Hence, we have 

\begin{equation}
    \mathbb{E}_{\mathbf{z}_{uv}^t} \left[ d_{uv}^t - d_{uv}^{t + \tau} \right] = \sqrt{\tau \pi} \mathbf{G}(t) L^{1/2}_{1/2} \left( \frac{- \| \mathcal{C}_u - \mathcal{C}_v \|^2}{2 \sqrt{2\tau} \mathbf{G}(t)} \right).
\end{equation}

\subsubsection{Fokker-Planck equation and forward process of the adjacency matrix distribution} \label{app: Fokker-Planck equation}
To find the reverse ODE which has the same marginal distribution as the forward diffusion, one may consider first deriving the Fokker-Planck equation~\citep{risken1996fokker, song2020score}. We consider the marginal distribution of the adjacency matrix corresponding to the forward diffusion process of coordinates given by

\begin{equation}
    \mathrm{d} \mathcal{C}_t = \mathbf{G}(t) \mathrm{d} \mathbf{w}, \quad \mathcal{C}^{t + \tau} = \mathcal{C}^t + \sqrt{\tau} \mathbf{G}(t) \mathbf{z}^{t, \mathcal{C}^t}.
\end{equation}

Consider the Dirac delta function

\begin{subequations}
    \begin{align}
        \delta \left( d_{uv} - d_{uv}^{t + \tau} \right) &= \delta \left( d_{uv} - d_{uv}^t + d_{uv}^t - d_{uv}^{t + \tau} \right) \\
        &= \delta \left( d_{uv} - d_{uv}^t \right) + \left( d_{uv}^t - d_{uv}^{t + \tau} \right) \nabla_{d_{uv}} \delta \left( d_{uv} - d_{uv}^t \right) \\
        & \quad + \frac{1}{2} \left( d_{uv}^t - d_{uv}^{t + \tau} \right)^2 \nabla_{d_{uv}}^2 \delta \left( d_{uv} - d_{uv}^t \right) + \lito(\tau) \nonumber
    \end{align}
\end{subequations}

\begin{subequations}
    \begin{align}
        p_{t + \tau}\left( d_{uv} \right) &= \mathbb{E}_{d_{uv}^{t + \tau}} \left[ \delta \left( d_{uv} - d_{uv}^{t + \tau}  \right) \right] \\
        &= \mathbb{E}_{d_{uv}^{t}, \mathbf{m}^{t, \mathcal{C}_t}} \left[ \delta \left( d_{uv} - d_{uv}^t \right) \right] +  \mathbb{E}_{d_{uv}^{t}, \mathbf{m}^{t, \mathcal{C}_t}} \left[ \left( d_{uv}^t - d_{uv}^{t + \tau} \right) \nabla_{d_{uv}} \delta \left( d_{uv} - d_{uv}^t \right) \right] \\
        & \quad + \frac{1}{2}  \mathbb{E}_{d_{uv}^{t}, \mathbf{m}^{t, \mathcal{C}_t}} \left[ \left( d_{uv}^t - d_{uv}^{t + \tau} \right)^2 \nabla_{d_{uv}}^2 \delta \left( d_{uv} - d_{uv}^t \right) \right] + \lito(\tau) \nonumber \\
        &= p_t(d_{uv}) + \mathbb{E}_{d_{uv}^{t}} \left[ - 3 \tau \mathbf{G}^2(t) / \| \mathcal{C}_u - \mathcal{C}_v \| \cdot \nabla_{d_{uv}} \delta \left( d_{uv} - d_{uv}^t \right) \right] \\
        & \quad + \frac{1}{2}  \mathbb{E}_{d_{uv}^{t}} \left[ 2 \tau \mathbf{G}^2(t) \cdot \nabla_{d_{uv}}^2 \delta \left( d_{uv} - d_{uv}^t \right) \right] + \lito(\tau) \nonumber \\
        &= p_t(d_{uv}) - \nabla_{d_{uv}} \left[ 3 \tau \mathbf{G}^2(t) / \| \mathcal{C}_u - \mathcal{C}_v \| \cdot p_t \left( d_{uv} \right) \right] + \tau \mathbf{G}^2(t) \cdot \nabla_{d_{uv}}^2 p_t \left( d_{uv} \right) + \lito(\tau)
    \end{align}
\end{subequations}

Take $\tau \rightarrow 0$, we have

\begin{equation}
    \frac{\partial}{\partial t} p_t (d_{uv}) = - \nabla_{d_{uv}} \left[ 3 \mathbf{G}^2(t) / \| \mathcal{C}_u - \mathcal{C}_v \| \cdot p_t \left( d_{uv} \right) \right] + \mathbf{G}^2(t) \cdot \nabla_{d_{uv}}^2 p_t \left( d_{uv} \right). \label{eq: FP for distance 1}
\end{equation}

The above Fokker-Planck equation has the solution of the forward diffusion process by~\citep{ottinger2012stochastic}

\begin{equation}
    \mathrm{d} d_{uv}^t = 3 \mathbf{G}^2(t) / d_{uv}^t \mathrm{d} t + \sqrt{2} \mathbf{G}(t) \mathrm{d} \mathcal{B}_t. \label{eq: forward sde of d 1}
\end{equation}

\subsubsection{Reverse ODE}
Note that we can rewrite Eq.~\eqref{eq: FP for distance 1} as 

\begin{subequations}
    \label{eq: FP for distance 2}
    \begin{align}
        \frac{\partial}{\partial t} p_t (d_{uv}) &= - \nabla_{d_{uv}} \left[ 3 \mathbf{G}^2(t) / \| \mathcal{C}_u - \mathcal{C}_v \| \cdot p_t \left( d_{uv} \right) \right] + \mathbf{G}^2(t) \cdot \nabla_{d_{uv}}^2 p_t \left( d_{uv} \right) \\
        &= - \nabla_{d_{uv}} \left[ 3 \mathbf{G}^2(t) / \| \mathcal{C}_u - \mathcal{C}_v \| \cdot p_t \left( d_{uv} \right) - \left( \mathbf{G}^2(t) - \sigma^2(t) \right) \nabla_{d_{uv}} p_t \left( d_{uv} \right) \right] + \sigma^2(t) \cdot \nabla_{d_{uv}}^2 p_t \left( d_{uv} \right) \\
        &= - \nabla_{d_{uv}} \left\{ \left[ 3 \mathbf{G}^2(t) / \| \mathcal{C}_u - \mathcal{C}_v \| - \left( \mathbf{G}^2(t) - \sigma^2(t) \right) \nabla_{d_{uv}} \log p_t \left( d_{uv} \right) \right]  p_t \left( d_{uv} \right) \right\} + \sigma^2(t) \cdot \nabla_{d_{uv}}^2 p_t \left( d_{uv} \right) 
    \end{align}
\end{subequations}

for any $\sigma(t) \leq \mathbf{G}(t)$. Note that the Eq.~\eqref{eq: FP for distance 2} has the solution of 

\begin{equation}
    \mathrm{d} d_{uv}^t = \left( 3 \mathbf{G}^2(t) / d_{uv}^t - \left( \mathbf{G}^2(t) - \sigma^2(t) \right) \nabla_{d_{uv}} \log p_t \left( d_{uv}^t \right) \right) \mathrm{d} t + \sqrt{2} \sigma(t) \mathrm{d} \mathcal{B}_t. \label{eq: forward sde of d 2}
\end{equation}

The Eq.~\eqref{eq: forward sde of d 1} and Eq.~\eqref{eq: forward sde of d 2} has the same marginal distribution. By setting $\sigma(t) = 0$ and ignoring the dependence of entries in the adjacency matrix, the reverse ODE flow is

\begin{equation}
    \mathrm{d} \gamma_1(t) = \left( 3 \mathbf{G}^2(t) / \gamma_1(t) - \mathbf{G}^2(t) \nabla_{\gamma_1} \log p_t \left( \gamma_1(t) \right) \right) \mathrm{d} t,
\end{equation}

which is the reverse adjacency matrix ODE flow of the Eq.~\eqref{eq: forward sde c ve case}. We can write the reverse ODE flow of the coordinates as 

\begin{subequations}
    \begin{align}
        d^{k-1} &= \pi_{M, d^k} \left( d^k + \tau \left( -3 \mathbf{G}^2(kT/N) / d^k + \mathbf{G}^2(kT/N) \nabla_{d} \log p_{kT/N} \left( d^k \right) \right) \right) \\
        \mathcal{C}^{k-1} &= \varphi \circ \pi_{M, d^k} \left( d^k + \tau \left( -3 \mathbf{G}^2(kT/N) / d^k + \mathbf{G}^2(kT/N) \nabla_{d} \log p_{kT/N} \left( d^k \right) \right) \right) \\
        &\simeq \mathcal{C}^k + \frac{\tau}{2 \left( n - 1 \right)} \left( -3 \mathbf{G}^2(kT/N) / d^k + \mathbf{G}^2(kT/N) \nabla_{d} \log p_{kT/N} \left( d^k \right) \right) \circledast \frac{\partial d^k}{\partial \mathcal{C}^k}
    \end{align}
\end{subequations}

Finally, the above discretization corresponds to the reverse ODE of 

\begin{equation}
    \frac{\mathrm{d} \mathcal{C}_t}{\mathrm{d} t} = \frac{\mathbf{G}^2(t)}{2 \left( n - 1 \right)} \left( \frac{3}{d_t} - \nabla_{d} \log p_t \left( d_t \right) \right) \circledast \frac{\partial d_t}{\partial \mathcal{C}_t}.
\end{equation}

\section{Error bound analysis} \label{app: scatter-mean approx error bound}
\subsection{Error bound computation}
We compute the maximum objective function value of $\sum_{i<j} \left( \left\| \hat{\mathbf{x}}_i - \hat{\mathbf{x}}_j \right\| - \hat{d}_{ij} \right)^2$ under our approximation in the case of $\hat{d} = \tilde{d} + \delta e_{uv}$ in Theorem~\ref{thm: error bound}. Let $\sim$ denote the adjacent relation and $S_N(i) = \{ (i, j) \mid \mathbf{x}_i \sim \mathbf{x}_j \}$ denotes the incident edges of node $i$, $S_{\emptyset}(i, j) = \{ (p, q) \mid p \not \in (i, j), q \not \in (i, j) \}$ denote the set of edges that are incident to $i$ and $j$, and $\operatorname{degree}_i$ denote the degree of node $i$. We use $\hat{\mathcal{C}} = [\hat{\mathbf{x}}_1, \ldots, \hat{\mathbf{x}}_n]$ to denote our approximated optimal solution, i.e. $\hat{\mathcal{C}} = \tilde{\mathcal{C}} + \frac{\delta}{2(n-1)} \frac{\partial \tilde{d}_{uv}}{\partial \tilde{\mathcal{C}}}$. Formally, we have 

\begin{subequations}
\begin{align}
    &\quad \quad \quad \quad \quad \quad \min_{\mathbf{x}_1, \ldots, \mathbf{x}_n} \sum_{i<j} \left( \left\| \mathbf{x}_i - \mathbf{x}_j \right\| - \hat{d}_{ij} \right)^2 \\
    &\leq \left( \left\| \hat{\mathbf{x}}_u - \hat{\mathbf{x}}_v \right\| - \hat{d}_{uv} \right)^2 + \sum_{(i, j) \in S_{\emptyset}(u, v)} \left( \left\| \hat{\mathbf{x}}_i - \hat{\mathbf{x}}_j \right\| - \hat{d}_{ij} \right)^2 \\
    & \quad + \sum_{(i, j) \in S_N(u) \setminus \{ (u, v) \} } \left( \left\| \hat{\mathbf{x}}_i - \hat{\mathbf{x}}_j \right\| - \hat{d}_{ij} \right)^2 + \sum_{(i, j) \in S_N(v) \setminus \{ (u, v) \} } \left( \left\| \hat{\mathbf{x}}_i - \hat{\mathbf{x}}_j \right\| - \hat{d}_{ij} \right)^2 \nonumber \\
    &= \left( \left\| (\tilde{\mathbf{x}}_u + \frac{\delta}{2 (n-1)} \boldsymbol{\lambda}_{uv}) - (\tilde{\mathbf{x}}_v - \frac{\delta}{2 (n-1)} \boldsymbol{\lambda}_{uv}) \right\| - \tilde{d}_{uv} - \delta \right)^2 \\
    & \quad + \sum_{(i, j) \in S_{\emptyset}(u, v)} \left( \left\| \tilde{\mathbf{x}}_i - \tilde{\mathbf{x}}_j \right\| - \tilde{d}_{ij} \right)^2 \nonumber \\
    & \quad + \sum_{(i, j) \in S_N(u) \setminus \{ (u, v) \} } \left( \left\| (\tilde{\mathbf{x}}_u + \frac{\delta}{2 (n-1)} \boldsymbol{\lambda}_{uv}) - \tilde{\mathbf{x}}_j \right\| - \tilde{d}_{uj} \right)^2 \nonumber \\
    & \quad + \sum_{(i, j) \in S_N(v) \setminus \{ (u, v) \} } \left( \left\| (\tilde{\mathbf{x}}_v - \frac{\delta}{2 (n-1)} \boldsymbol{\lambda}_{uv}) - \tilde{\mathbf{x}}_j \right\| - \tilde{d}_{vj} \right)^2 \nonumber \\
    &\leq \left(         \left\| \boldsymbol{\lambda}_{uv} \left( \left\| \tilde{\mathbf{x}}_u - \tilde{\mathbf{x}}_v \right\| + \frac{\delta}{n - 1} \right) \right\| - \tilde{d}_{uv} - \delta      \right)^2 + 0 \\
    & \quad + \max_{\pm_{(i, j)}} \sum_{(i, j) \in S_N(u) \setminus \{ (u, v) \} } \left( \left\| \tilde{\mathbf{x}}_u  - \tilde{\mathbf{x}}_j \right\| \pm \left\| \frac{\delta}{2 (n-1)} \boldsymbol{\lambda}_{uv} \right\| - \tilde{d}_{uj} \right)^2 \nonumber \\
    & \quad + \max_{\pm_{(i, j)}} \sum_{(i, j) \in S_N(v) \setminus \{ (u, v) \} } \left( \left\| \tilde{\mathbf{x}}_v - \tilde{\mathbf{x}}_j \right\| \pm \left\| \frac{\delta}{2 (n-1)} \boldsymbol{\lambda}_{uv} \right\| - \tilde{d}_{vj} \right)^2 \nonumber \\
    &= \left( 1 - \frac{1}{n-1} \right)^2 \delta^2 + \sum_{(i, j) \in S_N(u) \setminus \{ (u, v) \} } \left( \frac{\delta}{2 (n-1)} \right)^2 + \sum_{(i, j) \in S_N(v) \setminus \{ (u, v) \} } \left( \frac{\delta}{2 (n-1)} \right)^2 \\
    &= \left( \left( \frac{n-2}{n-1} \right)^2 + \frac{\operatorname{degree}_u - 1}{4(n-1)^2} + \frac{\operatorname{degree}_v - 1}{4(n-1)^2} \right) \delta^2 \\
    &\leq \left( \left( \frac{n-2}{n-1} \right)^2 + \frac{n - 2}{4(n-1)^2} + \frac{n - 2}{4(n-1)^2} \right) \delta^2 \\
    &= \frac{2n^2 - 7n + 6}{2 (n-1)^2} \delta^2 
\end{align}
\label{eq: error bound proof}
\end{subequations}

\subsection{Experiments of approximated optimal value}
\paragraph{Experiment settings.} We randomly generate $\mathcal{C}_i \sim \mathcal{N}(\mathbf{0}, \boldsymbol{I})$ with $n$ nodes, where $n \sim \operatorname{Uniform}([10, 19])$. We compute the adjacency matrix $d$ of the obtained coordinates and randomly perturb the adjacency matrix to obtain $\hat{d} = d + \delta \bar{e}_{uv}$. We aim to compare the magnitude of the optimal values computed under different algorithms or approximations. The optimal value is defined as the solution of the optimization problem $f(\hat{d}) = \underset{\hat{\mathbf{x}}_1, \ldots, \hat{\mathbf{x}}_n}{\min} \sum_{i<j} \left( \left\| \hat{\mathbf{x}}_i - \hat{\mathbf{x}}_j \right\| - \hat{d}_{ij} \right)^2$.

\begin{figure}[H]
    \centering
    \includegraphics[width=\linewidth]{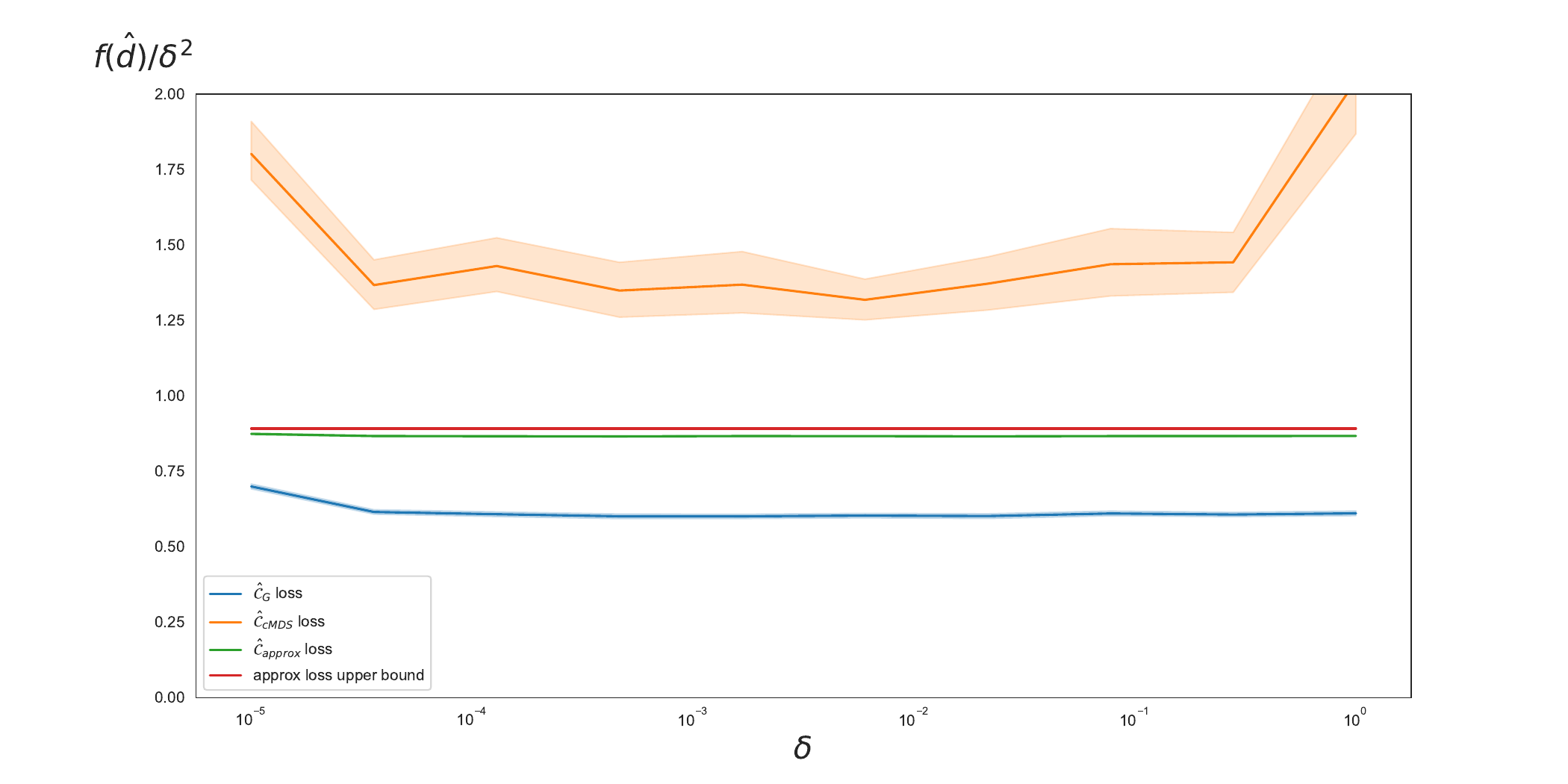}
    \caption{Optimization loss of MDS w.r.t. different optimization algorithm. $f(\hat{d})$ is the objective function of the optimization problem.}
    \label{fig: scatter-mean loss error comparison}
\end{figure}

\paragraph{Algorithms.} To our best knowledge, there is no simple algorithm for solving the metric MDS problem $\hat{\mathbf{x}}_1, \ldots, \hat{\mathbf{x}}_n = \underset{\hat{\mathbf{x}}_1, \ldots, \hat{\mathbf{x}}_n}{\operatorname{argmin}} \sum_{i<j} \left( \left\| \hat{\mathbf{x}}_i - \hat{\mathbf{x}}_j \right\| - \hat{d}_{ij} \right)^2$. The usual gradient descent algorithm is inapplicable in such case since $\| \mathbf{x} \|$ is not differentiable at $\mathbf{x} = \mathbf{0}$. Usual convergence theorems for gradient methods are invalid under such cases and local minimum points do not need to satisfy the stationary equations~\citep{de2005applications}. Hence, we slightly modify the gradient descent process to

\begin{equation}
    \hat{C}^{(t+1)} = \hat{C}^{(t)} + \nabla_{\hat{C}^{(t)}} \sum_{i<j} \left( \left( \left\| \hat{\mathbf{x}}_i^{(t)} - \hat{\mathbf{x}}_j^{(t)} \right\|^2 + \epsilon \right)^{1/2} - \hat{d}_{ij} \right)^2, \label{eq: mds gt gradient descent}
\end{equation}

for some sufficiently small $\epsilon > 0$. From experimental results, the above methods can monotonically decrease the loss when the step size is chosen properly and when $\hat{x}$ is not very close to the local minimum. We set $\hat{C}^{(0)} = \tilde{C} = [\tilde{\mathbf{x}}_1, \ldots, \tilde{\mathbf{x}}_n] $ as the initialized value and apply the gradient descent algorithm. We visualize the magnitude of the MDS objective function obtained at $\hat{\mathcal{C}}_{\textsubscript{approx}}, \hat{\mathcal{C}}_{\textsubscript{G}}, \hat{\mathcal{C}}_{\textsubscript{c-MDS}}$, computed from the proposed approximation (Eq.~\eqref{eq: approximate scatter mean compact}), gradient descent (Eq.~\eqref{eq: mds gt gradient descent}) and the algorithm for the classic MDS problem, respectively. We also visualize the error bound proved in Theorem~\ref{thm: error bound}. The results can be seen in Fig.~\ref{fig: scatter-mean loss error comparison}. The algorithm for the classic MDS problem (c-MDS) is stated below~\citep{wickelmaier2003introduction}:

\begin{itemize}
    \item Set up the squared proximity matrix $D =\left(d_{i j}^2\right)_{n \times n}$.
    \item Apply double centering: $B=-\frac{1}{2} P D P$ using the centering matrix $P=I-\frac{1}{n} J_n$, where $n$ is the number of objects, $I$ is the $n \times n$ identity matrix, and $J_n$ is an $n \times n$ matrix of all ones.
    \item Determine the 3 largest eigenvalues $\lambda_1, \lambda_2, \lambda_3$ and corresponding eigenvectors $e_1, e_2, e_3$ of $B$.
    \item Now, $\mathcal{C}=E_3 \Lambda_3^{1 / 2}$, where $E_3$ is the matrix of $3$ eigenvectors and $\Lambda_3$ is the diagonal matrix of $3$ eigenvalues of $B$. 
\end{itemize}

Metric MDS and c-MDS seek to find $\mathcal{C} = [\mathbf{x}_1, \ldots, \mathbf{x}_n]$ such that $\| \mathbf{x}_i - \mathbf{x}_j \| \approx d_{ij}$ but the optimal solution of c-MDS generally differs from the metric MDS. 

\paragraph{Experimental results.} We see that our approximation leads to a much smaller loss than the c-MDS's, and compared with the optimal value, the gap is acceptable. As there is still improvement for optimal values, this further suggests that applying a more refined projection operator can yield additional improvements in the model's performance~\citep{zhou2023learning}. 

\section{Corrector analysis} \label{sec: corrector analysis}
\subsection{Reasons for introduction of the corrector}
For simplicity, we consider the revised version of the reverse ODE given by Eq.~\eqref{eq: naive ODE in c}. Suppose that we have a model $\mathbf{s}_{\boldsymbol{\theta}} (d_t, t) = \frac{1}{2(n-1)} \nabla_d \log p_t\left(d_t\right) \circledast \frac{\partial d_t}{\partial \mathcal{C}_t} + \boldsymbol{\epsilon}(d_t, t)$, where $\boldsymbol{\epsilon}(d_t, t)$ is the prediction error from the approximation of 1) the mappings between the adjacency matrix, coordinates, 2) distance distribution, and 3) ignoration of dependence of distance entries. We assume $\boldsymbol{\epsilon}(d_t, t)$ follows a Gaussian distribution. Consider the discretization of reverse ODE between the timestamp from $s$ to $t$ (note that in variance exploding SDEs, $\mathbf{G}^2(t) = \frac{\mathrm{d} \sigma^2(t) }{\mathrm{d} t}$):

\begin{subequations}
\begin{align}
    \mathcal{C}_t &= \mathcal{C}_s - \int_{s}^{t} \frac{\mathbf{G}^2(u)}{2(n-1)} \nabla_d \log p_u\left(d_u\right) \circledast \frac{\partial d_u}{\partial \mathcal{C}_u} \mathrm{d} u \\
    &\approx \mathcal{C}_s + \left( \sigma^2(s) - \sigma^2(t) \right) \mathbf{s}_{\boldsymbol{\theta}} (d_s, s) + \left( \sigma^2(t) - \sigma^2(s) \right) \boldsymbol{\epsilon}(d_s, s)
\end{align}
\label{eq: reverse error from where}
\end{subequations}

The term $\left( \sigma^2(t) - \sigma^2(s) \right) \boldsymbol{\epsilon}(d_s, s)$ can be seen as an addtional noise injected to $\mathcal{C}_s$. Hence, after one iteration of the denoising process, the obtained $\tilde{C}_t$ does not exactly follow the distribution $q_t$ but follows $q_{t+\lambda(s-t)}$ for some $\lambda \in (0, 1)$. As the magnitude of the prediction error increases, the corresponding value of the parameter $\lambda$ exhibits a proportional augmentation. This motivates us to choose a larger drift. Hence, we introduce a multiplier ``corrector'' $k_{\mathbf{s}_{\boldsymbol{\theta}}}(d_s, s, t) > 1$ as an addition term to remedy the model's prediction errors, and the iteration rule becomes 

\begin{align}
    \mathcal{C}_t &\approx \mathcal{C}_s + k_{\mathbf{s}_{\boldsymbol{\theta}}}(d_s, s, t) \left( \sigma^2(s) - \sigma^2(t) \right) \mathbf{s}_{\boldsymbol{\theta}} (d_t, t). \label{eq: Rn conformation reverse approx}
\end{align}

\subsection{Experiments}
To find the corrector $k_{\mathbf{s}_{\boldsymbol{\theta}}}$, we further assume $k_{\mathbf{s}_{\boldsymbol{\theta}}}(d_s, s, t) \approx k_{\mathbf{s}_{\boldsymbol{\theta}}}(p_{\textsuperscript{data}})$, i.e., for each dataset, we can find a hyper-parameter to approximate $k_{\mathbf{s}_{\boldsymbol{\theta}}}(d_s, s, t)$ for all $d_t \sim p_{t}(d_t)$. From the experimental results, we find that the magnitude of $k_{\mathbf{s}_{\boldsymbol{\theta}}}$ increases along with the increase of the model's prediction error, while other factors such as node number and node degree do not have a significant influence on the choice of $k_{\mathbf{s}_{\boldsymbol{\theta}}}$. 

\subsubsection{Experiments for positive correlation between prediction error and corrector $k_{\mathbf{s}_{\theta}}$} \label{app: relations error and corrector settings}

\begin{wrapfigure}{r}{.33\textwidth}
\vspace{-20pt}
\captionsetup[subfigure]{labelformat=empty}
\centering
\includegraphics[width=0.33\textwidth]{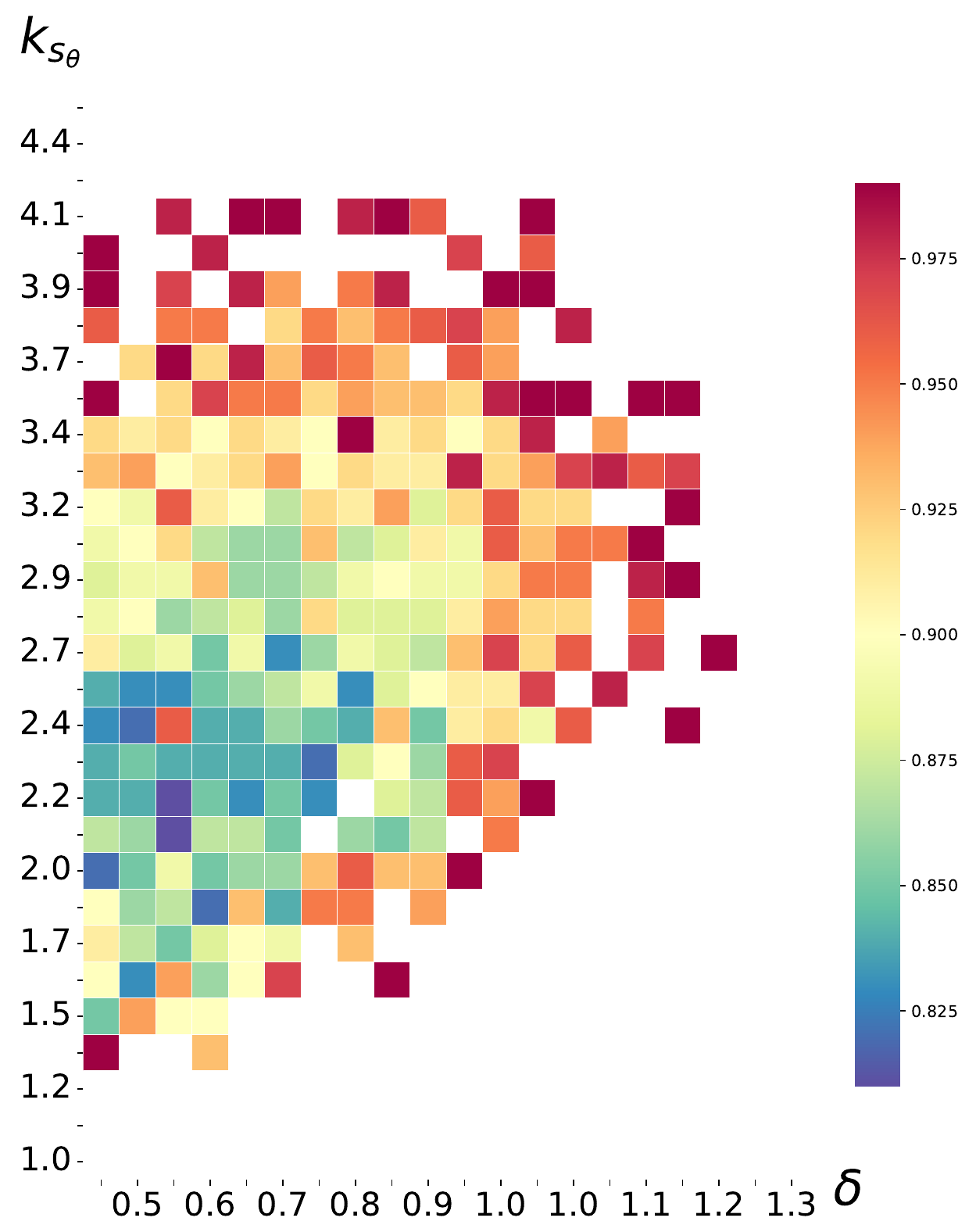}
\caption{Relations between the prediction error $\delta$ and corrector $k_{\mathbf{s}_{\theta}}$. Grid color indicates the convergent time.}
\label{fig: error and corrector}
\vspace{-30pt}
\end{wrapfigure}

\paragraph*{Dataset settings.} We develop datasets $Q_i = \{ \mathcal{C}_0^{(i)} \}$ that only contain one set of coordinates of 20 nodes. Each conformation is a 4-regular graph (so that SE(3)-invariant conformation and pairwise distance manifolds are bijective) and $\mathcal{C}_0^{(i)} \sim \mathcal{N}(\mathbf{0}, \mathbf{I}), \forall i = 1, \ldots, n$. Thus, the ground truth of the denoising process is fixed. We try to sample 3D coordinates in 100 steps.

\paragraph*{Model settings.} During the denoising process, our model has access to the ground truth $d_0$ and following the Gaussian assumption of the distance distribution, we want to develop a model $\mathbf{s}_{\boldsymbol{\theta}}(d_t, t) = \frac{1}{2\cdot\operatorname{degree}} \nabla_d \log p_t\left(d_t\right) \circledast \frac{\partial d_t}{\partial \mathcal{C}_t}$. As discussed in Eq.~\eqref{eq: reverse error from where}, we cannot access the accurate information about the denoising timestamp during the denoising process, hence, we assume that $\mathbf{s}_{\boldsymbol{\theta}}(d_t, t)$ has the standard deviation (std) to be 1. Such assumption comes from the fact that the standard deviation of the score matching ground truth $\frac{1}{2\cdot\operatorname{degree}} \nabla_d \log p_t\left(d_t \mid d_0 \right) \circledast \frac{\partial d_t}{\partial \mathcal{C}_t}$ approximately equals 1. Thus, the obtained model trends to output a score whose std equals 1. So, we force the output std of our model to be 1. Also, we use a hyperparameter $\delta$ to add prediction errors by letting $\mathbf{s}_{\boldsymbol{\theta}}(d_t, t, \delta) = \operatorname{Norm}_{\textsubscript{std}} \left( \frac{1}{2\cdot\operatorname{degree}} \nabla_d \log p_t\left(d_t\right) \circledast \frac{\partial d_t}{\partial \mathcal{C}_t} + \boldsymbol{\epsilon}^{d}(\delta, t) \right)$. But note that when $\delta = 0$, there are still prediction errors due to the approximated transformation between distances and coordinates and the inaccurate distance distribution hypothesis. Finally, we define $\boldsymbol{\epsilon}^{d}(\delta, t) = 2(\operatorname{sigmoid(\sigma_t \cdot \delta) - 0.5}) \delta \mathbf{z}$, where $\mathbf{z} \sim \mathcal{N}(\mathbf{0}, \mathbf{I})$.

\paragraph*{Detailed settings of convergence.} We find that even with access to the ground truth of $d_0$, our model cannot always converge to the ground truth at the end of the reverse process. Hence, we only consider the convergence for most samples (90\% samples). Given a noise level $\delta$ and $k_{\mathbf{s}_{\theta}}$, we repeatedly apply the denoising process. If at time $t_0$, at least 90\% samples have converged, then we say that under $\delta, k_{\mathbf{s}_{\theta}}$, the model converges at time $t_0$.

If $\| d_t - d_0 \|_{\infty} < h$ for some predefined threshold $h > 0$, we say that the reverse process converges at $t$ and define the minimal $1 - t / T$ to be the convergent time. We aim to find the convergent time under different noise levels $\delta$ and corrector $k_{\mathbf{s}_{\theta}}$. We grid-search the above two parameters and visualize the convergence of the model. The results can be seen in Fig.~\ref{fig: error and corrector} and the color of each grid represents the convergent time. Grids with no color imply that under such noise level $\delta$ and model error $k_{\mathbf{s}_{\theta}}$, the model diverges. We can see a positive correlation between the model's error and the corrector, which matches the hypothesis in Eq.~\eqref{eq: reverse error from where}.

\subsubsection{Ablation study on factors influencing the value of the corrector} \label{app: relations other factors and corrector}

In the context of our investigation, we employ $\| d_t - d_0\|_{\infty}$ as a reduced-dimensional representation of the reverse flow phenomenon, where $d_t$ denotes the edge lengths of the graph at time $t$, $d_0$ denotes the ground truth edge lengths and $\| \cdot \|_{\infty}$ denotes the maximum difference. It is well-established that the selection of the corrector $k_{\mathbf{s}_{\theta}}$ exerts a notable influence on the characteristics of the flow. Our primary objective is to systematically investigate whether while maintaining a constant corrector parameter $k_{\mathbf{s}_{\theta}}$, other variables such as the number of nodes and node degrees exhibit substantial alterations in the reverse flow patterns. Should our findings indicate minimal variation in the reverse flow with respect to these aforementioned factors, it would enable us to posit that the choice of corrector $k_{\mathbf{s}_{\theta}}$ is relatively independent of their influence. We use the same settings of the sigma scheduler and the model, as discussed in the Appendix~\ref{app: relations error and corrector settings} but develop different toy datasets that contain graphs of distinct node numbers and distinct node degrees. 

\begin{figure}[h]
\centering
\subfigure[node number v.s. corrector]{\includegraphics[width=.25\textwidth] {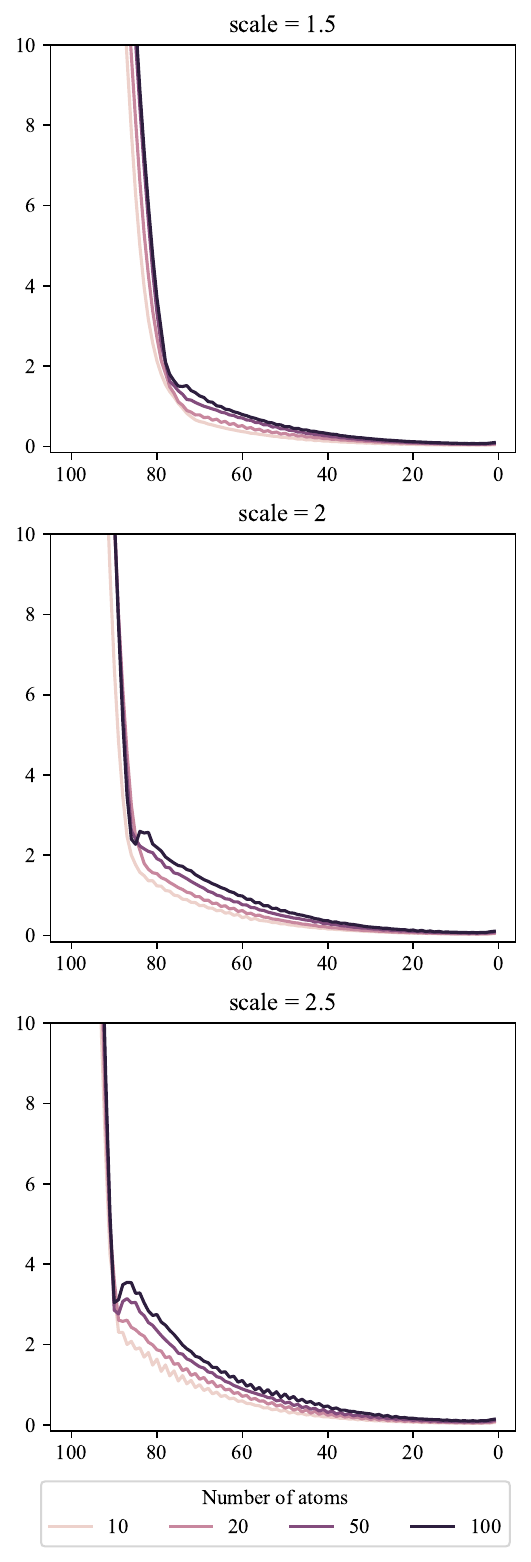} \label{fig: ablation-atom}}
\subfigure[node degree v.s. corrector]{\includegraphics[width=.25\textwidth] {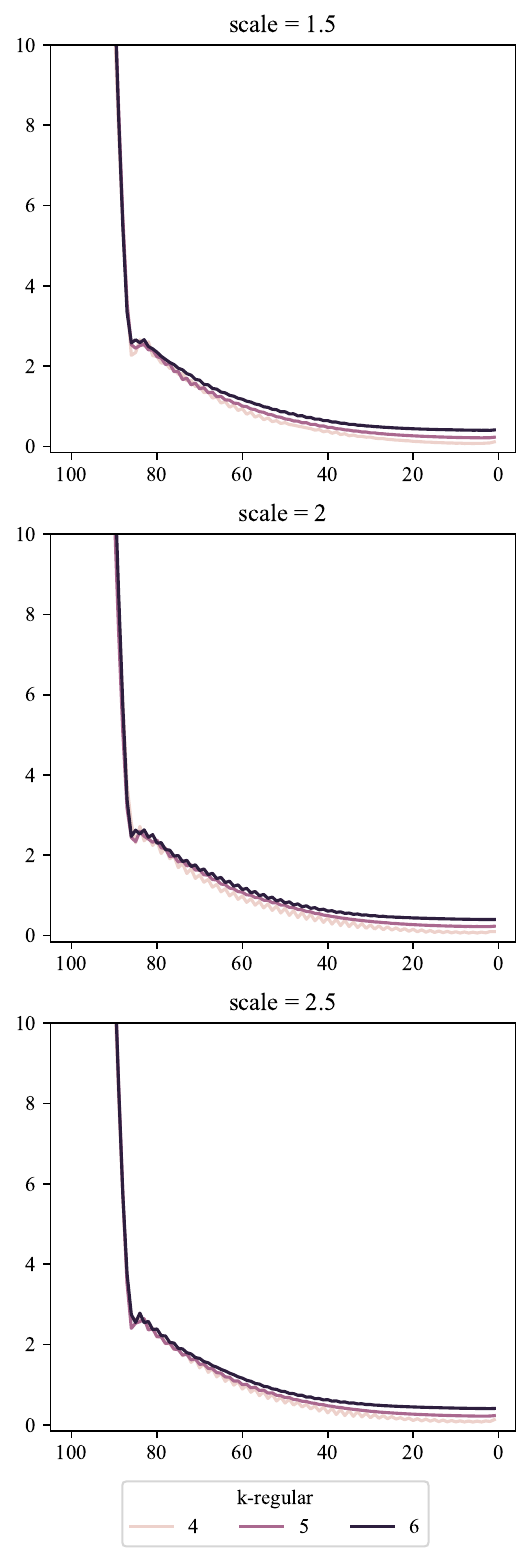} \label{fig: ablation-k}}
\subfigure[model error v.s. corrector]{\includegraphics[width=.25\textwidth] {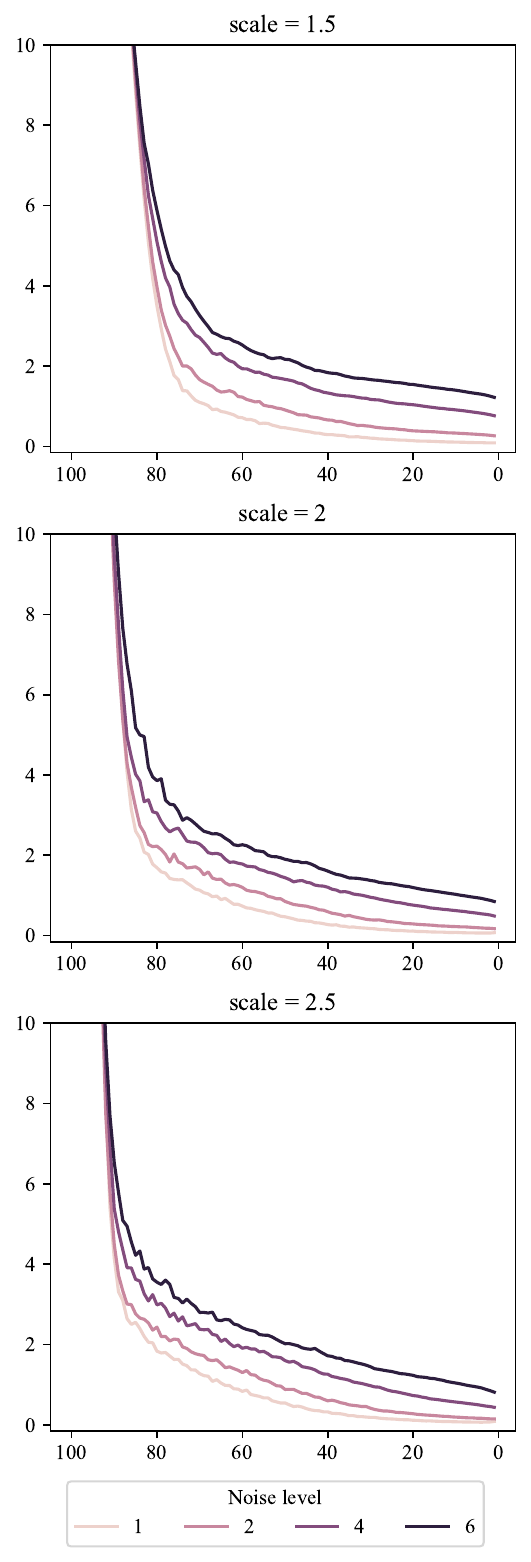} \label{fig: ablation-noise}}

\caption{Ablation study of factors that may affect the corrector $k_{\mathbf{s}_{\theta}}$. The y-axis denotes the flow represented by $\| d_t - d_0\|_{\infty}$ and the x-axis denotes the denoising steps $t$. We can see that the prediction error greatly affects the reverse flow and other factors do not have a significant impact on the reverse flow.}
\label{fig: ablation}
\end{figure}

\paragraph{Dataset settings.} We develop two datasets with one containing a fixed node number of [10, 20, 50, 100], and each graph is a 4-regular graph and the other contains conformations of 20 nodes, and each graph is a $k$-regular graph, where $k=4, 5, 6$. 

\paragraph{Experimental results.} We visualize the flow (represented by $\| d_t - d_0\|_{\infty}$) under different node numbers and node degrees and find that these two factors have a limited effect on the flow when the corrector $k_{\mathbf{s}_{\theta}}$ is fixed. Results can be seen in Fig.~\ref{fig: ablation-atom} and Fig.~\ref{fig: ablation-k}. We also visualize the reverse flow of 4-regular graphs with 20 nodes under different prediction errors. Results are shown in Fig.~\ref{fig: ablation-noise}. Compared with Fig.~\ref{fig: ablation-atom} and Fig.~\ref{fig: ablation-k}, we can see that the flow under different settings of the model's error shows great difference and we conclude that prediction error is the main factor that affects the choice of corrector $k_{\mathbf{s}_{\theta}}$.

\section{Experiment extension} \label{app: experiment extension}
In this section, we introduce training details first, and then provide more analysis and results for molecular conformation generation task and human pose generation task, respectively.

\subsection{Training details} \label{app: training details}
\paragraph{Backbone.}
Our two tasks can both be viewed as generating 3D coordinates based on a given graph. A molecule is natural to be represented as a graph. Notably, we add ``virtual edges" between nodes that are not originally bonded. Note that we find that embedding time causes a decrease in performance. Hence, we do not embed the diffusion time stamp in the model.

For the molecular generation task, the molecules can be viewed as both local and global components. Same as~\citet{xu2022geodiff}, we utilize GIN~\citep{xu2018powerful} to process the local part of the graph and SchNet~\citep{schutt2017schnet} for the global part. These two models can ensure Permutation Invariance and Graph Isomorphism, which can be beneficial to capturing intricate dependencies within complex graph data.

\begin{figure}[tb]
    \centering
    \includegraphics[width=.2\linewidth]{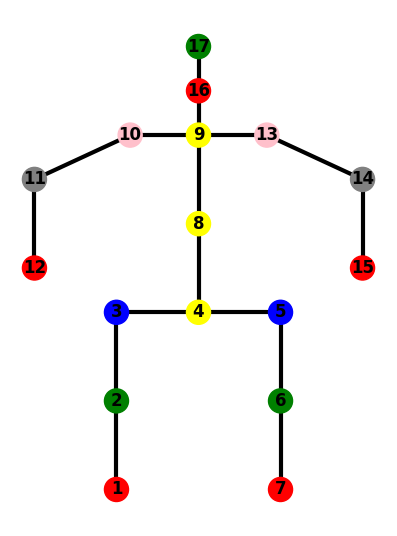} \includegraphics[width=.2\linewidth]{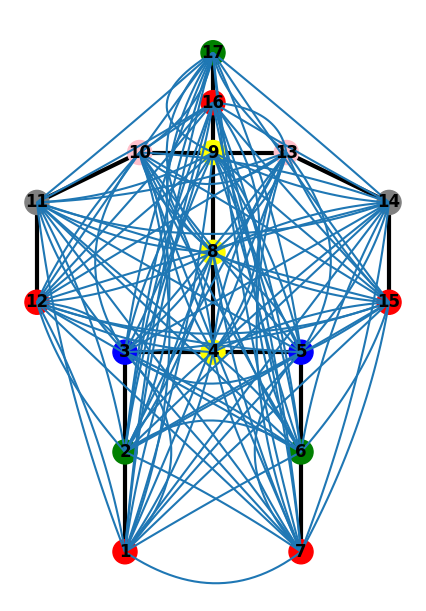}
    \caption{We conceptualize human pose as a graph where nodes represent key points, and edges represent human limbs shown as the left figure; we add ``virtual edges" to guarantee full connectivity within the graph shown as the right figure where the blue edges are new-added ``virtual edges".}
    \label{fig: human-skeleton}
\end{figure}

For the human pose generation task, we conceptualize human pose as a graph where nodes represent key points, and edges signify human limbs. Each node contains 3D coordinates. We perform a normalization of 3D coordinates and then introduce ``virtual edges" to guarantee full connectivity within the graph. The graph This adjustment is essential since we consider that each particle requires at least four connections to determine its position, as illustrated in Fig.~\ref{fig: human-skeleton}. 

Since the backbone we use is a graph neural network, we believe that it is beneficial to add additional edges to the original graph, which not only provides more global information for local nodes but also can better generate high-quality samples with more stability due to the strong connectivity between nodes.

\paragraph{Configuration.}
We trained the model on a single NVIDIA GeForce RTX 3090 GPU and Intel(R) Xeon(R) Silver 4210R CPU @ 2.40GHz CPU. We use the Adam optimizer for training, with a maximum of 1000 epochs. We set the learning rate to 0.001, and it decreases by a factor of 0.6 every 2 epochs. The batch size is 64. As for diffusion configuration, we set the steps of the forward diffusion process to 5000, the noise scheduler to $\sigma_t=\sqrt{\frac{\bar{\alpha}_t}{1 - \bar{\alpha}_t}}$ where $\bar{\alpha}_t=\prod_i^t (1-\beta_i)$ and $\beta_t=\operatorname{sigmoid}(t)$. $t$ is uniformly selected in [1e-7, 2e-3], and the magnitude of $\sigma_t$ is in the range of 0 to around 12.

\subsection{Molecular conformation generation}

\subsubsection{Evaluation} \label{app: geom experiment setting}
We use the COV and MAT metrics for Recall (R) and Precision (P) to evaluate the diversity and quality of generated conformers. These metrics are built on the root-mean-square-deviation (RMSD) of heavy atoms. COV can reflect the converge status of ground truth conformers, and MAT denotes the average minimum RMSD. The calculation for COV-R and MAT-R is:
\begin{equation*}
    \operatorname{COV-R} =\frac{1}{|S_r|} \{\mathcal{C} \in S_r | \operatorname{RMSD} (\mathcal{C}, \mathcal{C}')<\tau,\exists \mathcal{C}' \in S_g\}, \quad
    \operatorname{MAT-R} =\frac{1}{|S_r|} \sum_{\mathcal{C} \in S_r} \min_{\mathcal{C}' \in S_g} \operatorname{RMSD} (\mathcal{C}, \mathcal{C}')
\end{equation*}
where $S_g$ and $S_r$ denote generated and ground truth conformations. Sweeping $S_g$ and $S_r$, we obtain the COV-P and MAT-P. The MAT is calculated under RMSD threshold $\tau$. Following previous work~\citep{xu2020learning, xu2022geodiff}, we set $\tau=0.5\SI{}{\angstrom}$ for GEOM-QM9 and $\tau=1.25\SI{}{\angstrom}$ for GEOM-Drugs. Different from the previous work, when iterations for sampling are reduced, there may be nan in the generated samples, resulting in MAT metric exploding. Hence, when a sample contains nans, we use another sample from the same conditional generation process to replace it.

\subsubsection{Hyper-parameter analysis} \label{app: hyperparameter of molecule}
In Section \ref{sec: corrector analysis}, a hyper-parameter known as the ``corrector'', denoted as $k_{\mathbf{s}_{\theta}}$, is introduced. In this section, we report the performance implications associated with varying $k_{\mathbf{s}_{\theta}}$ values. The results for different $k_{\mathbf{s}_{\theta}}$ values are displayed in Table \ref{tab: corrector-Drugs}. Analysis of these results indicates satisfactory performance for $k_{\mathbf{s}_{\theta}}$ values ranging approximately from 8 to 16, for both the QM9 and Drugs datasets, suggesting a relative insensitivity of our proposed SDE and ODE to variations in the corrector values.

\begin{table}[h]
	\centering 
	\caption{Influence of corrector values on our proposed SDE and ODE in Drugs and QM9 dataset. Higher values for COV indicate that the model can generate samples with higher diversity. Lower values for MAT imply the model can generate high-quality samples with lower differences compared with the ground truth samples.}
	\label{tab: corrector-Drugs}

    \subtable[SDE on Drugs dataset]{
        \begin{sc}
        \begin{tabular}{>{\centering\arraybackslash}p{.9cm}p{1.8cm}|*4{>{\centering\arraybackslash}p{1.2cm}}|*4{>{\centering\arraybackslash}p{1.2cm}}}
    	\toprule
    	\multirow{2}{*}{\# stpes} &\multirow{2}{*}{Corrector} & \multicolumn{2}{c}{COV-R(\%) $\uparrow$} & \multicolumn{2}{c|}{MAT-R($\SI{}{\angstrom}$) $\downarrow$} & \multicolumn{2}{c}{COV-P(\%) $\uparrow$}& \multicolumn{2}{c}{MAT-P($\SI{}{\angstrom}$) $\downarrow$} \\ 
    	& & Mean & Median & Mean & Median & Mean & Median & Mean & Median \\ 
    	\midrule
    
         \multirow{4}{*}{100} 
        	& 4 & 60.55 & 68.13 & 1.1862 & 1.1758 & 27.38 & 21.47 & 1.5511 & 1.4989\\ 
        	& 8 & 85.13 & 94.06 & 0.9636 & 0.9518 & 56.26 & 55.46 & 1.2575 & 1.2224\\ 
        	& 16 & \textbf{86.44} & \textbf{93.08} & \textbf{0.9261} & \textbf{0.9132} & \textbf{64.53} & \textbf{67.83} & \textbf{1.1584} & \textbf{1.1279}\\ 
        	& 20 & \textbf{84.78} & \textbf{92.02} & \textbf{0.9332} & \textbf{0.9280} & \textbf{66.50} & \textbf{71.93} & \textbf{1.1426} & \textbf{1.0990}\\ 
        
        \midrule
        \midrule
        
        \multirow{4}{*}{1000} 
         	& 4 & 68.08 & 78.22 & 1.1421 & 1.1253 & 33.20 & 26.87 & 1.4961 & 1.4524\\ 
        	& 8 & \textbf{83.39} & \textbf{91.02} & \textbf{0.9728} & \textbf{0.9784} & \textbf{58.43} & \textbf{60.28} & \textbf{1.2393} & \textbf{1.1814}\\ 
        	& 16 & \textbf{75.26} & \textbf{80.65} & \textbf{1.0069} & \textbf{1.0058} & \textbf{64.68} & \textbf{70.24} & \textbf{1.1749} & \textbf{1.1030}\\ 
        	& 20 & 74.11 & 78.58 & 1.0167 & 1.0125 & 65.77 & 71.82 & 1.1578 & 1.0916\\ 
        
    	\bottomrule
    	\end{tabular}
     \end{sc}

    }

    \subtable[ODE on Drugs dataset]{
    \begin{sc}
        \begin{tabular}{>{\centering\arraybackslash}p{.9cm}p{1.8cm}|*4{>{\centering\arraybackslash}p{1.2cm}}|*4{>{\centering\arraybackslash}p{1.2cm}}}
    	\toprule
     
        \multirow{5}{*}{100} 
            & 2 & 77.90 & 93.73 & 1.0293 & 1.0144 & 41.19 & 37.28 & 1.4281 & 1.3852\\ 
        	& 4 & \textbf{87.76} & \textbf{98.59} & \textbf{0.9105} & \textbf{0.8846} & \textbf{53.32} & \textbf{54.27} & \textbf{1.8528} & \textbf{1.3563}\\ 
        	& 8 & \textbf{85.54} & \textbf{94.77} & \textbf{0.9201} & \textbf{0.9011} & \textbf{61.23} & \textbf{67.43} & \textbf{2.1393} & \textbf{1.5605}\\ 
        	& 16 & 85.06 & 93.76 & 0.9272 & 0.9099 & 60.72 & 67.62 & 1.8572 & 1.4149\\ 
        	& 20 & 85.46 & 93.28 & 0.9229 & 0.9064 & 62.70 & 69.27 & 1.9726 & 1.3828\\ 

            \midrule
            \midrule

        \multirow{5}{*}{1000} 
            & 4 & 87.07 & 98.76 & 0.9110 & 0.9006 & 52.80 & 52.59 & 1.2756 & 1.2319\\ 
        	& 8 & \textbf{90.64} & \textbf{98.85} & \textbf{0.8493} & \textbf{0.8270} & \textbf{59.67} & \textbf{62.12} & \textbf{1.1915} & \textbf{1.1420}\\ 
        	& 16 & \textbf{89.46} & \textbf{95.39} & \textbf{0.8407} & \textbf{0.8277} & \textbf{65.27} & \textbf{69.49} & \textbf{1.1170} & \textbf{1.0634}\\ 
        	& 20 & 88.03 & 94.08 & 0.8587 & 0.8483 & 66.82 & 70.82 & 1.1012 & 1.0612\\ 
    	\bottomrule
    	\end{tabular} 
    \end{sc}
     }
\end{table}

\begin{table}[tb] 
	\centering 
	\label{tab: corrector-QM9}

    \subtable[SDE on QM9 dataset]{
    \begin{sc}
         \begin{tabular}{>{\centering\arraybackslash}p{.9cm}p{1.8cm}|*4{>{\centering\arraybackslash}p{1.2cm}}|*4{>{\centering\arraybackslash}p{1.2cm}}}
    	\toprule
        \multirow{4}{*}{100} 
        	& 4 & 52.08 & 51.12 & 0.4928 & 0.4932 & 14.84 & 12.88 & 0.7506 & 0.7084\\ 
        	& 8 & \textbf{89.83} & \textbf{92.58} & \textbf{0.2858} & \textbf{0.2894} & \textbf{53.21} & \textbf{51.24} & \textbf{0.4957} & \textbf{0.4720}\\ 
        	& 16 & \textbf{88.34} & \textbf{91.59} & \textbf{0.2763} & \textbf{0.2757} & \textbf{53.48} & \textbf{51.25} & \textbf{0.4870} & \textbf{0.4564}\\ 
        	& 20 & 86.98 & 91.06 & 0.2928 & 0.2854 & 52.62 & 49.90 & 0.6315 & 0.4828\\ 
        
        \midrule
        \midrule
        
        \multirow{4}{*}{1000} 
         	& 4 & 64.09 & 65.42 & 0.4614 & 0.4617 & 21.14 & 18.62 & 0.6938 & 0.6605\\ 
        	& 8 & 90.03 & 94.65 & 0.2623 & 0.2706 & 54.17 & 52.12 & 0.4781 & 0.4617\\ 
        	& 16 & \textbf{89.25} & \textbf{94.02} & \textbf{0.2215} & \textbf{0.2180} & \textbf{55.15} & \textbf{52.42} & \textbf{0.4323} & \textbf{0.4164}\\ 
        	& 20 & \textbf{88.21} & \textbf{90.16} & \textbf{0.2251} & \textbf{0.2229} & \textbf{55.81} & \textbf{53.99} & \textbf{0.4273} & \textbf{0.4095}\\ 
        
    	\bottomrule
    	\end{tabular}   
    \end{sc}

     }
	\subtable[ODE on QM9 dataset]{
     \begin{sc}
        \begin{tabular}{>{\centering\arraybackslash}p{.9cm}p{1.8cm}|*4{>{\centering\arraybackslash}p{1.2cm}}|*4{>{\centering\arraybackslash}p{1.2cm}}}
    	\toprule
        \multirow{5}{*}{100} 
            & 2 & 87.03 & 92.32 & 0.3269 & 0.3322 & 42.86 & 41.65 & 0.6528 & 0.5582\\ 
        	& 4 & \textbf{90.81} & \textbf{96.56} & \textbf{0.2691} & \textbf{0.2765} & \textbf{48.73} & \textbf{47.16} & \textbf{0.5190} & \textbf{0.5026}\\ 
        	& 8 & \textbf{89.75} & \textbf{94.19} & \textbf{0.2699} & \textbf{0.2713} & \textbf{48.98} & \textbf{47.57} & \textbf{0.5140} & \textbf{0.4902}\\ 
        	& 16 & 85.50 & 90.58 & 0.2928 & 0.2945 & 48.57 & 47.09 & 0.5067 & 0.4906\\ 
        	& 20 & 81.58 & 86.64 & 0.3190 & 0.3205 & 47.08 & 45.98 & 0.5211 & 0.5035\\ 

            \midrule
            \midrule
            
        \multirow{4}{*}{1000} 
        	& 4 & \textbf{90.24} & \textbf{95.04} & \textbf{0.2052} & \textbf{0.1956} & \textbf{53.76} & \textbf{52.26} & \textbf{0.4304} & \textbf{0.4251}\\ 
        	& 8 & \textbf{88.99} & \textbf{93.49} & \textbf{0.2139} & \textbf{0.2037} & \textbf{54.41} & \textbf{53.60} & \textbf{0.4230} & \textbf{0.4099}\\ 
        	& 16 & 78.89 & 82.31 & 0.2790 & 0.2680 & 54.14 & 52.56 & 0.7337 & 0.4345\\ 
        	& 20 & 69.17 & 70.73 & 0.3303 & 0.3341 & 54.64 & 52.76 & 0.8217 & 0.4242\\ 
        	\bottomrule
    	\end{tabular} 
     \end{sc}
        
     }

\end{table}

To facilitate a fair comparison with our approach and ensure robust validation, we have conducted a thorough examination of the optimal step size $\alpha_t$, as delineated in Eq.~\ref{eq: LD sample}, for the baseline method, LD. Detailed results of the LD sampling over 100 steps on the QM9 dataset are presented in Table \ref{tab: QM9 GeoDiff ld 100}. Additionally, the results for LD sampling over 5000 and 1000 steps, as reported in the GeoDiff study, are included for reference in Table \ref{tab: GEOM}.

\begin{table}[tb] 
	\centering 
	\caption{Influence of step size variation on sampling via LD over 100 steps in QM9 dataset.}
	\label{tab: QM9 GeoDiff ld 100}
     \begin{sc}
        \begin{tabular}{c|cccc|cccc}
    	\toprule
    	\multirow{2}{*}{Step size} & \multicolumn{2}{c}{COV-R(\%) $\uparrow$} & \multicolumn{2}{c|}{MAT-R($\SI{}{\angstrom}$) $\downarrow$} & \multicolumn{2}{c}{COV-P(\%) $\uparrow$}& \multicolumn{2}{c}{MAT-P($\SI{}{\angstrom}$) $\downarrow$} \\ 
    	& Mean & Median & Mean & Median & Mean & Median & Mean & Median \\ 
    	\midrule
    	3e-7  & 16.57 & 09.34 & 0.6573 & 0.6318 & 03.19 & 01.75 & 1.6711 & 1.5551\\ 
    	4e-7  & 76.70 & 81.61 & 0.3998 & 0.3947 & 31.52 & 27.16 & 0.6954 & 0.6308\\ 
    	5e-7  & \textbf{88.40} & \textbf{91.70} & 0.3135 & 0.3120 & \textbf{48.84} & \textbf{47.00} & 0.5411 & 0.5024\\ 
    	6e-7  & \textbf{88.88} & \textbf{91.83} & 0.2898 & 0.2962 & \textbf{53.48} & \textbf{51.64} & 0.5051 & 0.4663\\ 
    	7e-7  & \textbf{88.24} & \textbf{91.08} & \textbf{0.2787} & \textbf{0.2898} & \textbf{54.61} & \textbf{52.00} & \textbf{0.4932} & \textbf{0.4581}\\ 
    	8e-7  & \textbf{88.07} & \textbf{90.20} & \textbf{0.2711} & \textbf{0.2808} & \textbf{54.80} & \textbf{52.42} & \textbf{0.5090} & \textbf{0.4552}\\ 
    	9e-7  & 87.22 & 89.42 & \textbf{0.2695} & \textbf{0.2763} & 54.87 & 51.73 & \textbf{0.5364} & \textbf{0.4600}\\ 
    	1e-6  & 87.10 & 90.73 & \textbf{0.2663} & \textbf{0.2759} & 54.75 & 53.01 & \textbf{0.5576} & \textbf{0.4667}\\ 
    	2e-6  & 83.00 & 87.88 & 0.2849 & 0.2858 & 49.91 & 45.83 & 1.5991 & 0.6705\\ 
    	3e-6  & 75.22 & 79.81 & 0.3460 & 0.3323 & 37.64 & 33.92 & - & -\\ 
    	4e-6  & 65.99 & 73.94 & - & 0.3647 & 31.54 & 26.74 & - & -\\ 
    	\bottomrule
    	\end{tabular}
     \end{sc}
    
\end{table}

\clearpage
\subsubsection{Visualization} \label{app: visualizaiton of molecule}
\begin{figure}[h]
    \centering
    \subfigure[QM9]{
        \tabcolsep=0.09cm 
        \hspace{-6em}
        \begin{tabular}{p{1.4cm}p{13cm}}
        \makecell{Method \\ / Steps}  \\
        \makecell{SDE\\1000} & \raisebox{-.4\totalheight}{\includegraphics[width=.9\textwidth]{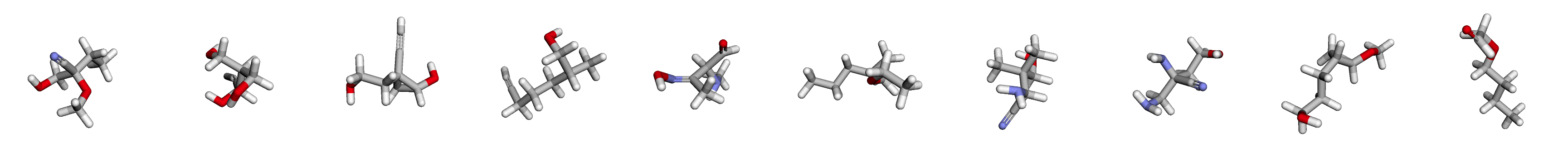}} \\
        \makecell{SDE\\100} & \raisebox{-.4\totalheight}{\includegraphics[width=.9\textwidth]{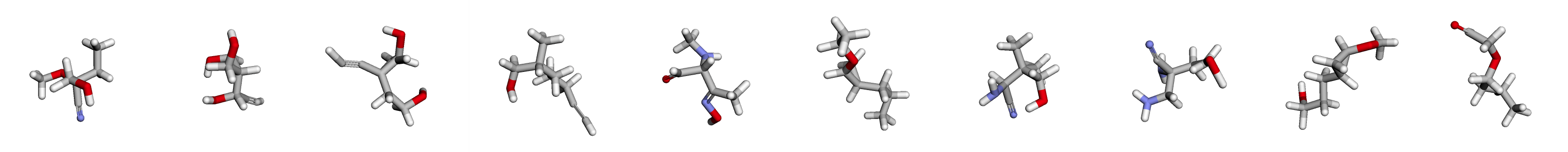}} \\
        \makecell{ODE\\1000} &  \raisebox{-.4\totalheight}{\includegraphics[width=.9\textwidth]{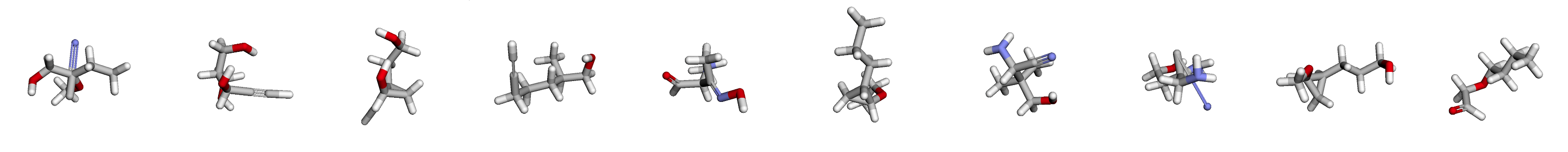}}\\
        \makecell{ODE\\100} & \raisebox{-.4\totalheight}{\includegraphics[width=.9\textwidth]{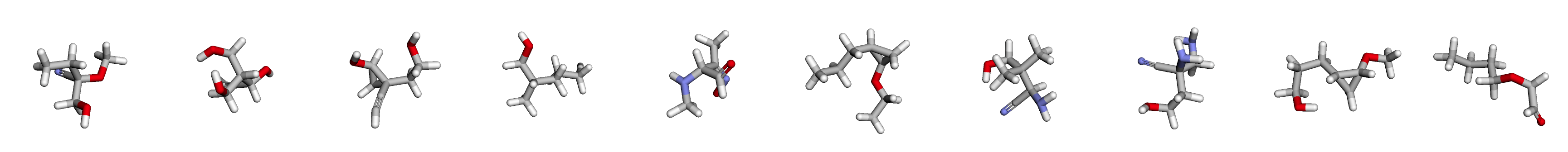}} \\
        & \raisebox{-.4\totalheight}{\includegraphics[width=.9\textwidth]{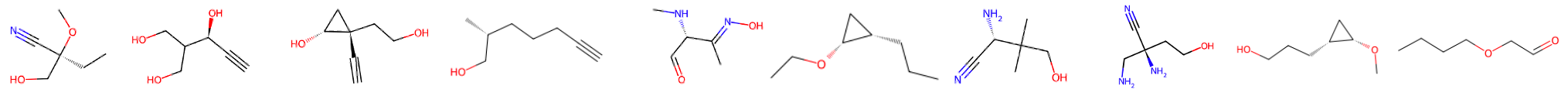}} \\
        \end{tabular}
    }
    \subfigure[Drugs]{
        \tabcolsep=0.09cm 
        \hspace{-6em}
        \begin{tabular}{p{1.4cm}p{13cm}}
        \makecell{SDE\\1000} & \raisebox{-.4\totalheight}{\includegraphics[width=.9\textwidth]{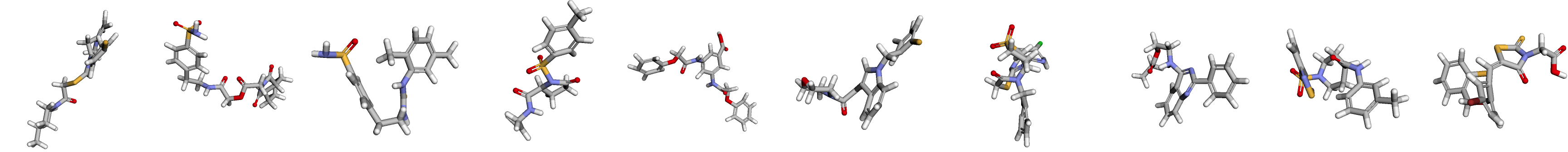}} \\
        \makecell{SDE\\100} & \raisebox{-.4\totalheight}{\includegraphics[width=.9\textwidth]{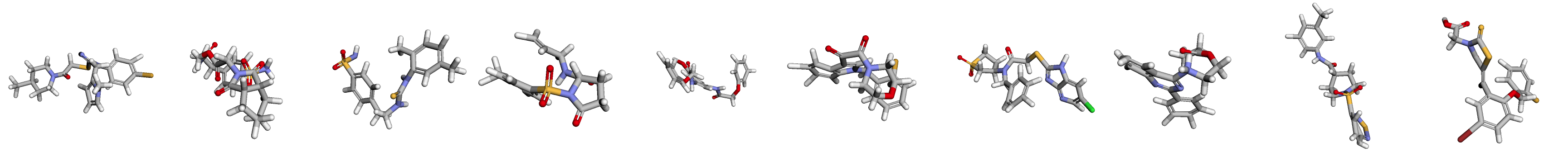}} \\
        \makecell{ODE\\1000} &  \raisebox{-.4\totalheight}{\includegraphics[width=.9\textwidth]{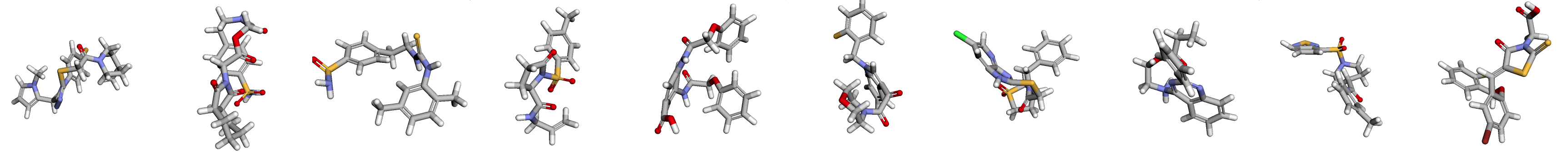}} \\
        \makecell{ODE\\100} & \raisebox{-.4\totalheight}{\includegraphics[width=.9\textwidth]{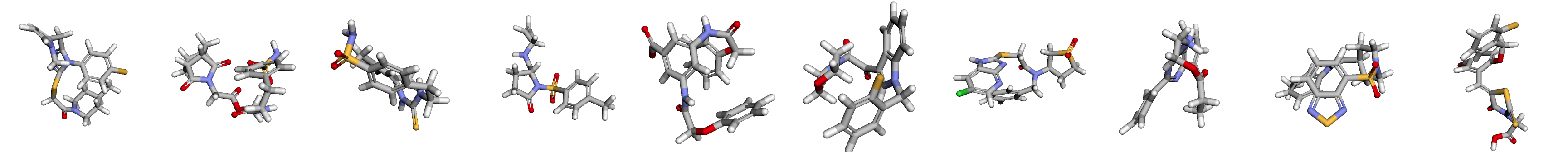}} \\
         & \raisebox{-.4\totalheight}{\includegraphics[width=.9\textwidth]{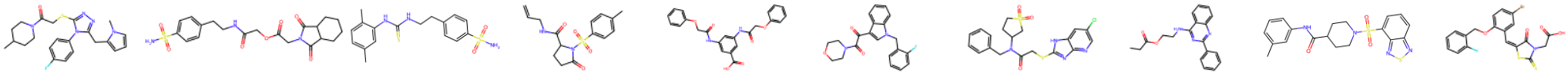}} \\
        \end{tabular}
        }
    \caption{Examples of molecular conformers sampled using our proposed SDE and ODE methods. Accompanying each example are the corresponding molecular formulas for reference.}
    \label{fig: vis-mol}
\end{figure}

For qualitative analysis, we visualized conformers generated using our methods over 100 and 1000 steps, as depicted in Fig.~\ref{fig: vis-mol}. These visualizations demonstrate that most molecular structures sampled through our approach are both meaningful and stable, even with a limited step count of only 100.

\begin{figure}[h]
    \centering
    \subfigure[QM9]{
        \tabcolsep=0.09cm 
        \hspace{-6em}
        \begin{tabular}{p{1.4cm}p{13cm}}
        \makecell{Methods \\/ Steps} & $\xrightarrow{\text{{\small Noise}}\hspace*{13cm}\text{{\small Molecule}}}$ \\
        \makecell{SDE\\1000} & \raisebox{-.4\totalheight}{\includegraphics[width=.9\textwidth]{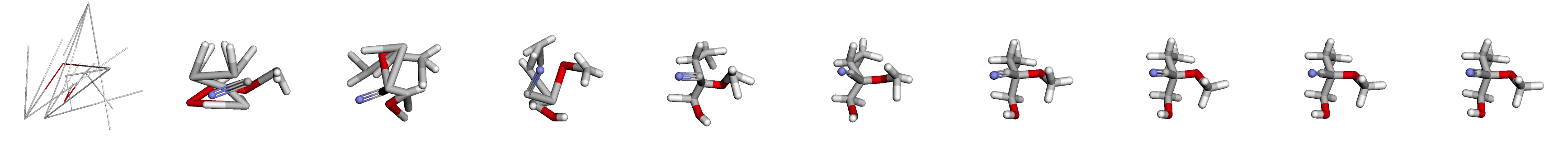}} \\
        \makecell{SDE\\100} & \raisebox{-.4\totalheight}{\includegraphics[width=.9\textwidth]{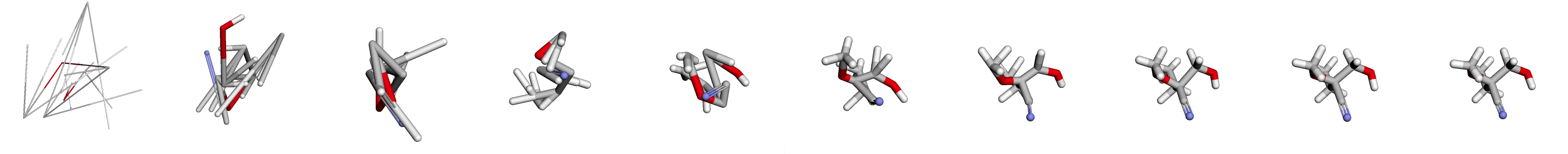}} \\
        \makecell{ODE\\1000} & \raisebox{-.4\totalheight}{\includegraphics[width=.9\textwidth]{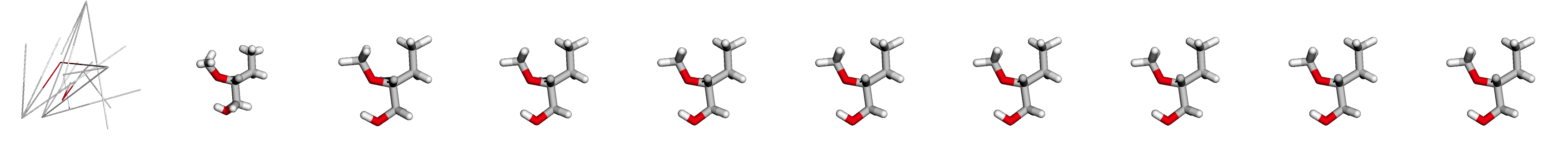}} \\
        \makecell{ODE\\100} & \raisebox{-.4\totalheight}{\includegraphics[width=.9\textwidth]{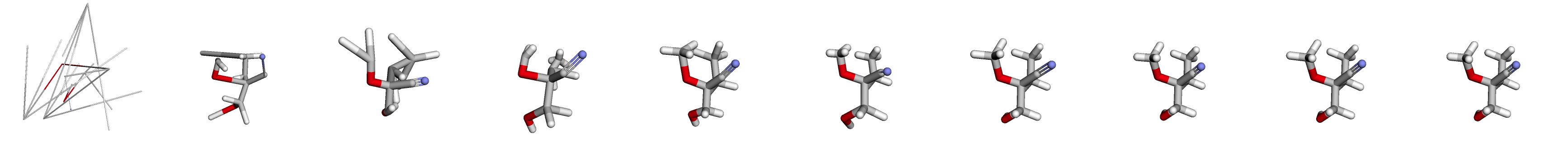}} \\
        \end{tabular}
    }

    \subfigure[Drugs]{
        \tabcolsep=0.09cm 
        \hspace{-6em}
        \begin{tabular}{p{1.4cm}p{13cm}}
        \makecell{SDE\\1000} & \raisebox{-.4\totalheight}{\includegraphics[width=.9\textwidth]{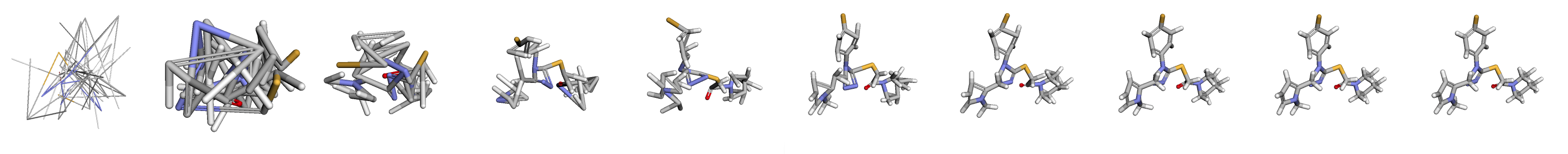}}\\
        \makecell{SDE\\100} & \raisebox{-.4\totalheight}{\includegraphics[width=.9\textwidth]{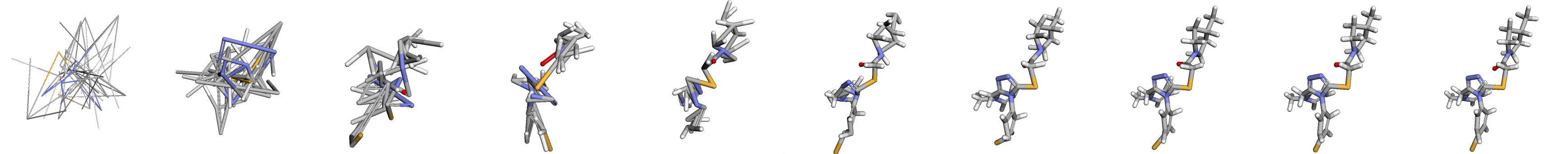}} \\
        \makecell{ODE\\1000} & \raisebox{-.4\totalheight}{\includegraphics[width=.9\textwidth]{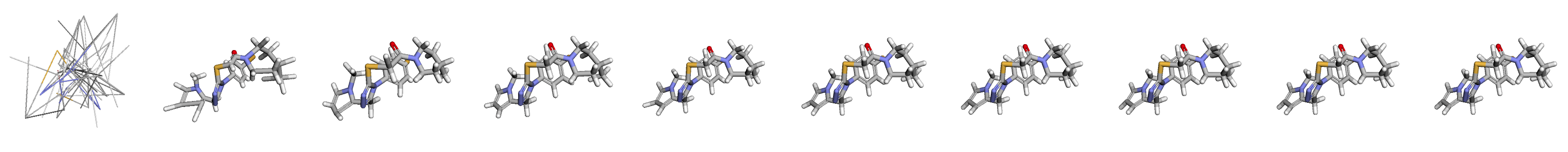}} \\
        \makecell{ODE\\100} & \raisebox{-.4\totalheight}{\includegraphics[width=.9\textwidth]{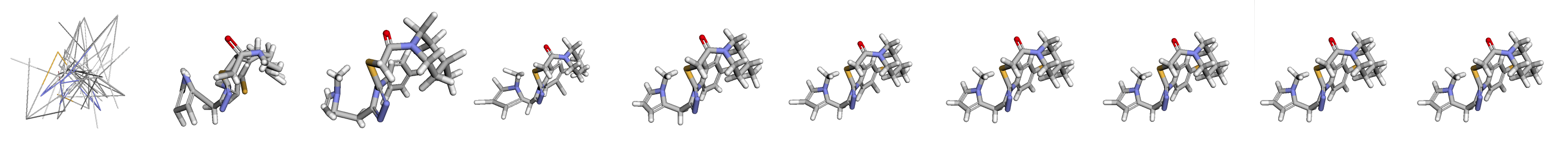}} \\
        \end{tabular}
        }
    \caption{This illustration showcases the sampling trajectories using our proposed SDE and ODE methods, progressing from random noise to coherent molecular structures. For clarity in visualization, we uniformly select and visualize 10 steps.}
    \label{fig: mol-traj}
\end{figure}

In Fig.~\ref{fig: mol-traj}, we depict the sampling trajectories of conformers using our proposed SDE and ODE over 100 and 1000 steps. Analysis reveals that conformers' basic structures are established in the initial half of the process, with only minor adjustments occurring in later steps. This pattern suggests the potential for developing a more expedited sampling method.

\clearpage
\subsection{Human pose generation}\label{app: humanpose experiment}

\subsubsection{Hyper-parameter analysis for human pose} \label{app: hyperparameter of humanpose}
\begin{figure}[h]
    \centering
    \includegraphics[height=.35\textwidth]{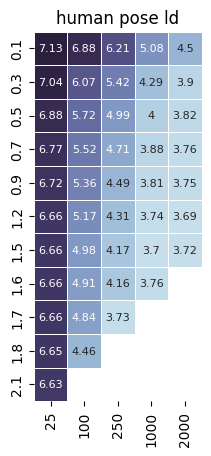}
    \includegraphics[height=.35\textwidth]{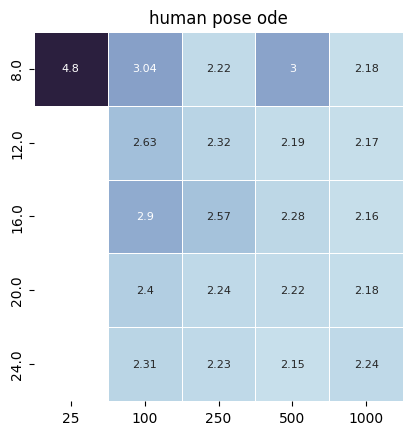}
    \includegraphics[height=.35\textwidth]{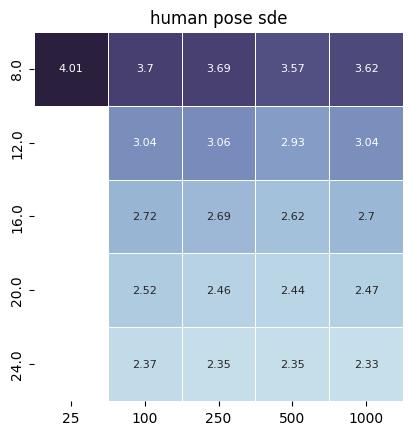}
    \caption{Hyperparameter analysis for human pose task. We investigate the influence of step size variation on sampling via LD and the influence of corrector on proposed SDE and ODE. The horizontal axis represents the number of steps, and the vertical axis represents the value of step size (for LD) or the corrector (for SDE/ODE). Under large values of step size and corrector, the generated human pose can be Nan, resulting unavailable AD score which is shown by the blank cells in the figure.}
    \label{fig: hyperparameter humanpose}
\end{figure}

For the human pose generation task, we conduct a hyperparameter analysis akin to that in Sec.~\ref{app: hyperparameter of molecule}, assessing both our SDE/ODE methods and LD. 
The AD scores are depicted in Fig.~\ref{fig: hyperparameter humanpose}. We observe that for our SDE/ODE sampling method, too large a corrector size fails to generate in small steps and also makes the generated results quite unsatisfactory when the number of steps is large. We choose 8-24 as the search range and find that the corrector size from an appropriate range does not have a large impact on the generated results. For LD, a grid search determines that the step size $\alpha_t$ has a minimal impact on the performance of LD in this context. Excessively large values for step size also lead to failure in generation.

\subsubsection{Visualization of more samples} \label{app: visualizaiton of humanpose}
In this section, we visualize more samples. All the poses are generated from the ``eating'' action, and we show 36 sampling results of each sampling method in various steps. It is observed that LD fails to yield meaningful structures with a small number of steps. However, when the step count exceeds 2000, LD consistently produces reasonable human structures. In contrast, our proposed SDE and ODE can generate meaningful structures in a small number of steps, as few as 50 or 25 steps.

\begin{figure}[h]
    \centering
    \subfigure[Samples from LD.]{
       \begin{tabular}{c c}
         Steps & Samples \\ 
         25 &  \raisebox{-.5\totalheight}{\includegraphics[width=.8\textwidth]{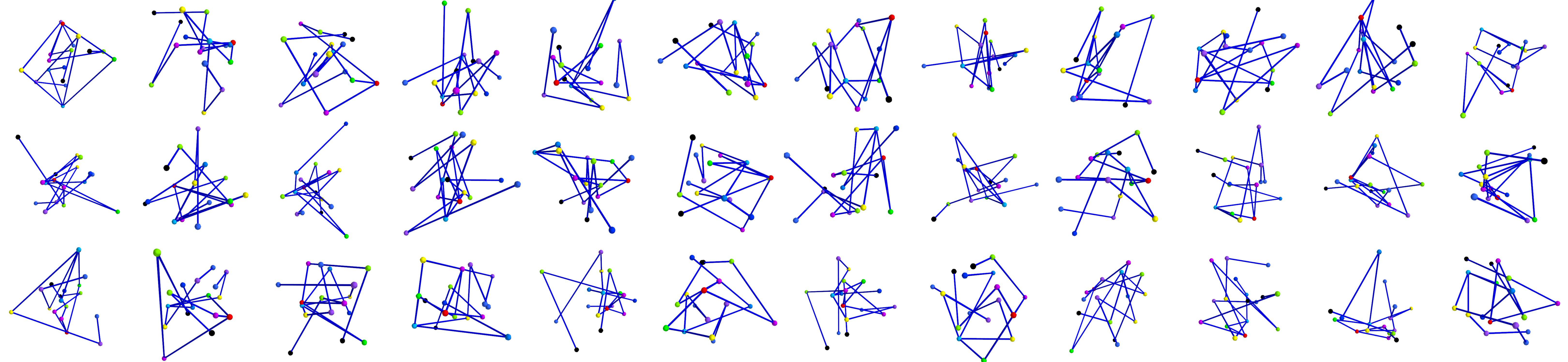}} \\ \hline
         100 & \raisebox{-.5\totalheight}{\includegraphics[width=.8\textwidth]{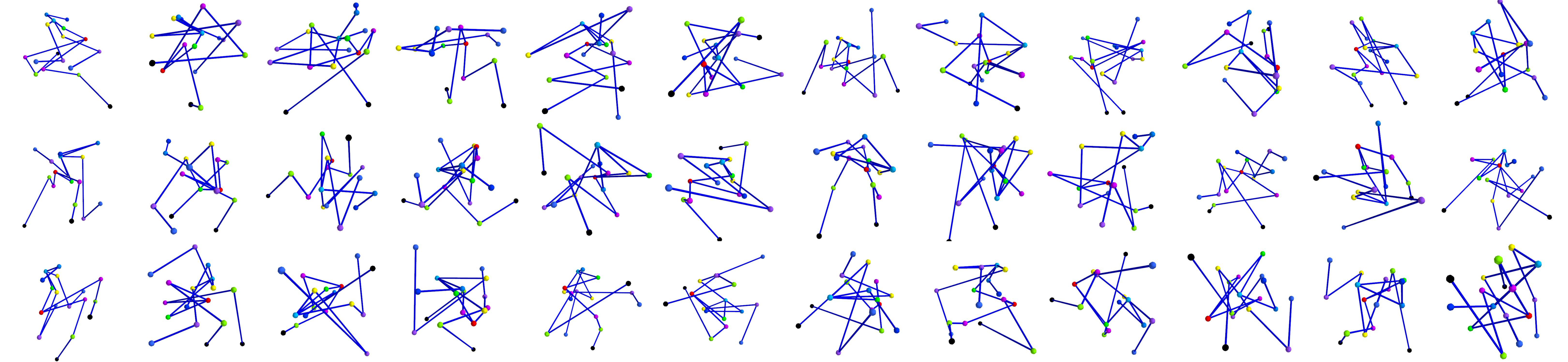}} \\ \hline
         500 & \raisebox{-.5\totalheight}{\includegraphics[width=.8\textwidth]{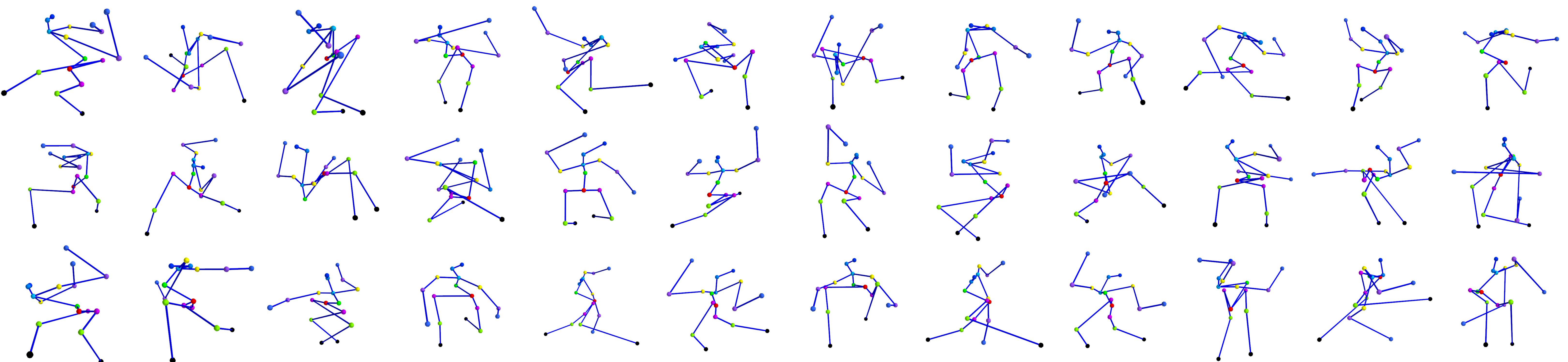}} \\ \hline
         2000 & \raisebox{-.5\totalheight}{\includegraphics[width=.8\textwidth]{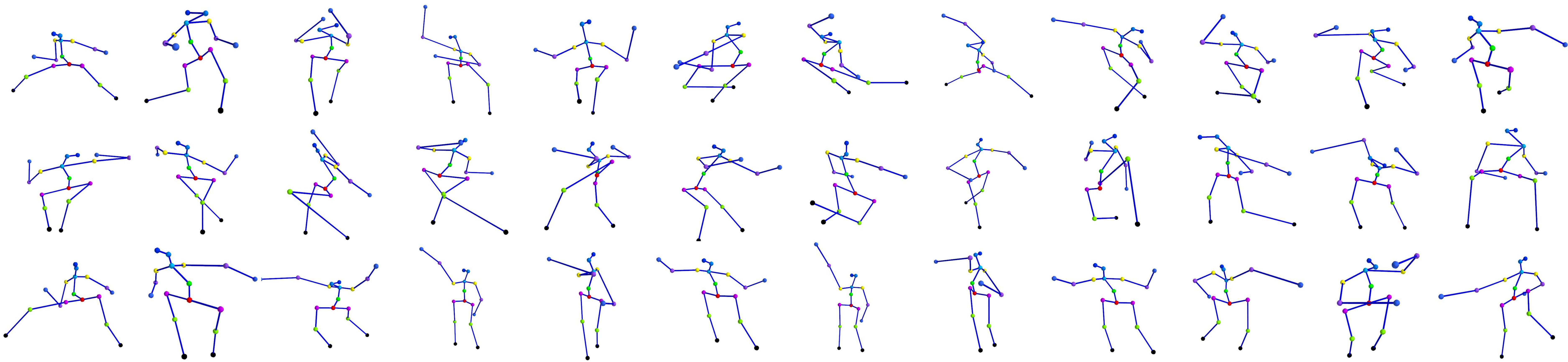}}
        \end{tabular}
        }
    \subfigure[Samples from SDE.]{
       \begin{tabular}{c c}
        25 & \raisebox{-.5\totalheight}{\includegraphics[width=.8\textwidth]{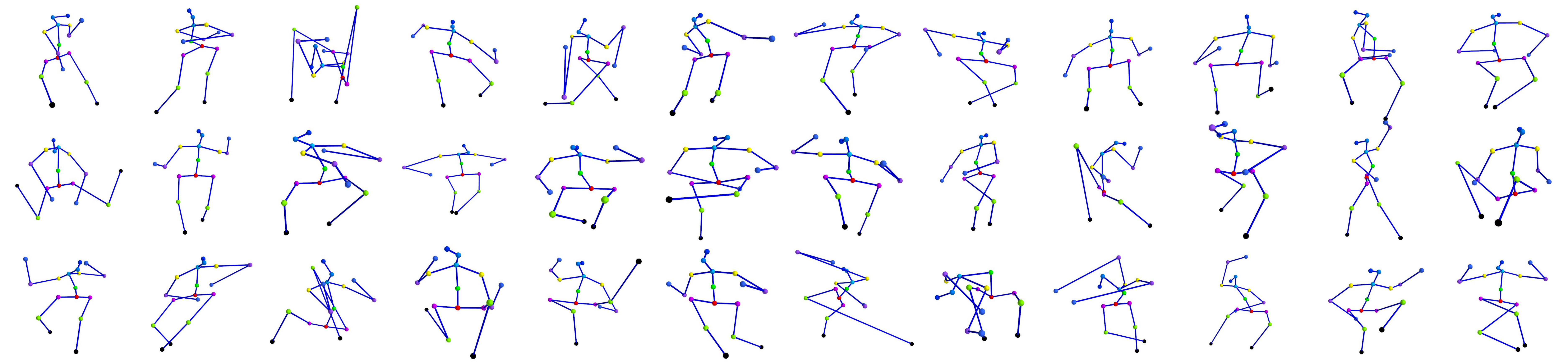}} \\ \hline
        100 & \raisebox{-.5\totalheight}{\includegraphics[width=.8\textwidth]{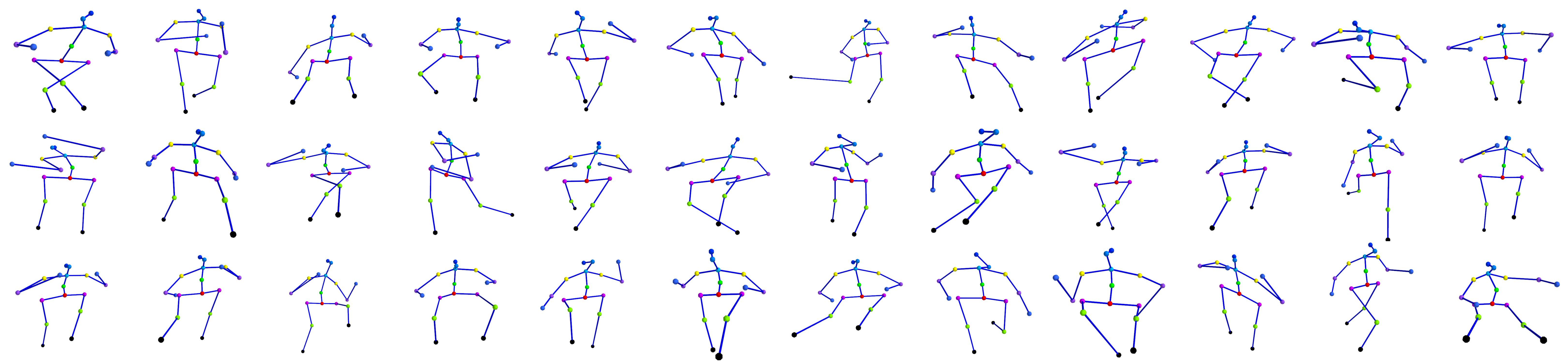}}
        \end{tabular}
    }

\end{figure}

\begin{figure}[h]
    \centering
    \subfigure[Samples from ODE.]{
       \begin{tabular}{c c}
       \midrule
         50 &  \raisebox{-.5\totalheight}{\includegraphics[width=.8\textwidth]{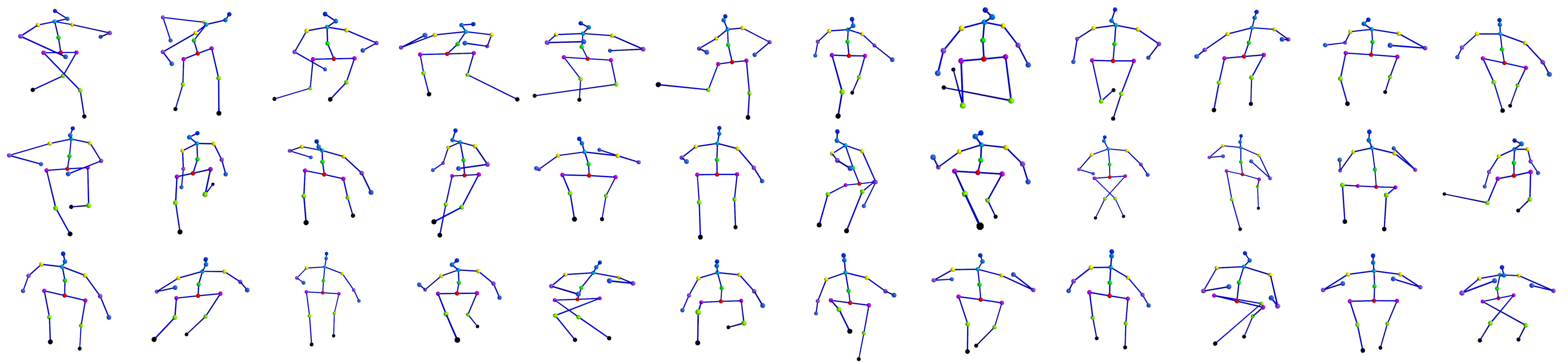}} \\ \hline
         100 &  \raisebox{-.5\totalheight}{\includegraphics[width=.8\textwidth]{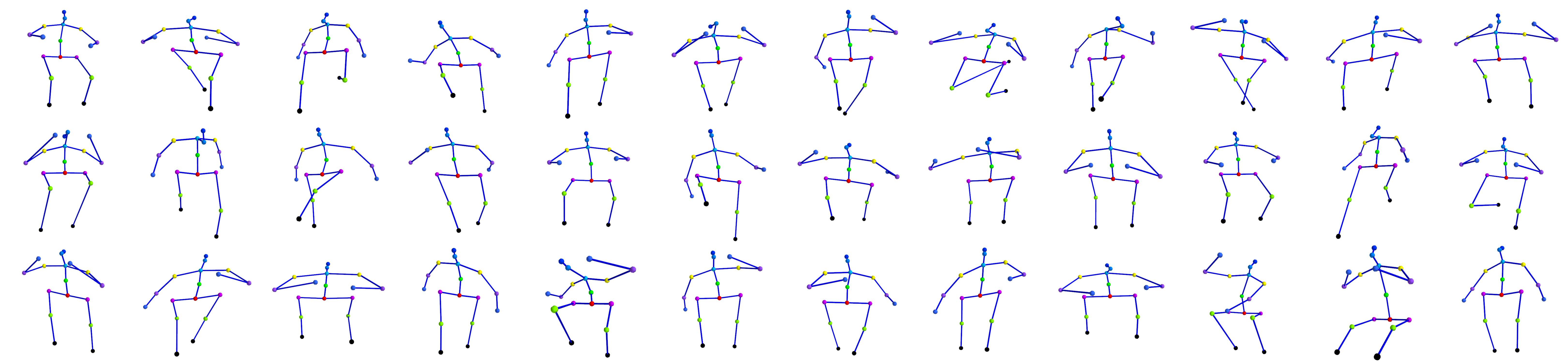}}
        \end{tabular}
    }
    \caption{To avoid occasionality, we randomly generate more samples by different methods in various 100 steps without manual selection to compare different sample methods.}
    \label{fig: Appendix-pose-more}
\end{figure}

\end{document}